\pdfoutput=1
\documentclass[format=acmsmall, review=false, screen=true]{acmart}
\usepackage{booktabs} 
\usepackage{subfigure}
\usepackage{xspace,amsmath,multirow,color,array,colortbl}
\usepackage{natbib}
\usepackage{url}
\usepackage[english]{babel}
\usepackage[ruled]{algorithm2e} 
\usepackage{textcomp}
\usepackage{color}
\usepackage{makecell}

\SetAlFnt{\small}
\SetAlCapFnt{\small}
\SetAlCapNameFnt{\small}
\SetAlCapHSkip{0pt}
\IncMargin{-\parindent}
\newtheorem{problem}{Problem}
\newtheorem{definition}{Definition}

\newcommand{\eat}[1]{}

\newcounter{ccc}

\begin{document}

\title[Passenger Mobility Prediction based on DDW]{Passenger Mobility Prediction via Representation Learning for Dynamic Directed and Weighted Graph}


\author{Yuandong Wang}\orcid{0000-0002-1807-2622}
\affiliation{%
  \institution{School of Computer Science and Engineering, Beihang University}
  \city{Beijing}
  \country{China}}
\email{wangyd@act.buaa.edu.cn}
	
\author{Hongzhi Yin$^*$}
\author{Tong Chen}
\affiliation{%
	\institution{School of Information Technology and Electrical Engineering, The University of Queensland}
	\city{Brisbane}
	\country{AU}}
\email{{h.yin1, tong.chen}@uq.edu.au}

\author{Chunyang Liu}
\author{Ben Wang}
\affiliation{%
	\institution{Didichuxing}
	\city{Beijing}
	\country{China}}
\email{{liuchunyang, wangben}@didiglobal.com}

\author{Tianyu Wo$^*$}
\affiliation{%
	\institution{School of Computer Science and Engineering, Beihang University}
	\city{Beijing}
	\country{China}}
\email{woty@act.buaa.edu.cn}

\author{Jie Xu}
\affiliation{%
	\institution{School of Computing, University of Leeds}
	\city{Leeds}
	\country{UK}
}
\email{J.Xu@leeds.ac.uk}

\thanks{$^*$Corresponding author.}

\begin{abstract}
In recent years, ride-hailing services have been increasingly prevalent as they provide huge convenience for passengers. As a fundamental problem, the timely prediction of passenger demands in different regions is vital for effective traffic flow control and route planning. As both spatial and temporal patterns are indispensable passenger demand prediction, relevant research has evolved from pure time series to graph-structured data for modeling historical passenger demand data, where a snapshot graph is constructed for each time slot by connecting region nodes via different relational edges (e.g., origin-destination relationship, geographical distance, etc.). Consequently, the spatiotemporal passenger demand records naturally carry dynamic patterns in the constructed graphs, where the edges also encode important information about the directions and volume (i.e., weights) of passenger demands between two connected regions.
However, existing graph-based solutions fail to simultaneously consider those three crucial aspects of dynamic, directed, and weighted (DDW) graphs, leading to limited expressiveness when learning graph representations for passenger demand prediction. Therefore, we propose a novel spatiotemporal graph attention network, namely \textbf{Gallat} (\underline{\textbf{G}}raph prediction with \underline{\textbf{all}} \underline{\textbf{at}}tention) as a solution. In Gallat, by comprehensively incorporating those three intrinsic properties of DDW graphs, we build three attention layers to fully capture the spatiotemporal dependencies among different regions across all historical time slots. Moreover, the model employs a subtask to conduct pretraining so that it can obtain accurate results more quickly. We evaluate the proposed model on real-world datasets, and our experimental results demonstrate that Gallat outperforms the state-of-the-art approaches. 
\end{abstract}

%





\maketitle

\renewcommand{\shortauthors}{Wang and Yin et al.}

\section{introduction}\label{sec:introduction}
Transportation plays a very important role in our daily lives. In 2018, commuters in Beijing spend about 112 minutes behind the wheel on average every day. With the prominent development of technologies like GPS and mobile Internet, various ride-hailing applications have emerged to provide drivers and passengers with more convenience, such as Didi, Lyft and Uber. For all ride-hailing platforms, analyzing and predicting the real-time passenger demand is the key to high-quality services, which has recently started to attract considerable research attention.

Initially, the majority of studies \cite{tong2017simpler, yao2018deep, wang2019unified} treat passenger demand prediction as a time series prediction problem which predicts the number of passenger demands in an arbitrary location during a given time period. However, such prediction paradigm only considers the origin of each passenger's trip, and neglects the destination information. To account for this important aspect in passenger demand prediction, recent research \cite{deng2016latent,hu2020stochastic} defines it as an Origin-Destination Matrix Prediction (ODMP) problem. In ODMP, each time slot has its own OD matrix, where the element indexed by $(i,j)$ describes the travel demand from region $i$ to region $j$. In this regard, tensor factorization models \cite{gong2018network, hu2020stochastic} and convolutional neural networks (CNNs) \cite{li2018diffusion, liu2019contextualized} can be conveniently adopted to extract latent representations from the OD matrices to support prediction. Additionally, ODMP not only estimates the number of passenger demands within the target area, but also foresees where these demands go, making it easier for ride-hailing companies to coordinate travel resources (i.e., vehicles) to maximize customer satisfaction and business revenue. 

More recently, research on passenger demand has introduced a new perspective by modelling the traffic data as graphs where different regions are viewed as nodes. Compared with OD matrices, graph-structured data is able to uniform various heterogeneous information to help boost the prediction accuracy. For instance, two region nodes can be connected by different edges representing specific relationships, e.g., origin-destination relationship in a time slot \cite{shi2020predicting}, geographical association (e.g., adjacency regions) \cite{wang2019origin}, and even functionality similarity by comparing their point-of-interest distributions \cite{geng2019spatiotemporal}. With the recent advances in graph neural networks (GNNs) \cite{hamilton2017inductive, velivckovic2017graph,kipf2016semi}, GNN-based models have emerged and yielded state-of-the-art performance in a wide range of passenger demand prediction and traffic modelling tasks \cite{shi2020predicting,wang2019origin,geng2019spatiotemporal,cui2018traffic}. On one hand, GNN-based models are capable of mining the complex and heterogeneous relationships between regions, thus thoroughly capturing the latent properties of each region to allow for accurate demand prediction. On the other hand, GNN-based models are versatile as they generalize the convolutional operations in CNNs to the non-Euclidean graph topologies without the need for conversion into OD matrices at each time slot. 

When modelling traffic data as graphs, a common practice is to construct a snapshot graph for each time slot. Consequently, this results in three major intrinsic properties of such graph-structured data, i.e., the constructed graphs are \textit{dynamic}, \textit{directed}, and \textit{weighted}. In our paper, this notion is referred to as dynamic, directed, and weighted (DDW) graphs, which contain both the spatial and temporal information across regions. From a temporal perspective, DDW graphs are time-sensitive due to complex real-life situations (e.g., peak hour traffic, special events, etc.), making it non-trivial to fully capture their dynamics. For every DDW snapshot graph, two regions are linked via an edge if there are observed trip orders between them, thus allowing GNN-based models to capture signals of spatial passenger flows. However, on top of that, in an origin-destination relationship, the edge between two region nodes should be directional. For example, a central business district tends to have substantially more inbound traffic flows than outbound flows during morning peak hours. Also, as the volume of passenger demand varies largely among different routes, those origin-destination edges should preserve such information as their weights. Unfortunately, existing methods tend to oversee these two important edge properties, leading to severe information loss. For instance, graph-structured data is used to extract temporal features of each region \cite{geng2019spatiotemporal}, but neither the direction nor the volume of passengers is captured by the constructed graphs. Though \cite{wang2019origin} considers the weights of edges when learning representations for each region nodes, it simply treats the passenger flows between two nodes from both directions equally without distinguishing their semantics. 

Meanwhile, as a widely reported issue in passenger demand research \cite{xue2015solving, wang2019origin,hu2020stochastic}, the geographically imbalanced distribution of trip orders inevitably incurs sparsity issues within the constructed graphs. Due to varied locations and public resource allocations, the observable passenger flows from/to some regions (e.g., rural areas) are highly scarce. While most existing studies seek solutions by gathering side information from multiple data sources (e.g., point-of-interests \cite{geng2019spatiotemporal,wang2019unified}, weather \cite{liao2018deep, wang2019origin}, real-time events \cite{tong2017simpler}, etc.), it is impractical to assume the constant availability and high quality of such auxiliary data. Hence, it further highlights the necessity of fully capturing the latent patterns and comprehensively modelling the information within DDW graphs constructed from historical trip orders. 

To this end, we propose Gallat, namely \underline{\textbf{G}}raph prediction with \underline{\textbf{all}} \underline{\textbf{at}}tention, which is a novel spatiotemporal graph neural network for passenger demand prediction. Specifically, having the self-attention as its main building block, Gallat consists of three main parts: the spatial attention layer, the temporal attention layer, and the transferring attention layer. In the spatial attention layer, we learn each region node's representation by discriminatively aggregating information from its three types of neighbor nodes with attention mechanism. To be specific, we innovatively define forward and backward neighborhoods respectively for its outbound and inbound travel records to distinguish the different semantics of two directions. The geographical neighborhood is also defined for each node to gather information from nearby regions and help alleviate data sparsity. 
In the temporal attention layer, we deploy a multi-channel self-attention mechanism to retrieve important contextual information from the history, which is then used to augment the learned node representations for demand prediction in the next time slot. 
With the generated representation of each node from the first two layers, we firstly predict the total number of passenger demands in each region, and then distribute it to all possible destination regions via a transferring probability generated in the final transferring attention layer.
By taking advantages of the attention mechanism, our model Gallat is highly expressive, and is able to capture the heterogeneous connectivity among regions to yield optimal prediction performance.

The main contributions of this work are as follows:
\begin{itemize}
	\item We investigate passenger demand from the perspective of DDW graphs, with a comprehensive take on modelling the dynamic, directed, and weighted properties within the passenger flow data simultaneously.
	\item We propose a novel spatiotemporal graph neural network named Gallat, which is an inductive solution to representation learning on DDW graphs for passenger demand prediction.
	\item Extensive experiments are conducted on two real-world datasets, and the results demonstrate the  superior effectiveness of our proposed model.
\end{itemize}

\section{Preliminaries}\label{sec:problem}
In this section, we provide key definitions and formally define the passenger mobility prediction problem.
\begin{definition}\label{sec:time}
	\textit{\textbf{Time Slot.}} We evenly partition the time into a sequence of $T$ slots, which are represented as $t\in\{1, 2, ... , T\}$. The interval between any two consecutive slots is constant. For example, we can divide a day into 24 one-hour time slots.
\end{definition}

\begin{definition}\label{sec:nodes}
	\textit{\textbf{Node.}} 
	The entire area of interest like a city is divided into $n$ non-overlapping regions. Each region is regarded as a node, and the node set of the specific city can be denoted as $\mathcal{V}=\{v_1, v_2, ..., v_{n}\}$. Following \cite{wang2019origin,hu2020stochastic,yao2018deep}, we determine regions by evenly diving a whole area into grids according to their longitudes and latitudes. Then, we calculate the physical distance between each pair of nodes using their central coordinates, which is stored in an adjacency matrix $\textbf{R}$. Every element $r_{i,j}\in \textbf{R}$ represents the geographical distance between node $v_i$ and node $v_j$.
\end{definition}

\begin{definition}\label{sec:ddw}
	\textit{\textbf{Dynamic, Directed, and Weighted graph.}} In a time slot, the passenger mobility in the region of interest can be modeled as interactions between nodes. Given a fixed region node set $\mathcal{V}=\{v_1, v_2, ..., v_{n}\}$, we use $e_{ij}=(v_i \rightarrow v_j, g_{ij})$ to denote the directional edge from $v_i$ to $v_j$, where $g_{ij}$ is the weight of the edge. In our paper, each $g_{ij}$ is directly defined as the number of passenger demands from region $v_i$ to $v_j$ in a specific time slot. If there are no trip orders from $v_i$ to $v_j$, then $g_{ij}=0$ denotes a non-existing edge. We use a sequence of adjacency matrices $\{\textbf{G}_t\}_{t=1}^{T}=\{\textbf{G}_1, \textbf{G}_2, ..., \textbf{G}_T\}$ to represent all DDW graphs in all $T$ time slots where $\textbf{G}_t = \{g_{ij}^t\}_{i,j=1}^{n} \in \mathbb{N}^{n \times n}$. 
\end{definition}
\vspace{-1ex}
\begin{figure*}[th!]
	\centering
	\setlength{\abovecaptionskip}{0.15cm} 
	\includegraphics[width=1.0\textwidth]{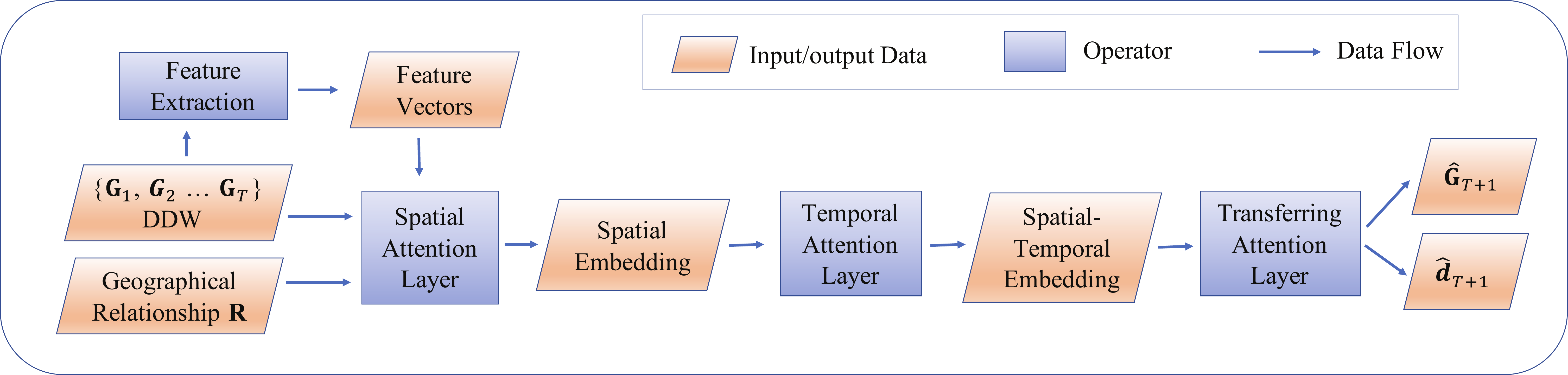}
	\caption{\textbf{The overview of Gallat model.} This figure depicts the key components and data flow of the model.} 
	\label{fig:model} 
\end{figure*}

Here, we formulate the passenger mobility prediction problem as follows:
\begin{problem}
	\textit{\textbf{Passenger Mobility Prediction.}} For a fixed region node set $\mathcal{V}$, given all DDW snapshot graphs $\{\textbf{G}_t\}_{t=1}^{T}$ in the past $T$ time slots and the geographical relationship $\textbf{R}$ among nodes, we define passenger mobility prediction as a DDW graph prediction problem, which aims to predict the DDW  snapshot graph $\textbf{G}_{T+1}$ in the next time slot.
\end{problem}

\begin{table}[h!]
	\centering
	\caption{\textbf{Notations}}
	\label{table:notations}
	\begin{tabular}{m{4cm}<{\centering}|m{9cm}<{\centering}}
		\toprule 
		Notation               &Description                                        \\ \hline 
		$t$                         & a time slot      \\\hline 
		$T$                        & the total number of  time slots in a sequence \\\hline 
		$\mathcal{V}$ &a node set  \\\hline 
		$v_i$                     & a node \\\hline 
		$\textbf{V}_t $                          &the matrix of  all nodes' feature vectors at $t$                \\\hline
		$\textbf{v}_i^t $   & the feature vector of node $v_i$ at time slot $t$ , $\textbf{v}_i^t  \in  \textbf{V}_t $         \\\hline
		
		$\textbf{R}$     & an adjacency matrix of  nodes' geographical relationship               \\\hline 
		$r_{i,j}$              & the geographical distance between node $v_i$ and $v_j$,   $r_{i,j} \in \textbf{R}$               \\\hline 
		$\textbf{G}_t$ & an adjacency matrix of DDW graphs at time slot $t$        \\\hline 
		$g_{ij}^t$          & the passenger demand between node  $v_i$ and $v_j$,   $g_{ij}^t \in \textbf{G}_t$                \\\hline 
		
		$\Psi^{t}_i , \Phi^{t}_i, \Theta_i $  & the sets of node $v_i$'s forward, backward and geographical neighbors' indexes at time slot $t$             \\\hline 
		$L$                  &the threshold of the distance to determine the size of $v_i$'s  geographical neighborhood                \\\hline 
		$\textbf{M}_t$          &the matrix of all nodes' representation vectors               \\\hline 
		$m_i^t$        &node $v_i$'s representation vector at time slot $t$,  $m_i^t \in  \textbf{M}_t$               \\\hline 
		$\psi_{ij}^{t}$, $\phi_{ij}^{t}$ and $\theta_{ij}^{t}$         &  attentive weights between  three different neighbors  of $v_i$ and $v_j$             \\\hline 
		$a_{j}^t$, $b_{j}^t$ , $c_{j}$          &pre-weights of  different kinds of  neighbors                \\\hline 
		$P$          &the number of historical time slots considered in each channel               \\\hline 
		$\mathcal{S} $          &  a channel's sequence of  $\textbf{M}_t$, $\mathcal{S} \in \{\mathcal{S}_1, \mathcal{S}_2, \mathcal{S}_3, \mathcal{S}_4\}$             \\\hline 
		$\textbf{M}_{\mathcal{S}}$          & the aggregated representation of the sequence $\mathcal{S} $               \\\hline 
		$\textbf{M}'_{T}$          &the final representation of time slot $T$                \\\hline 
		$\textbf{V}_{T+1}$          &the matrix that stores n nodes' features at time slot $T+1$                \\\hline 
		$\textbf{d}_{T+1}$, $\textbf{G}_{T+1}$          &the outbound passenger demands and the snapshot of DDW graph at time slot $T+1$                \\\hline 
		$\hat{\textbf{d}}_{T+1}$, $\hat{\textbf{G}}_{T+1}$          &the predicted results of $\textbf{d}_{T+1}$ and $\textbf{G}_{T+1}$             \\\hline 
		$\hat{d_{i}}$, $\hat{g_{ij}}$ & the elements of  $\hat{\textbf{d}}_{T+1}$ and $\hat{\textbf{G}}_{T+1}$  \\\hline 
		$q_{ij}$          & the transferring probability between node $v_i$ and $v_j$               \\\hline 
		$\mathcal{L}, \mathcal{L}_{d}, \mathcal{L}_{o}$          &loss functions                \\\hline 
		$\eta_{d}, \eta_{o}$          &the weights of loss functions                \\\hline 
		$\textbf{a}, \textbf{w}, \textbf{b} $          &the weights to learn               \\\hline 
		$d, d_e, d_v $          &the dimensions of feature vectors during $\{1, ..., t\}$, embedding layers and feature vectors at time slots $T+1$               \\\hline 
		$\textbf{W},$ $\textbf{W}_s,$  $\textbf{W}_a, $ $\textbf{W}'_a, $ $\textbf{W}^K_{\mathcal{S}},$ $\textbf{W}^Q_{\mathcal{S}},$  $\textbf{W}^V_{\mathcal{S}}, $  $\textbf{W}'^K_{\mathcal{S}},$ $\textbf{W}'^Q_{\mathcal{S}},$ $\textbf{W}'^V_{\mathcal{S}}$         &the weight matrices to learn                \\\hline 
		\bottomrule
	\end{tabular}	
\end{table}

\section{Solution}\label{sec:solution}

In this section, we present the detail of our model Gallat, a spatiotemporal attention network for passenger demand prediction. Figure \ref{fig:model} depicts the overview of our proposed model. With the DDW graph sequence $\{\textbf{G}_t\}_{t=1}^T$ and the geographical relationship $\textbf{R}$ as the model input, the feature extraction module generates the feature vector $\textbf{v}^t_i$ for node $v_i$ at the $t$-th time slot. Then, we define a spatial attention layer to learn a spatial representation (i.e., embedding) for every node at time $t$ by aggregating information from nodes within three distinct types of spatial neighborhoods. Afterwards, all nodes' representations will be fed into a temporal attention layer, which updates its current embedding with the captured temporal dependencies among its historical embeddings. In our transferring attention layer, we first calculate the total number of outbound passenger demands departing from each region and the transferring probability between every pair of nodes with the current embeddings, and then use the resulted probabilities to map each region's total passenger demands to the corresponding destination regions, which will compose the information in the next DDW snapshot graph $\textbf{G}_{T+1}$. 

\subsection{Feature Extraction}
Given a region node $v_i$, we first construct its feature vector by merging relevant information from multiple sources. Specifically, the feature vector $\textbf{v}_i^t \in \mathbb{R}^d$ for node $v_i$ at time $t$ is the concatenation of feature embeddings from all feature fields (e.g., weather, day of the week, etc.):
\begin{equation}
	\label{eq:embedding}
	\textbf{v}_i^t = \textbf{v}_i^{t,1}\oplus \cdots \oplus \textbf{v}_i^{t,f} \oplus \cdots \oplus \textbf{v}_i^{t,F},	
\end{equation}
where $F$ is the total number of feature fields, and $\oplus$ denotes the concatenation operation. Note that $\textbf{v}_i^{t,f}$ can be either a dense embedding vector for categorical features (e.g., node ID) or a real-valued number for continuous features (e.g., temperature). As such, for each snapshot DDW graph $\textbf{G}_t$, we can obtain the features of all $n$ nodes $\textbf{V}_t = \{ \textbf{v}_i^t\}_{i=1}^n$ to support subsequent computations. 
{In our experiments, a node $v_i$'s feature vector $\textbf{v}_i^t$ is the concatenation of  its row, column, out degree, in degree in the graph $\textbf{G}_t$, as well as the embeddings of the node ID, time slot $t$ and the corresponding day of a week.} 

\subsection{Spatial Attention Layer}\label{sec:spatial}
\begin{figure*}[tb!]
	\centering
	\setlength{\abovecaptionskip}{0.15cm}
	\subfigure[Forward neighbors at 3:00am]{
		\label{fig:degree:sub1}
		\includegraphics[width=0.44\textwidth]{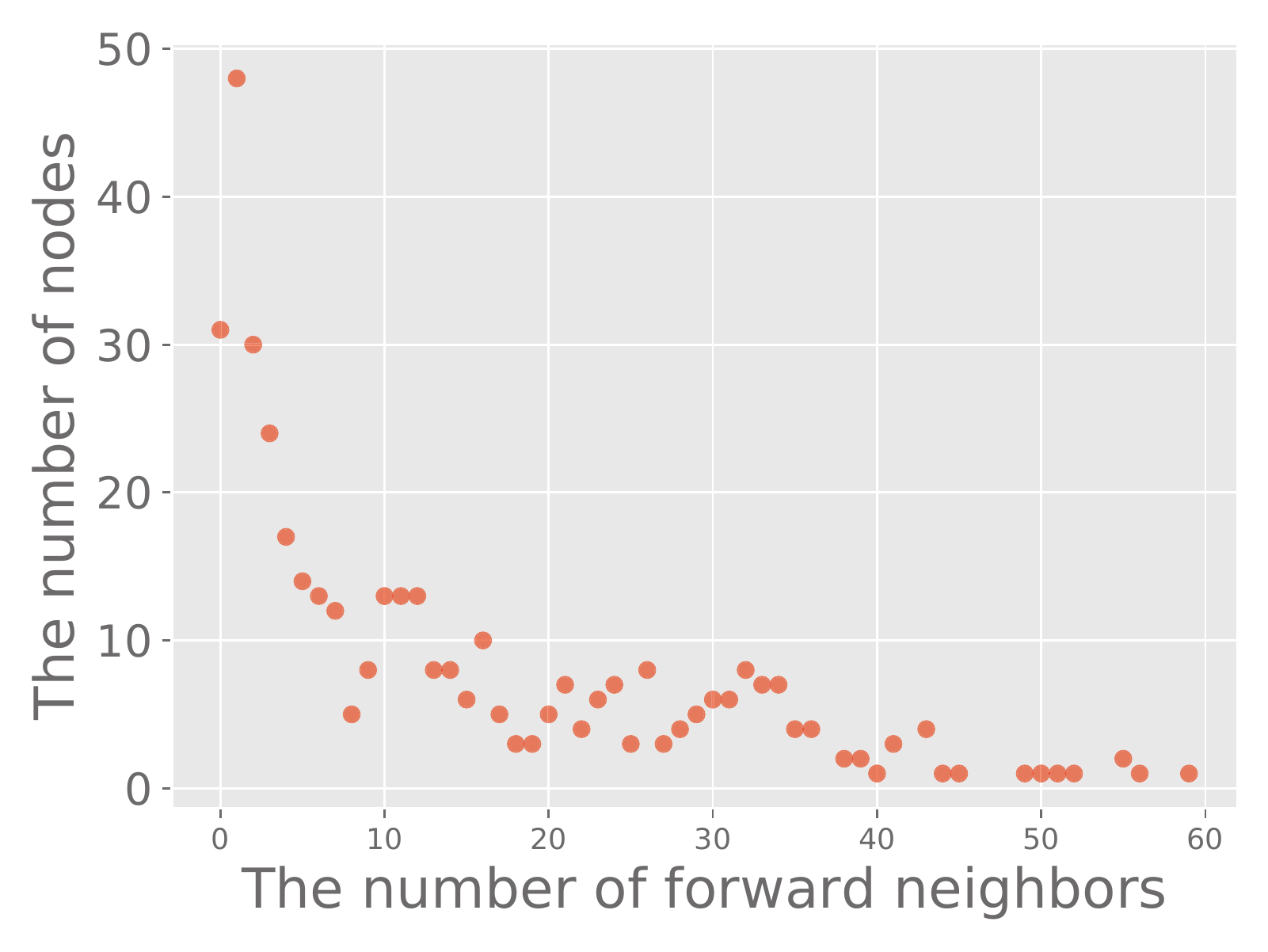}}
	\subfigure[Backward neigbors at 3:00am]{
		\label{fig:degree:sub2}
		\includegraphics[width=0.44\textwidth]{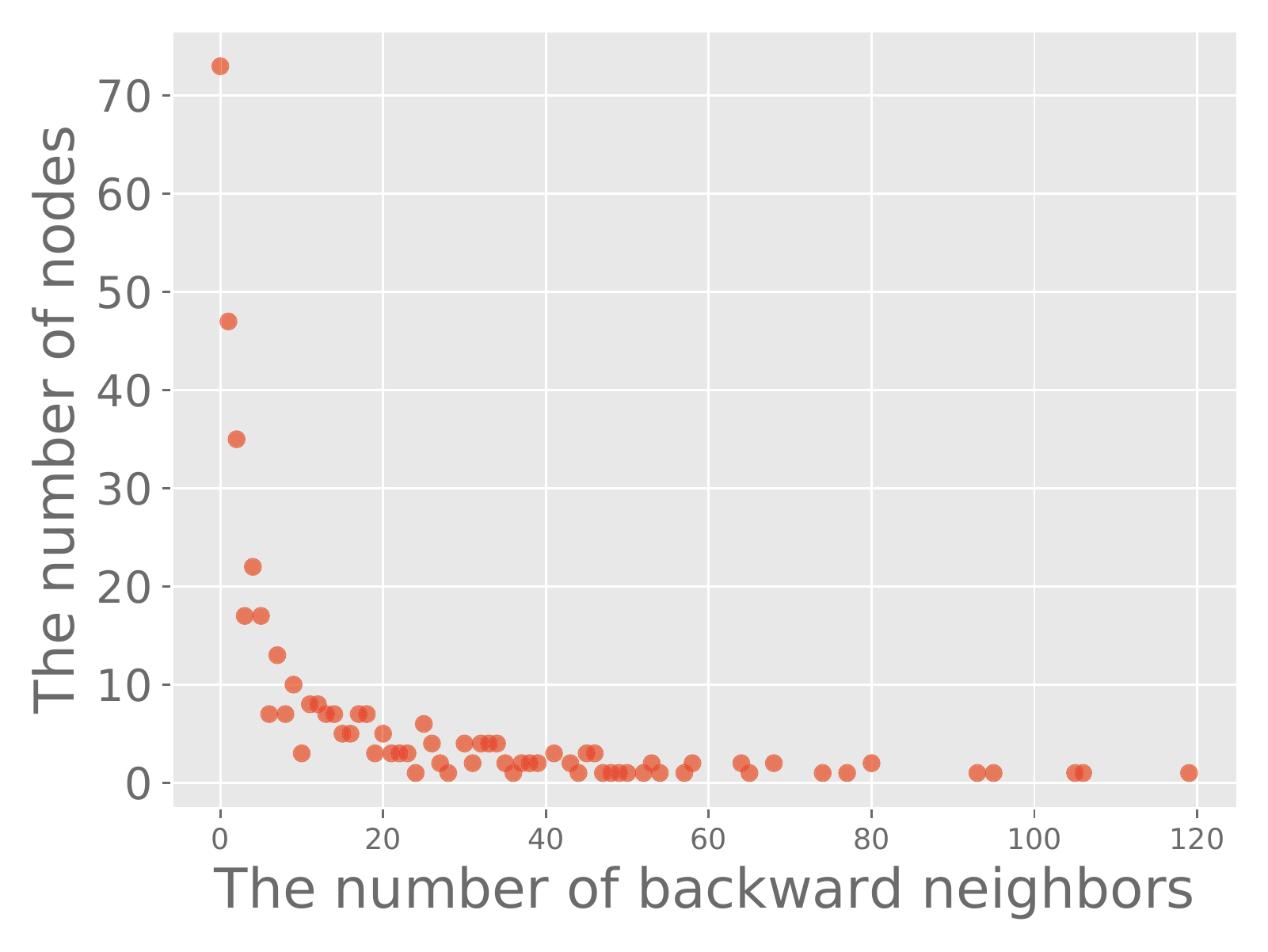}}
	\subfigure[Forward neighbors at 9:00am]{
		\label{fig:degree:sub3}
		\includegraphics[width=0.44\textwidth]{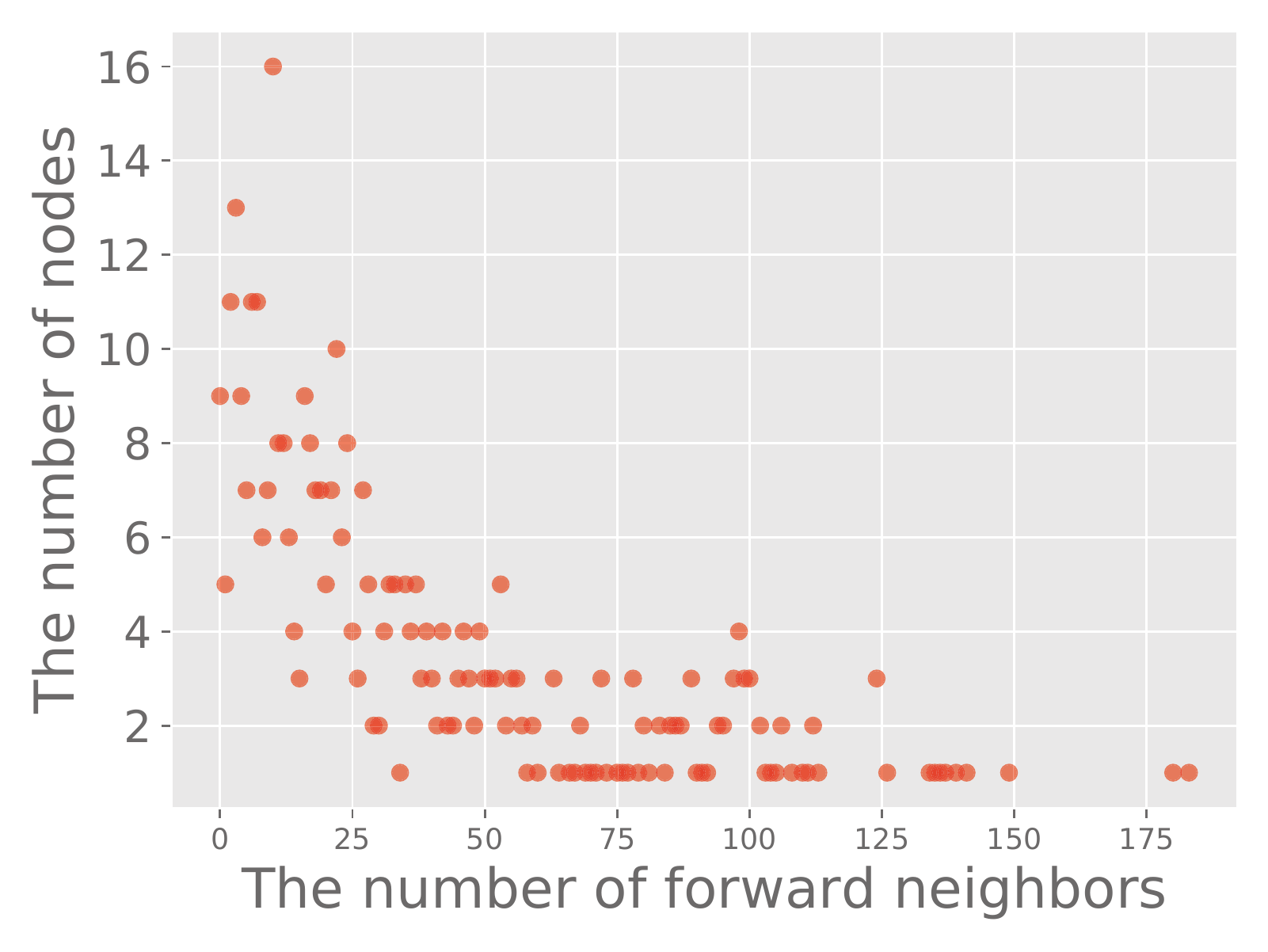}}
	\subfigure[Backward neighbors at 9:00am]{
		\label{fig:degree:sub4}
		\includegraphics[width=0.44\textwidth]{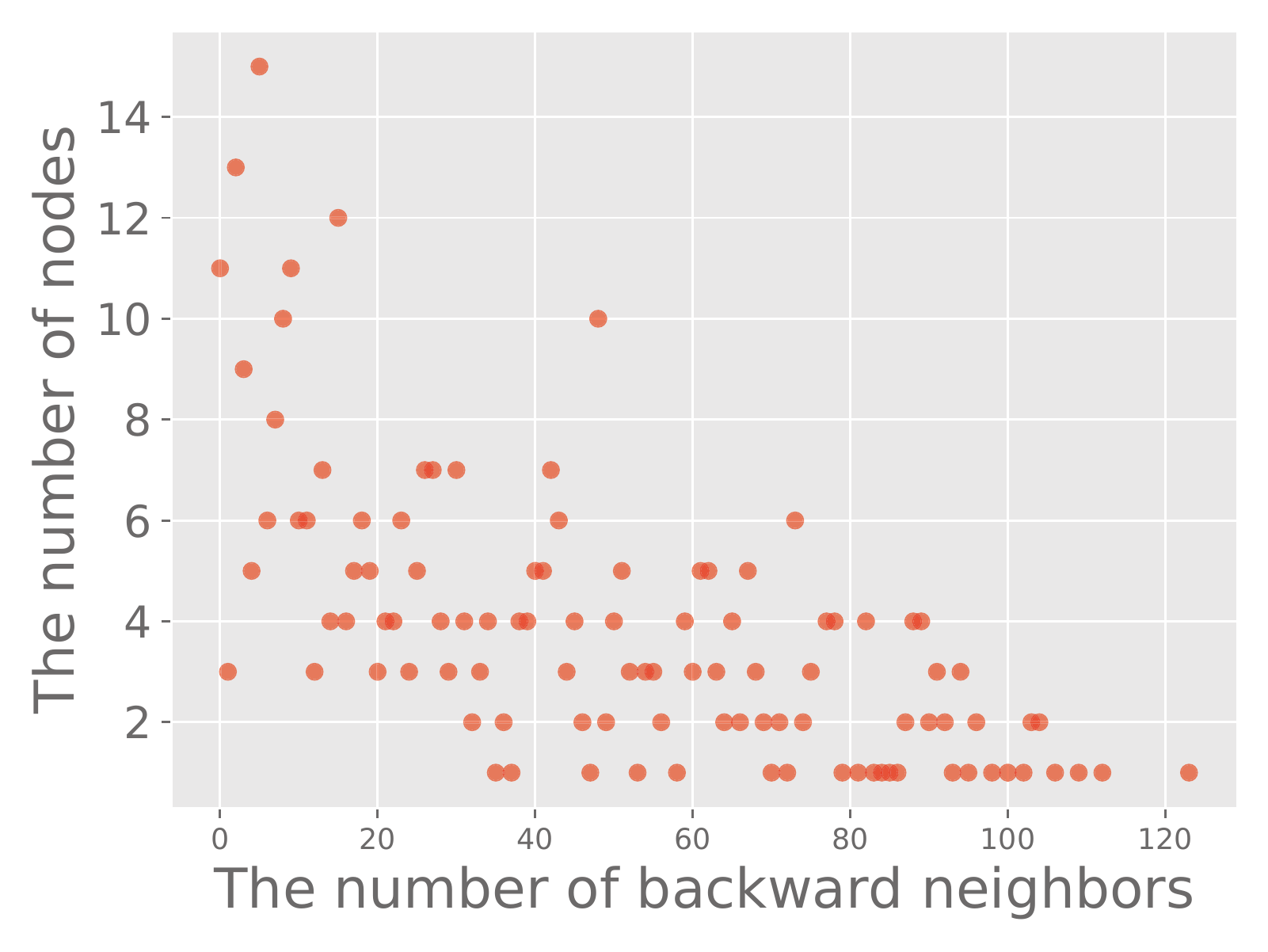}}
	\caption{\textbf{The forward and backward neighbors' distribution at different time slots}}
	\label{fig:degree} 
\end{figure*}

In this section, inspired by the inductive graph representation learning approach~\cite{hamilton2017inductive}, we learn a latent representation $\textbf{m}_i$ for node $v_i$ at time $t$ by effectively aggregating the features of its neighbors. Different from \cite{hamilton2017inductive} which only focuses on a single type of neighbors, we define three types of neighbors in DDW graphs, namely forward neighbors and backward neighbors based on passenger mobility, and geographical neighbors based on physical locations. 
{We make the statistics of real-world data in Beijing during one month. As shown in Figure \ref{fig:degree}, we summarize the mean values of a non-peak time slot (3:00am) and a peak one (9:00am). It can be seen that the forward neighbors and backward neighbors exhibit different distributions in the same time slot and the same type of neighbors also show different distributions at different time slots. Hence, it is meaningful to distinguish the forward and backward neighbors. }We first give the definition of forward and backward neighborhoods for each node as follows.
\begin{figure}[th!]
	\centering
	\setlength{\abovecaptionskip}{0.15cm} 
	\includegraphics[width=0.80\textwidth]{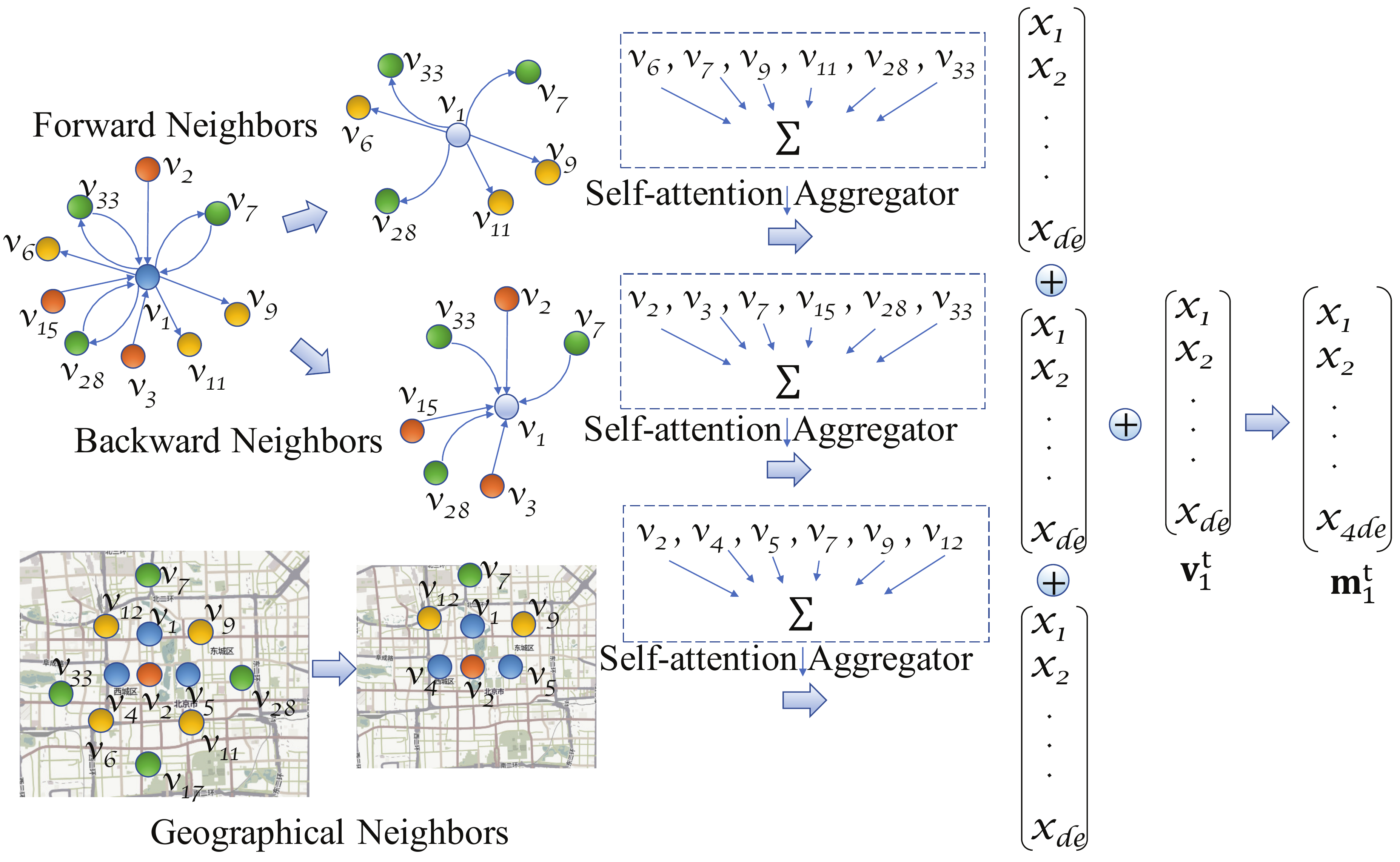}
	\caption{\textbf{The illustration of spatial attention layer.} This figure describes the neighborhood definition and aggregation in spatial attention layer. $\sum$ means calculating the attention weighted and element-wise mean of the neighbors' feature vectors, and $\oplus$ represents the concatenation of vectors.} 
	\label{fig:spatial}
\end{figure}
\subsubsection{\textbf{Forward Neighborhood}}\label{sec:forward}
If there is at least one demand starting from region node $v_i$ and ending at $v_j$, i.e., $g_{ij}^t > 0$, then $v_j$ is a forward neighbor of $v_i$. For node $v_i$, its forward neighborhood at an arbitrary time slot $t$ is a set of node indexes defined as:
\begin{equation}
	\label{eq:fn}
	\Psi^{t}_i = \{ j | g_{ij}^t > 0, g_{ij}^t \in \textbf{G}_{t}\}.
\end{equation}  

\subsubsection{\textbf{Backward Neighborhood}}\label{sec:backward}
Similarly, if there is at least one demand starting from region node $v_j$ and ending at $v_i$, then $v_j$ is a backward neighbor of $v_i$. For node $v_i$, at an arbitrary time slot $t$, we can obtain a set of its backward neighbors' indexes via:
\begin{equation}
	\label{eq:bn}
	\Phi^{t}_i = \{ j | g_{ji}^t > 0, g_{ji}^t \in \textbf{G}_{t}\}.
\end{equation}  

According to Eq.(\ref{eq:fn}) and Eq.(\ref{eq:bn}), it is worth mentioning that the numbers of different nodes' forward and backward neighbors are asymmetrical and time-dependent. 
The rationale of defining forward and backward neighborhoods is that the characteristics of each region are not only determined by its intrinsic features, but also affected by its interactions with other regions. Intuitively, if more trip orders are observed between nodes $v_i$ and $v_j$ at time $t$, they are more likely to possess a higher semantic affinity in that time slot. Hence, by propagating the representations of node $v_i$'s neighbors to itself, the properties from closely related nodes can complement its original representation, thus producing an updated embedding of node $v_i$ with enriched contexts. Moreover, when modelling passenger demands as DDW graphs, the direction of an edge carries crucial information about different mobility patterns, and indicates varied functionalities of regions. For example, it is common to see a residential area has many people travelling to the central business district for work during the morning rush hours; while in the evening when people return home from work, a large passenger flow may be observed in the reverse direction. Hence, for two linked nodes, unlike \cite{wang2019origin} that treats the passenger demands from both directions equally, we define the forward and backward neighbors to distinguish their semantics and allow them to contribute differently to the resulted node embedding. 

Apart from that, we also capture the information from geographically connected regions in the spatial attention layer. Geographical neighbors of region node $v_i$ are nodes that are physically close to it, which are constant across $T$ time slots.
\subsubsection{\textbf{Geographical Neighborhood}}\label{sec:geograohical}
As Definition \ref{sec:nodes} states, the nodes in our DDW graphs are manually-divided and non-overlapping grids with all pairwise geographical relationship $\textbf{R}$. For a node $v_i$, its static geographical neighborhood $\Theta_i$ is formulated as:
\begin{equation}
	\label{eq:gn}
	\Theta_i = \{j | r_{i, j} \leq L , r_{i,j} \in \textbf{R}\},
\end{equation}
where $L$ is a threshold of the distance to determine the size of $v_i$'s neighborhood. A node's geographical neighbors are important for its embedding. Intuitively, being the geographical neighbors of $v_i$, the nodes in $\Theta_i$ are more likely to own similar inner properties as $v_i$, thus leading to close distributions of passenger demands. For example, if node $v_i$ and $v_j$ are adjacent and located in a sparsely populated suburb, then both of them are likely to have fewer demands. The geographical neighbors also help alleviate the data sparsity problem. For example, if a node has very few forward and backward neighbors at a specific time $t$, the features of its geographical neighbors will become a key supplementary information resource to ensure the learning of discriminative node representation.

It is worth mentioning that these three kinds of neighborhoods carry different information from both semantic and geographical perspectives. They need to be transfered to the next step of the model separately so we utilize the attention-based aggregator to gather information within each kinds of neighborhood and concatenate the aggregating results of them in the following steps.

\subsubsection{\textbf{Attention-based Aggregator}}
With three types of node neighborhoods defined, we devise an attention-based aggregator to merge the respective node information within $\Psi^{t}_i$, $\Phi^{t}_i$ and $\Theta_i$. Taking the feature vectors of all nodes in the neighbor sets, i.e., $\{\textbf{v}_j|j\in \Psi^{t}_i\}$,  $\{\textbf{v}_j|j\in \Phi^{t}_i\}$ and $\{\textbf{v}_j|j\in \Theta_i\}$ as the input, the attention-based aggregator fuses them into a unified vector representation $\textbf{m}^t_i$ for each node $v_i$ at time $t$. 
Before detailing our attention-based aggregator for node embedding, we first briefly introduce the naive form of aggregator introduced in~\cite{hamilton2017inductive}, which is originally designed for static graphs without any direction or weight information on edges:
\begin{equation}
	\label{eq:mean}
	\textbf{m}^t_{i}=\sigma \Big{(}\textbf{W} \cdot  \big{(}\textbf{v}^t_{i} \oplus Aggregate\{\textbf{v}^t_{j}|j \in \mathcal{N}\}  \big{)} \Big{)},
\end{equation}
where $\textbf{m}_{i}^t$ represents the resulted embedding vectors of node $v_i$, $\textbf{W}$ is the weight matrix to learn, and $\sigma$ is the nonlinear $Sigmoid$ activation function. To aggregate the information passed from neighbor nodes using the aggregation function $Aggregate(\cdot)$, the GraphSAGE proposed in \cite{hamilton2017inductive} creates a paradigm that allows the model to sample and aggregate a fixed number of neighbors $\mathcal{N}$. However, GraphSAGE uniformly samples a relatively small number of  neighbors, which treats all neighbor nodes evenly and neglects their varied importance of  a node's neighbors. Also, for a popular region, only sampling a small subset of its neighbors will lead to severe information loss. Consequently, the learned embedding of $v_i$ will likely fail to pay sufficient attention to closely related neighbors, and be vulnerable to the noise from semantically irrelevant nodes. In addition, as \cite{hamilton2017inductive} simply assumes there is only one homogeneous type of neighborhood relationship in a graph (i.e., $\mathcal{N}$), Eq.(\ref{eq:mean}) is unable to simultaneously handle information within the heterogeneous neighbors we have defined for passenger demand prediction. In light of this, we propose an attention-based aggregation scheme to discriminatively select important nodes via learned attentive weights, which also extends Eq.(\ref{eq:mean}) to our three heterogeneous neighborhood sets:
\begin{equation}
	\label{eq:aggregator}
	\! \textbf{m}_i^t \!=\! \textbf{W}_s\textbf{v}_i^t \!\oplus\!\! \sum_{j \in \Psi_i^{t}}\!\! \psi_{ij}^{t} \textbf{W}_s\textbf{v}_j^t \!\oplus\!\! \sum_{j \in \Phi_i^t}\!\! \phi_{ij}^{t} \textbf{W}_s\textbf{v}^t_{j} \!\oplus\!\! \sum_{j \in \Theta^i}\!\! \theta_{ij}^{t}\textbf{W}_s\textbf{v}_j^{t},\!
\end{equation}
where $\textbf{W}_s \in \mathbb{R}^{d_e \times d}$ is a shared weight matrix that projects all feature vectors onto the same $d_e$ dimensional embedding space. Notably, the specific $Aggregate(\cdot)$ function we adopt is the weighted sums of node information within $\Psi^{t}_i$, $\Phi^{t}_i$, and $\Theta_i$, respectively. $\psi_{ij}^{t}$, $\phi_{ij}^{t}$ and $\theta_{ij}^{t}$ are the attentive weights between nodes $v_i$ and $v_j$ in the corresponding neighborhoods. 
{In this attention layer, we focus on mining the fine-grained pairwise importance from the neighbor node to the target node. Hence, we employ the self-attention calculation of GAT \cite{velivckovic2017graph} in  Eq.(\ref{eq:attention1}) which is designed for graph node representation learning and would learn a more expressive representation for each node.} To compute the attentive weights, we firstly define a shared attention network denoted by $AttentionNet(\cdot,\cdot)$. $AttentionNet(\cdot,\cdot)$ produces a scalar that quantifies the semantic affinity between nodes $v_i$ and $v_j$ using their features. Take one arbitrary node pair $(v_i, v_j)$ at time $t$ as an example, then $AttentionNet(\textbf{v}_i^t, \textbf{v}_j^t)$ is calculated as follows:
\begin{equation}
	\label{eq:attention1}
	AttentionNet(\textbf{v}_i, \textbf{v}_j) = \mu(\textbf{a}^{\top}(\textbf{W}_a\textbf{v}_i^t \oplus \textbf{W}_a\textbf{v}_j^t)),
\end{equation}
where $\mu$ represents the $LeakyReLU$ function that applies nonlinearity, $\textbf{W}_a \in \mathbb{R}^{d_e \times d}$ is the learnable weight matrix, $\textbf{a} \in \mathbb{R}^{2d_e \times 1}$ is the project weight that maps the concatenated vector to a scalar output. Then, we compute $\psi_{ij}^{t}$, $\phi_{ij}^{t}$ and $\theta_{ij}^{t}$ by applying $softmax$ \cite{mnih2009a} to normalize all the attention scores between $v_i$ and its forward, backward, and geographical neighbors:
\begin{equation}
	\label{eq:self-attention1}
	\begin{split}
		\psi_{ij}^{t} = \frac{\exp(AttentionNet(\textbf{v}_i^t, a_{j}^t\textbf{v}_j^t))}{\sum_{ k \in \Psi_i^{t}}\exp(AttentionNet(\textbf{v}_i^t, a_{k}^t\textbf{v}_k^t))},\\
		\phi_{ij}^{t} = \frac{\exp(AttentionNet(\textbf{v}_i^t, b_{j}^t\textbf{v}_j^t))}{\sum_{k \in \Phi_i^{t}} \exp(AttentionNet(\textbf{v}_i^t, b_{k}^t\textbf{v}_k^t))},\\
		\theta_{ij}^{t} = \frac{\exp(AttentionNet(\textbf{v}_i^t, c_{j}\textbf{v}_j^t))}{\sum_{k \in \Theta_i} \exp(AttentionNet(\textbf{v}_i^t, c_{k}\textbf{v}_k^t))},
	\end{split}
\end{equation}
which enforces $\sum_{j\in\Psi_i^{t}}{\psi_{ij}^{t}} = \sum_{j\in\Phi_i^{t}}{\phi_{ij}^{t}} = \sum_{j\in\Theta_i}{\theta_{ij}^{t}} = 1$ and thus can be viewed as three probability distributions over the corresponding type of neighborhoods. The attention network is shared across the computations for all three neighborhoods. In Eq.(\ref{eq:self-attention1}), it is worth noting that, before being fed into the attention network, every neighbor node of $v_i$ is weighted by a factor $a$, $b$ and $c$ for the forward, backward, and geographical neighborhood, respectively. Next, we explain the rationale of involving these weights in the computation of Eq.(\ref{eq:self-attention1}) and present the details of three pre-weighted functions for generating the weights $a$, $b$ and $c$.

\subsubsection{\textbf{Pre-weighted Functions}}
{The core idea behind our pre-weighted functions is to timely help sense the sparsity of the data and provide additional prior knowledge for the subsequent attention-based information aggregation.} This is achieved by taking advantage of the observed weights on each DDW graph $\textbf{G}_t$ and the geographical relationship $\textbf{R}$. Given the target node $v_i$, for any of its forward, backward or geographical neighbor $v_j$ in $\Psi_i^{t}$, $\Phi_i^{t}$ or $\Theta_i$, we derive three statistics-driven pre-weighted functions to compute the corresponding weight for $v_j$, i.e., $a_{j}^t$, $b_{j}^t$ or $c_{j}$:
\begin{equation}
	\label{eq:pre-weighted1}
	\begin{split}
		& a_{j}^{t} = \frac{g_{ij}}{\sum_{j\in \Psi^{t}_i} g_{ij}+\epsilon},  \,\, g_{ij} \in \textbf{G}_{t},\\
		& b_{j}^{t} = \frac{g_{ji}}{\sum_{j\in \Phi^{t}_i} g_{ji} +\epsilon}, \,\, g_{ji} \in \textbf{G}_{t},\\
		& c_{j} = \frac{\frac{1}{r_{ij}}}{\sum_{j \in \Theta_i} \frac{1}{r_{ij}}}, \,\, r_{ij} \in \textbf{R},
	\end{split}
\end{equation}
where $\epsilon$ is a small additive term in case the denominator is $0$ (i.e., $\Psi^{t}_i= \varnothing$ or $\Phi^{t}_i= \varnothing$ in highly sparse data). As suggested by Eq.(\ref{eq:pre-weighted1}), the weights $a_{j}^{t}$ and $b_{j}^{t}$ reflect $v_j$'s intensity of passenger demands at time $t$. Therefore, the attention weights $\psi_{ij}^{t}$ and $\phi_{ij}^{t}$ obtained in Eq.(\ref{eq:self-attention1}) is not only dependent on the semantic similarity between node features $\textbf{v}_i^t$ and $\textbf{v}_j^t$, but also the real-time popularity of neighbor region node $v_j$ at time $t$. Besides, motivated by Harmonic mean, the geographical weighting factor $c_j$ essentially assigns larger weights to region nodes that are geographically closer to the target node $v_i$. As such, by coupling the pre-weighted functions with our attention-based aggregator, the embedding $\textbf{m}_i^t$ generated with Eq.(\ref{eq:aggregator}) is an expressive blend of its inner properties and the characteristics of three distinct neighborhoods.

\subsection{Temporal Attention Layer}\label{sec:temproal}
So far, we can obtain a set of embeddings $\{\textbf{m}_i^t\}_{i=1}^n \in \mathbb{R}^{4d_e}$ for all regions in each snapshot DDW graph $\textbf{G}_t$. Specifically, for each time slot $t$, we use a feature matrix $\textbf{M}_t= [\textbf{m}_1^t,\textbf{m}_2^t,...,\textbf{m}_n^t]^{\top} \in \mathbb{R}^{n\times 4d_e}$ to vertically stack all $n$ node embeddings at time $t$. Hence, for the DDW graph sequence $\{\textbf{G}_t\}_{t=1}^{T}$, we can obtain $T$ time-varying feature matrices $\{\textbf{M}_t\}_{t=1}^{T}$ to represent the DDW graphs at corresponding time slots. For the current time slot $T$, since $\textbf{M}_T$ only carries the spatial information within the $T$-th DDW graph. To account for the dynamics within our DDW graphs, we develop a temporal attention layer to firstly capture the sequential dependencies among the learned representations in $\{\textbf{M}_t\}_{t=1}^{T}$, and generate a spatiotemporal representation $\textbf{M}_{T}'$ for predicting the passenger demand in the next time slot. Obviously, a straightforward approach is to gather information from $P$ most recent and consecutive DDW graphs, i.e., $\{\textbf{M}_t\}_{t=T-P+1}^T$. However, in real-life scenarios, for the DDW graph at time $T$, only DDW graphs from time slots that are temporally close to $T$ will exploit similar characteristics. In contrast, if there is a relatively big time gap between two DDW graphs, their characteristics will vary significantly, e.g., the traffic flow in the central business district will be much lower during midnight than in the morning. As a result, merely using consecutive time slots can introduce a large amount of noise when learning the spatiotemporal DDW graph representation $\textbf{M}_{T}'$. 
\begin{figure}[th!]
	\centering
	\setlength{\abovecaptionskip}{0.15cm} 
	\includegraphics[width=0.75\textwidth]{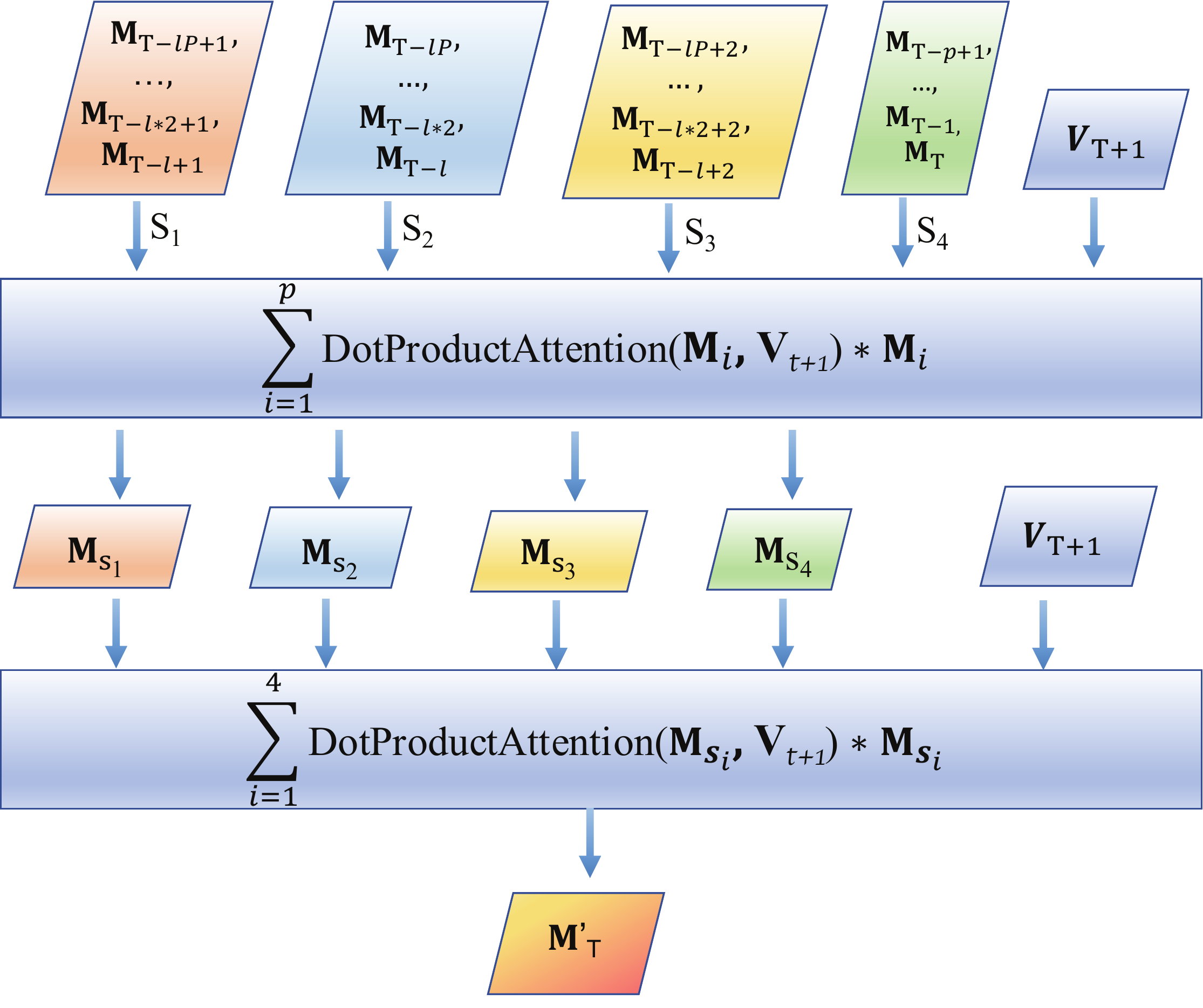}
	\caption{\textbf{The architecture of temporal attention layer.} The temporal attention layer utilizes dot product attention mechanism to capture the dynamic property of DDW via four channels.}
	\label{fig:temporal} 
\end{figure}

On this occasion, as Figure \ref{fig:temporal} shows, we design a multi-channel structure to capture the temporal patterns among different DDW graphs. To enhance the capability of learning useful information from historical DDW graphs, we infuse the periodicity within passenger demands into the temporal attention layer. Specifically, apart from the DDW graph sequence $\{\textbf{M}_t\}_{t=T-P+1}^T$, we derive three periodical sequences to augment the long-term temporal information about the DDW graph at time $T$. First, we collect $P$ historical DDW graphs from the same time slot of each day. For example, if we divide each day into 24 1-hour slots, then we can collect DDW graphs from the same 8:00--8:59 am slot from $P$ consecutive days. Mathematically, we represent such sequence as $\mathcal{S}_1 = \{\textbf{M}_t|t=T-lp+1, p\in[1,P]\}$, where $l$ is the number of time slots in a day ($l=24$ in our case), and $P \leq \lfloor\frac{T}{l}\rfloor$. Similarly, to leverage the close contexts in directly adjacent time slots, we consider two periodical sequences for $T$'s prior and subsequent time slots (i.e., $T-1$ and $T+1$), which results in $\mathcal{S}_2 = \{\textbf{M}_t|t=T-lp, p\in[1,P]\}$ and $\mathcal{S}_3 = \{\textbf{M}_t|t=T-lp+2, p\in[1,P]\}$, respectively. The non-periodical sequence $\mathcal{S}_4 = \{\textbf{M}_t\}_{t=T-P+1}^T$ is also used to capture short-term passenger demand fluctuations.

Correspondingly, we build our temporal attention layer with four channels to attentively aggregate the information within sequences $\mathcal{S}_1$, $\mathcal{S}_2$, $\mathcal{S}_3$, and $\mathcal{S}_4$:  
\begin{equation}
	\label{eq:weightaverage1}
	\begin{split}
		\textbf{M}_{\mathcal{S}} = \sum_{\textbf{M}_t \in \mathcal{S}}\varrho(\frac{\textbf{V}_{T+1}\textbf{W}^Q_{\mathcal{S}} \cdot(\textbf{M}_t\textbf{W}^K_{\mathcal{S}})^\top}{\sqrt{4d_e}})\cdot\textbf{M}_t\textbf{W}^V_{\mathcal{S}}, \mathcal{S} \in \{\mathcal{S}_1, \mathcal{S}_2, \mathcal{S}_3, \mathcal{S}_4\}
	\end{split}
\end{equation}
where $\varrho$ represents the row-wise $softmax$ function. $\textbf{W}^K_{\mathcal{S}}$, $\textbf{W}^V_{\mathcal{S}} \in \mathbb{R}^{4d_e \times 4d_e}$ and $\textbf{W}^Q_{\mathcal{S}} \in \mathbb{R}^{d_v \times 4d_e}$ are query, key and value weight matrices dedicated each channel $\mathcal{S}$. $\textbf{V}_{T+1} = [\textbf{v}_1^{T+1},\textbf{v}_2^{T+1},...,\textbf{v}_n^{T+1}]^{\top} \in \mathbb{R}^{n\times d_v}$ is the feature matrix that stores $n$ nodes' features at time $T+1$. Each feature $\textbf{v} \in \mathbb{R}^{d_v}$ is generated with the same process as in Eq.(\ref{eq:embedding}). Note that $d_v \leq d$ because some features for time slot $T+1$ might be unavailable at the current time $T$ (e.g., the number of trip orders), and are hence excluded. 
{Specifically, we formulate Eq.(\ref{eq:weightaverage1}) with the notion of scaled dot product attention \cite{vaswani2017attention} which is different from the self-attention in Eq.(\ref{eq:attention1}). In this attention layer, we are more concerned on capturing the graph-level association in the time domain. To more efficiently and effectively compute an attentive feature matrix for the current graph, we adopt the scaled dot-product attention in Eq.(\ref{eq:weightaverage1}).} The rationale is that, the row-wise $softmax$ first produces an $n\times n$ attention matrix $\textbf{A}$, where the $i$-th row $\{a_{ij}\}_{j=1}^n \in \textbf{A}$ is a probability distribution indicating the affinity between region $i$ at $T+1$ and each region $j$ at $t\leq T$. Then, by multiplying $\textbf{A}$ with the projected feature matrix $\textbf{M}_t$, we can obtain an updated representation for each time slot $t$ by selectively focusing on regions that are more similar to the contexts of region $v_i$ at $T+1$. By taking the sum of representations for all $P$ time slots, Eq.(\ref{eq:weightaverage1}) generates a temporal representation for each channel, denoted by $\textbf{M}_{\mathcal{S}_1}$, $\textbf{M}_{\mathcal{S}_2}$, $\textbf{M}_{\mathcal{S}_3}$ and $\textbf{M}_{\mathcal{S}_4}$.

After obtaining the channel-wise representations, we merge all information into a unified spatiotemporal representation by sharing another self-attention unit across all four channels:
\begin{equation}
	\label{eq:weightaverage2}
	\begin{split}
		\textbf{M}_{T}' = \sum_{\textbf{M}_{\mathcal{S}} \in \mathcal{S}'}\varrho(\frac{\textbf{V}_{T+1}\textbf{W}_Q' \cdot(\textbf{M}_{\mathcal{S}}\textbf{W}'_K)^\top}{\sqrt{4d_e}})\cdot \textbf{M}_{\mathcal{S}}\textbf{W}'_V, \mathcal{S}'=\{\textbf{M}_{\mathcal{S}_1}, \textbf{M}_{\mathcal{S}_2}, \textbf{M}_{\mathcal{S}_3}, \textbf{M}_{\mathcal{S}_4}\},
	\end{split}
\end{equation}
with weights $\textbf{W}'_K, \textbf{W}'_V \in \mathbb{R}^{4d_e \times 4d_e}$ and $\textbf{W}'_Q \in \mathbb{R}^{d_v \times 4d_e}$ to learn. As such, the resulted matrix $\textbf{M}_{T}'\in \mathbb{R}^{n \times 4d_e}$ is the final representation learned from all DDW graphs in $\{\textbf{G}_t\}_{t=1}^T$. Essentially, $\textbf{M}_{T}'$ now encodes both the spatial and temporal information contained in DDW graphs up to time $T$, which can provide strong predictive signals for estimating the upcoming passenger demands.


\subsection{Transferring Attention Layer}\label{sec:transferring}
With the spatiotemporal representation $\textbf{M}_T'$, we deploy a feed-forward layer to firstly compute an $n$-dimensional vector $\hat{\textbf{d}}_{T+1}=[\hat{d}_1,\hat{d}_2,...,\hat{d}_n] \in\mathbb{R}^{n}$, where each element $\hat{d}_i$ represents the total amount of outbound passenger demands (i.e., the out-degree of node $v_i$) at the next time slot $T+1$:
\begin{equation}
	\hat{\textbf{d}}_{T+1} = \sigma(\textbf{M}_T' \textbf{w} + \textbf{b}),
\end{equation}
with weight $\textbf{w}\in \mathbb{R}^{4d_e}$ and bias $\textbf{b} \in \mathbb{R}^{n}$ to learn. Intuitively, we derive $\hat{\textbf{d}}_{T+1}$ to firstly capture the general trend and intensity of trip demands in each region, then distribute the total demands to different destinations in a fine-grained way. Meanwhile, it can also support traditional  destination-unaware passenger demand prediction tasks \cite{tong2017simpler,wang2019unified}. We denote this task as \textbf{Demand task} in this paper. In the following experiment, we conduct pretraining on this task before the formal training process so that we can obtain the accuracy results more quickly.

To map the total passenger demands $\hat{d}_i$ from region node $v_i$ to all $n$ nodes, we calculate a transferring probability distribution $\{q_{ij}\}_{j=1}^{n}$ to indicate the likelihood of observing a passenger trip from $v_i$ to each destination region $v_j$ at the next time slot $T+1$. Specifically, as the $i$-th row $\textbf{m}_i'\in \textbf{M}_T'$ is a row vector carrying the spatiotemporal representation of node $v_i$, we calculate each probability $q_{ij} \in \{q_{ij}\}_{j=1}^{n}$ via the following attention mechanism:
\begin{equation}
	\label{eq:transfer_attention}
	q_{ij} = \frac{\exp(AttentionNet({\textbf{m}_i'}^{\top}, {\textbf{m}_j'}^{\top}))}{\sum_{i=j}^{n}\exp(AttentionNet({\textbf{m}_i'}^{\top}, {\textbf{m}_j'}^{\top}))},
\end{equation}
where the $AttentionNet(\cdot,\cdot)$ has the same structure as in Eq.(\ref{eq:self-attention1}) but uses a different set of parameters $\textbf{a}'\in\mathbb{R}^{4d_e}$ and $\textbf{W}_a'\in\mathbb{R}^{4d_e\times 4d_e}$. Finally, we can estimate every element in the next DDW graph $\hat{g}_{ij} \in \hat{\textbf{G}}_{T+1}$ (which is denoted as the \textbf{OD task}) :
\begin{equation}
	\hat{g}_{ij} = \hat{d}_{i}q_{ij}, \,\, \hat{g}_{ij}\in \hat{\textbf{G}}_{T+1}, \,\,\hat{d}_{i}\in \hat{\textbf{d}}_{T+1}.
\end{equation}

Note that we only consider the start time of the passenger demand, that is, how many trip requests will generate between two nodes at the time slot $T+1$ no matter whether the trips will be finished in $T+1$. 

\subsection{Optimization Strategy}\label{sec:optimization}
We formulate the overall loss function as follows:
\begin{equation}
	\begin{split}
		&\mathcal{L} = \eta_{d}\mathcal{L}_{d} + \eta_{o}\mathcal{L}_{o},	\\
		&\mathcal{L}_{d} = SmoothL1Loss( \hat{\textbf{d}}_{T+1},  {\textbf{d}}_{T+1}),\\
		&\mathcal{L}_{o} = SmoothL1Loss(\hat{\textbf{G}}_{T+1},  {\textbf{G}}_{T+1}),
	\end{split}
\end{equation}
where $SmoothL1Loss$~\cite{girshick2015fast} is a variant of the mean-squared-error loss which uses a squared term if the absolute element-wise error falls below $1$ and an $L1$ term otherwise; $ \eta_{d}$,  $\eta_{o}$ are two hyperparameters balancing the importance of two tasks. The motivation of defining $\mathcal{L}$ is to push our Gallat model to generate accurate predictions on both the overall demands ($\hat{\textbf{d}}_{T+1}$) and the origin-destination demands ($\hat{\textbf{G}}_{T+1}$). In addition, we conduct pretraining for our model on the Demand task's loss function $\mathcal{L}_{d}$ firstly. Then with the relative accurate prediction of  $\hat{\textbf{d}}_{T+1}$, we train the model for the further prediction of $\hat{\textbf{G}}_{T+1}$ (OD task) based on the pretrained model. All parameters are optimized with the Stochastic Gradient Descent (SGD) method. Specifically, we use Adam~\cite{kingma2014adam}, a variant of SGD to optimize the parameters in our model.

\subsection{Complexity Analysis}\label{sec:computingcomplexity}
In this section, we analyze both the time and space complexity of Gallat.

\subsubsection{\textbf{Time Complexity}}\label{sec:timecomplexity} Putting away the convenient concatenation operation and weighted sum, the major computational cost of Gallat comes from the attention mechanisms used in our spatial, temporal, and transferring layers. For every node feature pair $(\textbf{v}_i, \textbf{v}_j)$, it takes $O(dd_e + 2d_e)$ time to compute a scalar attention score in Eq.(\ref{eq:attention1}). As there are $n^2$ possible combinations for $n$ nodes, the total time complexity is $O(n^2dd_e + 2n^2d_e)$ for the computation in Eq.(\ref{eq:self-attention1}). At the same time, the self-attention modules in the temporal attention layer (i.e., Eq.(\ref{eq:weightaverage1}) and Eq.(\ref{eq:weightaverage2})) consumes $O((P+1)\times(8nd_e^2+nd_ed_v+2n^2d_e))$ time to calculate. Similar to the spatial attention layer, Eq.(\ref{eq:transfer_attention}) has the time complexity of $O(4n^2d_e^2 + n^2d_e)$. As $P$, $d_e$, and $d_v$ are typically small (see Section \ref{sec:evaluation}), the predominant factor in Gallat's time complexity is the total number of regions $n$. Also, as $P$, $d_e$, $d_v$ and $n$ are fixed in our model, the time complexity of Gallat is linearly associated with the scale of the data.

\subsubsection{\textbf{Space Complexity}}\label{sec:spatialcomplexity}
The trainable parameters from the spatial, temporal and transferring attention layers are $\{\textbf{W}_s, \textbf{W}_a, \textbf{a}\}$, $\{\textbf{W}^{K}_{\mathcal{S}}, \textbf{W}^{Q}_{\mathcal{S}}, \textbf{W}^V_{\mathcal{S}}, \textbf{W}'_{K}, \textbf{W}'_{Q}, \textbf{W}'_V\}$, and $\{\textbf{w}, \textbf{b}, \textbf{W}_a', \textbf{a}'\}$, respectively. This results in a total parameter size of $(2dd_e+2d_e) + (160d_e^2 + 20d_ed_v) + (16d_e^2 + 8d_e+ n) = 176d_e^2 + 20d_ed_v + 2dd_e + 10d_e +n$. Hence, the dominating term in the parameter size of Gallat is $176d_e^2$, which has the space complexity of $O(d_e^2)$.

\section{Experiment}\label{sec:evaluation}
In this section, we conduct experiments on real-world datasets to showcase the advantages of Gallat in passenger demand prediction tasks. In particular, we aim to
answer the following research questions via the experiments:
\begin{itemize}
	\item[\textbf{RQ1:}] How effectively does Gallat work on passenger demand prediction tasks?
	\item[\textbf{RQ2:}] How does Gallat benefit from each component of the proposed model structure?
	\item[\textbf{RQ3:}] How do the major hyperparameters affect the prediction performance of Gallat?
	\item[\textbf{RQ4:}] What about the scalability of  Gallat with the number of time slots/grids increasing?
	\item[\textbf{RQ5:}] What's the effectiveness of the pretraining process on Demand task?
	\item[\textbf{RQ6:}] Can Gallat learn useful mobility patterns from real data?
\end{itemize}
\begin{table}[h!]
	\setlength{\abovecaptionskip}{0.1cm}
	\centering
	\caption{\textbf{A summary of datasets}}
	\label{table:datasets}
	\begin{tabular}{c c c}
		\toprule 
		dataset               &Beijng              &Shanghai                           \\ \hline 
		time span           & 4 months       & 4 months\\
		total area & $53 \times 52$ km$^2$ & $57 \times 74 $ km$^2$\\
		grid granularity &$2.65 \times 2.6$ km$^2$ &$2.86 \times 3.70$ km$^2$ \\
		time slot granularity & 1 hour & 1 hour\\
		\bottomrule
	\end{tabular}	
\end{table}
\vspace{-3ex}
\subsection{Datasets}
We conduct experiments on two real-world datasets generated by Didi, which are both desensitized. Some similar datasets are publicly available\footnote{https://outreach.didichuxing.com/research/opendata/}. Table \ref{table:datasets} summarizes the characteristics of two datasets. The first dataset is collected in Beijing covering the area within 6th Ring Road. The second dataset covers the urban area of Shanghai. Both datasets are collected from June to September in 2019.
{We divide Beijing and Shanghai into 400 grids based on the granularities shown in Table \ref{table:datasets} as the average time for a car to travel such distance is 5 minutes, which is a reasonable waiting time for passengers~\cite{wei2016zest}. The DDW graphs on both datasets are constructed with 1-hour granularity. As the prediction results for passenger demands are mainly used as a reference for vehicle dispatching, so one-hour granularity can provide enough time for having the dispatching strategy operated in advance.}

\subsection{Baselines}
To evaluate the performance of our model Gallat on demand prediction, we compare with the following baseline methods.

\begin{itemize}
	\item{\textbf{HA:}} We adopt History Average to passenger demand prediction by calculating the mean of historical data at the same time slot of days and the same day of weeks. 
	
	\item{\textbf{LSTNet:}} LSTNet~\cite{lai2018modeling} is a state-of-the-art time series prediction model, which combines both LSTM and CNN for spatiotemporal feature modelling.
	
	\item{\textbf{GCRN:}} The recently proposed GCRN~\cite{seo2018structured} combines GCN with RNN to jointly identify spatial correlations and dynamic patterns. 
	
	\item{\textbf{GEML:}} GEML~\cite{wang2019origin} employs graph embedding in the spatial perspective and a LSTM-based multi-task Learning architecture in the temporal perspective to predict the passenger demands from one region to another. 
\end{itemize}

\subsection{Experimental Settings}\label{sec:expsetting}
We conduct experiments on passenger demand prediction settings, i.e., origin-destination demand prediction (denoted by ``OD'') and origin-only demand prediction (denoted by ``Demand''), which corresponds to our model output of $\hat{\textbf{G}}_{T+1}$ and $\hat{\textbf{d}}_{T+1}$, respectively. We measure the prediction accuracy with Mean Absolute Percentage Error (MAPE) and Mean Absolute Error (MAE) which have been widely used to evaluate the model performance in regression tasks:
\begin{equation}
	\label{eq:mape}
	MAPE = \frac{1}{n} \sum_{i=1}^{n}|\frac{\hat{y_i} - y_i}{y_i + 1}|,  MAE = \frac{1}{n} \sum_{i=1}^{n}|\hat{y_i} - y_i|,
\end{equation}
where $n$ is the total number of instances, $\hat{y_i}$ represents the predicted result and $y_i$ represents the ground truth. 
{In real-life applications, ride-hailing platforms concern more about the areas with more passenger demands. Regions having almost no passenger demand are less profitable and are paid less attention to. As a result, they only calculate the metrics for those records whose values are above some thresholds. In our experiments, we select three thresholds i.e., 0, 3, and 5 to calculate the metrics results which are termed as MAPE-0, MAPE-3, MAPE-5 and MAE-0, MAE-3, MAE-5, respectively. These are also used to evaluate the demand prediction on Didi platform now.}

\begin{table*}[t]
	\centering
	\setlength{\abovecaptionskip}{0.1cm}
	\setlength\tabcolsep{3.1pt}
	\caption{Results of Different Methods-Beijing}
	\begin{tabular}{|c|c||c|c|c|c|c|c|}
		\hline
		\multirow{2}*{Task}&\multirow{2}*{Method}&
		\multicolumn{6}{c|}{Beijing}\\ \cline{3-8}
		&&MAPE-0&MAPE-3&MAPE-5&MAE-0&MAE-3&MAE-5\\
		\hline
		\multirow{5}*{OD}&HA&2.7454&3.0059&3.1332&13.3953&39.7657&54.0634\\
		&LSTNet& 2.9443&4.3750&6.7874&14.8374&42.4918&95.2971\\
		&GCRN& 0.8347&0.9549&0.9693&5.0278&15.7346&23.3819\\
		&GEML& 0.8736&0.9244&0.9832&5.4396&12.6831&24.9918\\
		&Gallat& \textbf{0.7283}&\textbf{0.8465}&\textbf{0.8896}&\textbf{2.6781}&\textbf{3.4139}&\textbf{6.2347}\\
		\hline
		\multirow{5}*{Demand}&HA& 4.1315&3.7744&3.7077&512.7970&564.1135&593.6073\\
		&LSTNet& 12.5007&7.0283&5.3786&1329.3238&1589.1176&1608.4531\\
		&GCRN& 10.1060&4.1456&3.1136&127.3906&134.4935&148.2324\\
		&GEML& 0.8710&0.7671&0.7232&24.7852&29.3469&32.5521\\
		&Gallat& \textbf{0.6902}&\textbf{0.3904}&\textbf{0.3613}&\textbf{17.8332}&\textbf{20.0048}&\textbf{23.9004}\\
		\hline
	\end{tabular}
	\label{table:overall1}
\end{table*}
\vspace{4ex}
\begin{table*}
	\centering
	\setlength{\abovecaptionskip}{0.1cm}
	\setlength\tabcolsep{3.1pt}
	\caption{Results of Different Methods-Shanghai}
	\begin{tabular}{|c|c||c|c|c|c|c|c|}
		\hline
		\multirow{2}*{Task}&\multirow{2}*{Method}&
		\multicolumn{6}{c|}{Shanghai}\\ \cline{3-8}
		&&MAPE-0&MAPE-3&MAPE-5&MAE-0&MAE-3&MAE-5\\ \hline
		\multirow{5}*{OD}&HA&2.6416&2.8522&2.9670&18.5351&53.3798&71.8696\\
		&LSTNet&3.8422&4.6471&7.7677&28.1293&87.2351&112.3874\\
		&GCRN&0.8714&0.9678&0.9783&4.1634&12.2219&16.4002\\
		&GEML&0.8922&0.9210&0.9792&6.0167&12.3469&15.9975\\
		&Gallat&\textbf{0.6813}&\textbf{0.8752}&\textbf{0.9143}&\textbf{3.6138}&\textbf{5.3497}&\textbf{8.6622}\\
		\hline
		\multirow{5}*{Demand}&HA&3.7065&3.4005&3.3711&414.8688&489.3267&522.7588\\
		&LSTNet&15.9938&5.0623&4.6083&625.9973&688.1231&745.8227\\
		&GCRN&8.1710&3.3146&2.4972&143.7841&149.9170&153.6679\\
		&GEML&1.0220&0.7297&0.6832&37.4469&41.0042&49.6711\\
		&Gallat&\textbf{0.6899}&\textbf{0.4109}&\textbf{0.3815}&\textbf{21.3526}&\textbf{24.7359}&\textbf{29.1875}\\
		\hline
	\end{tabular}
	\label{table:overall2}
\end{table*}
\vspace{-3ex}
In experiments, we leave out the last two weeks of each dataset as the test set and the rest is training set. The last 10\% of the training set is used for validation. We implement Gallat with Pytorch 1.5.0 on Python 3.8. When training Gallat, we use the Demand task to do pre-training first, which can relieve the data sparsity problem and speed the training process. The default values of batch size, epoches, embedding dimension $d_e$, historical time slots $P$ , and loss weights $(\eta_d, \eta_o)$ are set as $20$, $200$, $16$, $7$ and $(0.8, 0.2)$. 

\subsection{Effectiveness Analysis (RQ1)}
Table~\ref{table:overall1} and Table~\ref{table:overall2} shows the results of state-of-the-art methods and Gallat under MAPE and MAE above thresholds on the test set. For the comparison with other models, we make the following observations from these two tables:
\begin{itemize}
	\item It clearly shows that the results of LSTNet and GCRN on Demand task are even worse than HA while they are doing better on OD task. GEML and Gallat show stable performance on both tasks. It may be because LSTNet and GCRN just focus on single task in their model structure and GEML and Gallat both involve the two tasks in their design.
	\item Methods tailored for graph-structured data (GCRN, GEML, Gallat) achieve better overall performance on OD task. It might prove that it's a better choice to model the passenger mobility prediction as a graph-based problem so that the complicated interactions between nodes can be fully captured. And in all of them, Gallat takes a more sufficient consideration on the graph representation, which tends to be the main reason that Gallat outperforms other spatiotemporal models.
	\item On Demand task, as the demand threshold increases the MAPE on Demand task is decreasing while MAE keeps increasing. One possible reason is that, the scale of the prediction targets in Demand task is larger than that in the OD task. Then, when we enlarge the threshold, the ground truth, i.e., the dominator of MAPE in Eq.(\ref{eq:mape}) is increasing more significantly than the absolute error in the numerator. Furthermore, Gallat's advantage against baselines is larger on MAPE-5, which demonstrates that our model is highly accurate in prediction the passenger flow for popular regions.
\end{itemize}
\begin{table*}[t!]
	\centering
	\setlength{\abovecaptionskip}{0.1cm}
	\setlength\tabcolsep{3.1pt}
	\caption{Results of Different Variants-Beijing}
	\begin{tabular}{|c|c||c|c|c|c|c|c|}
		\hline
		\multirow{2}*{Task}&\multirow{2}*{Method}&
		\multicolumn{6}{c|}{Beijing}\\ \cline{3-8}
		&&MAPE-0&MAPE-3&MAPE-5&MAE-0&MAE-3&MAE-5\\
		\hline
		\multirow{6}*{OD}&Gallat-S1&0.7728&0.9082&0.9347&3.8855&11.6697&15.7926\\
		&Gallat-S2&0.7497&0.8689&0.9050&3.4523&10.2289&13.8688\\
		&Gallat-S3&0.9534 &0.8689&0.9049&3.5206&10.4229&14.1817\\
		&Gallat-S4&0.7421 &0.8846&0.9135&3.4258&10.2555&13.8887\\
		&Gallat-S5&0.7598 &\textbf{0.7712}&0.9462&3.3208&8.7657&11.6541\\
		&Gallat& \textbf{0.7283}&0.8465&\textbf{0.8896}&\textbf{2.6781}&\textbf{3.4139}&\textbf{6.2347}\\
		\hline
		\multirow{6}*{Demand}&Gallat-S1&2.2931&1.1786&0.9723&52.5624&55.6956&57.5813\\
		&Gallat-S2&0.7637&0.4291&0.3641&25.0391&26.4881&27.4193\\
		&Gallat-S3&1.0001 &0.5030&0.4109&26.5299&27.7005&28.4539\\
		&Gallat-S4&1.0748 &0.4858&0.3925&24.7973&25.8946&26.6563\\
		&Gallat-S5&0.9714 &0.4730&0.3810&24.3334&25.4282&26.1744\\
		&Gallat& \textbf{0.6902}&\textbf{0.3904}&\textbf{0.3613}&\textbf{17.8332}&\textbf{20.0048}&\textbf{23.9004}\\
		\hline
	\end{tabular} 
	\label{table:variants1}
\end{table*}
\vspace{4ex}
\begin{table*}
	\centering
	\setlength{\abovecaptionskip}{0.1cm}
	\setlength\tabcolsep{3.1pt}
	\caption{Results of Different Variants-Shanghai}
	\begin{tabular}{|c|c|c|c|c|c|c|c|}
		\hline
		\multirow{2}*{Task}&\multirow{2}*{Method}&\multicolumn{6}{c|}{Shanghai}\\ \cline{3-8}
		&&MAPE-0&MAPE-3&MAPE-5&MAE-0&MAE-3&MAE-5 \\
		\hline
		\multirow{6}*{OD}&Gallat-S1&0.8635&0.8945&0.9194&5.6163&16.0487&21.5877\\
		&Gallat-S2&0.8458&0.8843&0.9093&5.1938&14.6175&19.5689\\
		&Gallat-S3&0.8673&0.9164&1.1901&5.3672&15.0821&20.1670\\
		&Gallat-S4&0.8450&0.8841&0.9091&5.1041&14.3365&19.2251\\
		&Gallat-S5&0.8338&0.8780&\textbf{0.9056}&5.1784&12.8859&16.5776\\
		&Gallat&\textbf{0.6813}&\textbf{0.8752}&0.9143&\textbf{3.6138}&\textbf{5.3497}&\textbf{8.6622}\\
		\hline
		\multirow{6}*{Demand}&Gallat-S1&4.8831&1.5453&1.1307&64.4367&70.1194&72.9258\\
		&Gallat-S2&1.0183&0.4603&0.3899&27.3659&30.2024&31.4659\\
		&Gallat-S3&1.3682&0.5138&0.4076&27.7591&31.3426&32.9075\\
		&Gallat-S4&1.0572&0.4402&0.3614&23.3665&26.2928&27.6677\\
		&Gallat-S5&0.9371&0.4248&\textbf{0.3605}&23.3078&26.0122&27.2486\\
		&Gallat&\textbf{0.6899}&\textbf{0.4109}&0.3815&\textbf{21.3526}&\textbf{24.7359}&\textbf{29.1875}\\
		\hline
	\end{tabular} 
	\label{table:variants2}
\end{table*}
\subsection{Component Analysis (RQ2)}
To validate the performance gain from each component of our model, we conduct an ablation study in which we change one component from Gallat each time to form a variant model. We implement the following variants of Gallat:
\begin{itemize}
	\item{\textbf{Gallat-S1:}} We use the existing method Graph Attention Networks (GATs) \cite{velivckovic2017graph} to replace the spatial attention layer, which doesn't distinguish the forward and backward neighbors and ignores the geographical neighbors.
	
	\item{\textbf{Gallat-S2:}} We treat forward and backward neighbors as semantic neighbors like \cite{wang2019origin} in the spatial attention layer.
	
	\item{\textbf{Gallat-S3:}} We replace the attention-based aggregator in the spatial attention layer with the default mean aggregator as used in \cite{hamilton2017inductive}.
	
	\item{\textbf{Gallat-S4:}} We use a mean aggregator to replace the dot product attention in the temporal attention layer.
	
	\item{\textbf{Gallat-S5:}} We replace the transferring attention layer with a simple dense layer.
\end{itemize}
By comparing with the different variants of Gallat, a few observations can be obtained from Table~\ref{table:variants1} and Table~\ref{table:variants2}:


\begin{itemize} 
	\item It is obvious that Gallat-S1 has the worst overall performance especially on the Demand task, which may indicates that the whole design of our spatial attention layer play an important role in the overall performance of our model. And then the second worst is Gallat-S3, which means our attention-based aggregator in the spatial attention layer leads to significantly better effectiveness than the simple mean aggregator. 
	\item The overall performance of Gallat-S2 is the better than Gallat-S3, especially on Shanghai dataset, indicating that the separation of forward and backward neighbor aggregation is necessary for providing more contexts for the prediction. 
	\item The performance of Gallat-S4 is as inferior as Gallat-S2. Hence, we can tell that the attention mechanisms in both the temporal attention layer and transferring attention layer can help the Gallat selectively learn useful patterns for passenger demand prediction.
	\item 
	{Gallat-S5 shows $3$ good results out of $24$ ones which may indicate the traditional dense layer has some advantages in learning transferring probabilities. However, Gallat-S5 is very unstable compared with other methods.}
\end{itemize}

\subsection{Hyperparameter Sensitivity Analysis (RQ3)}
In this section, we discuss three important hyperparameters, that is, weights of task losses, i.e.$\eta_{d}$ and $\eta_{o}$, and the number of historical time slots $P$ in each channel. As hyperparameters are closely related to the performance of the model, we conduct experiments by varying their settings and record the new prediction results achieved. In what follows, we discuss the impact of these hyperparameters.
\begin{figure*}[tb!]
	\centering
	\setlength{\abovecaptionskip}{0.15cm}
	\subfigure{
		\label{fig:eva1:sub1}
		\includegraphics[width=0.23\textwidth]{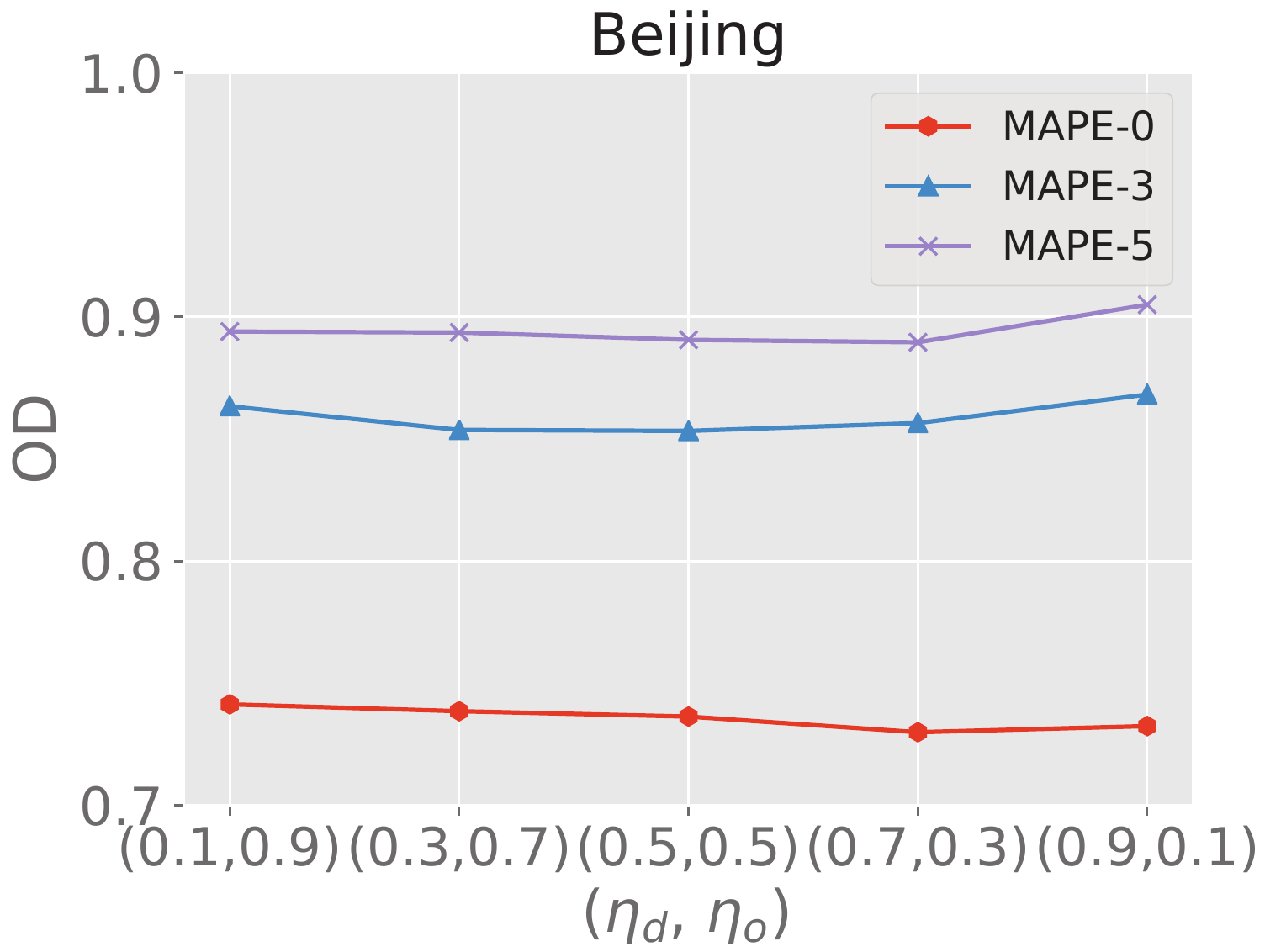}}
	\vspace{-2ex}
	\subfigure{
		\label{fig:eva1:sub2}
		\includegraphics[width=0.23\textwidth]{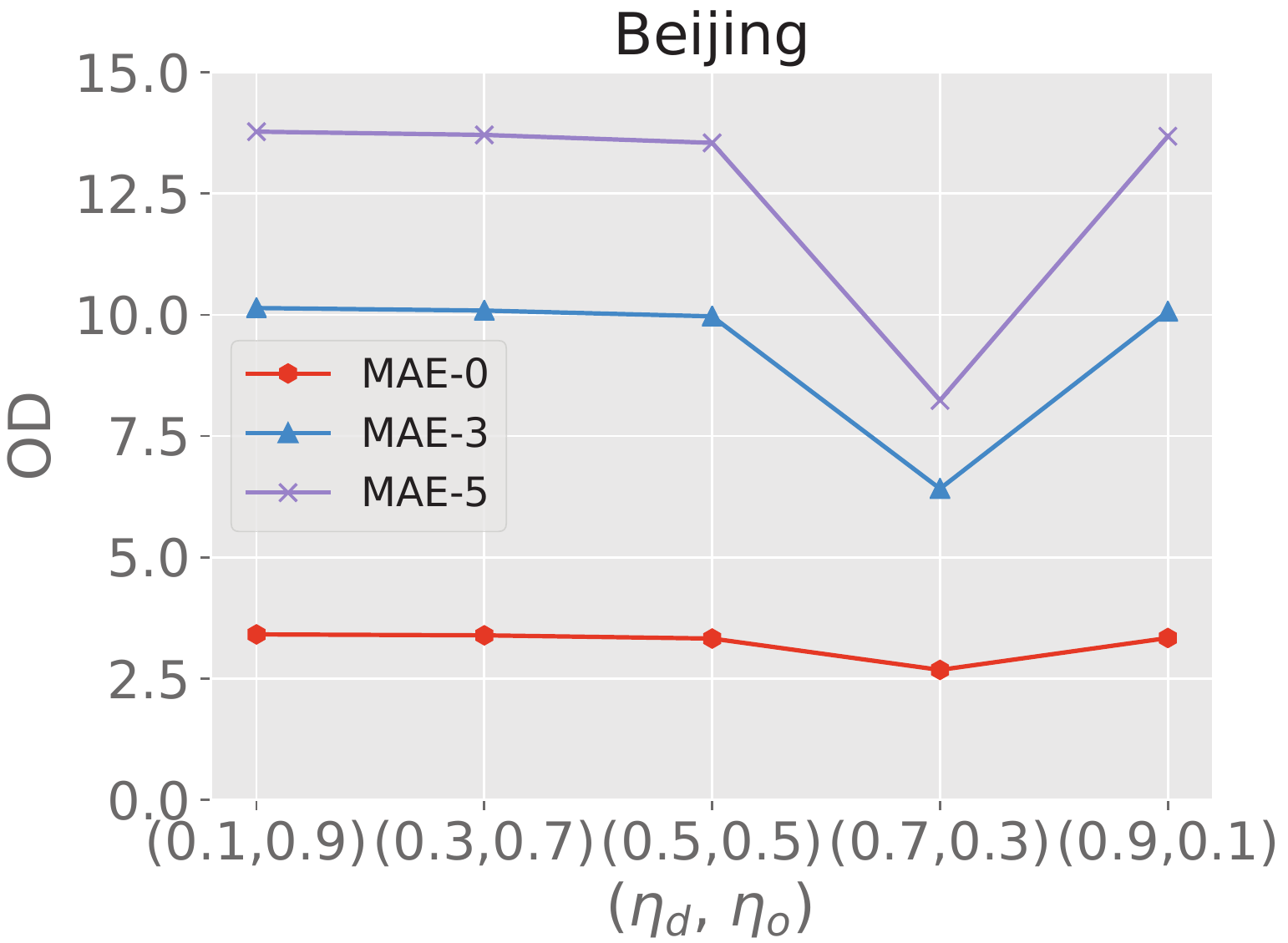}}
	\subfigure{
		\label{fig:eva1:sub3}
		\includegraphics[width=0.23\textwidth]{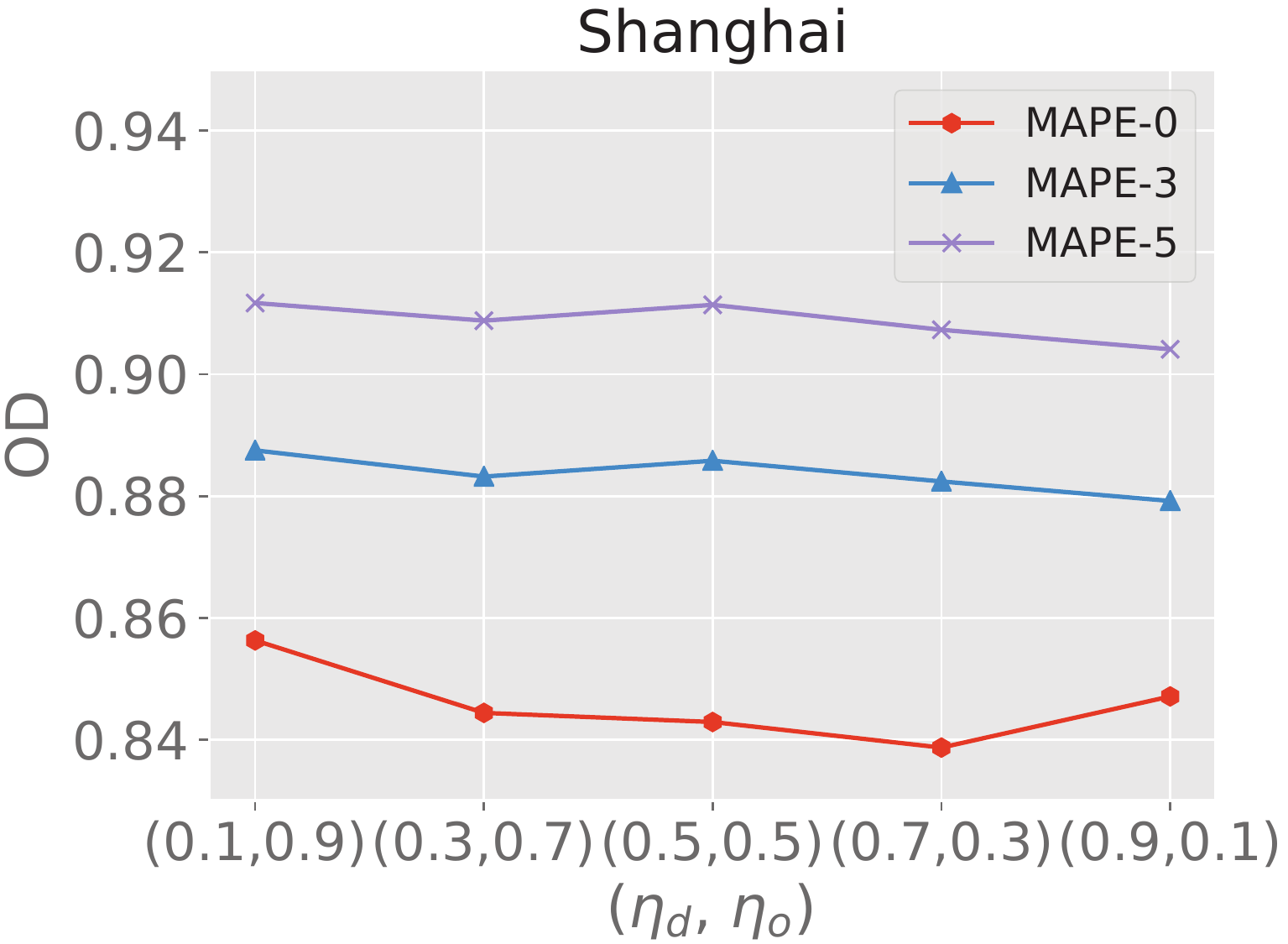}}	
	\subfigure{
		\label{fig:eva1:sub4}
		\includegraphics[width=0.23\textwidth]{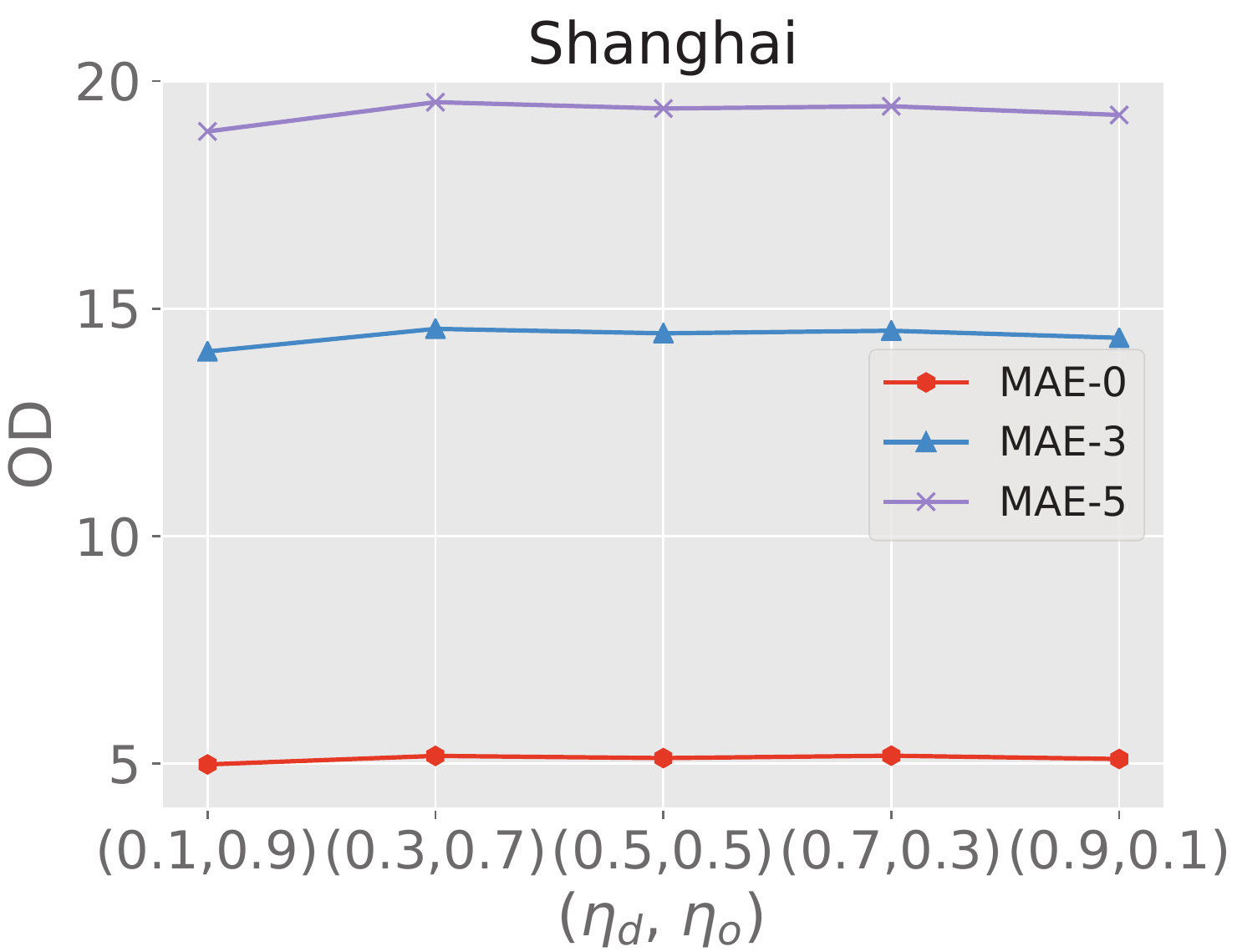}}
	\subfigure{
		\label{fig:eva1:sub5}
		\includegraphics[width=0.23\textwidth]{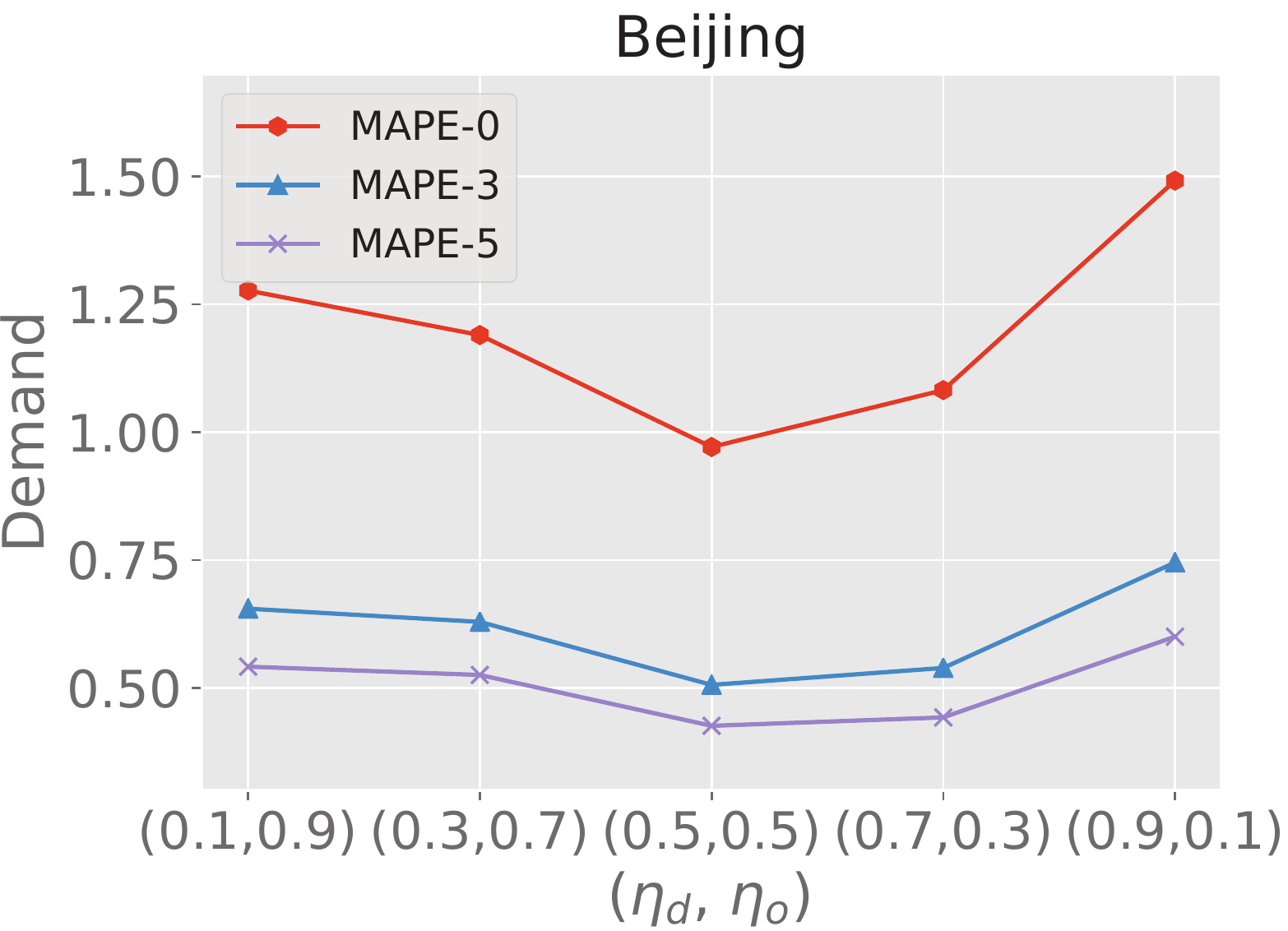}}
	\subfigure{
		\label{fig:eva1:sub6}
		\includegraphics[width=0.23\textwidth]{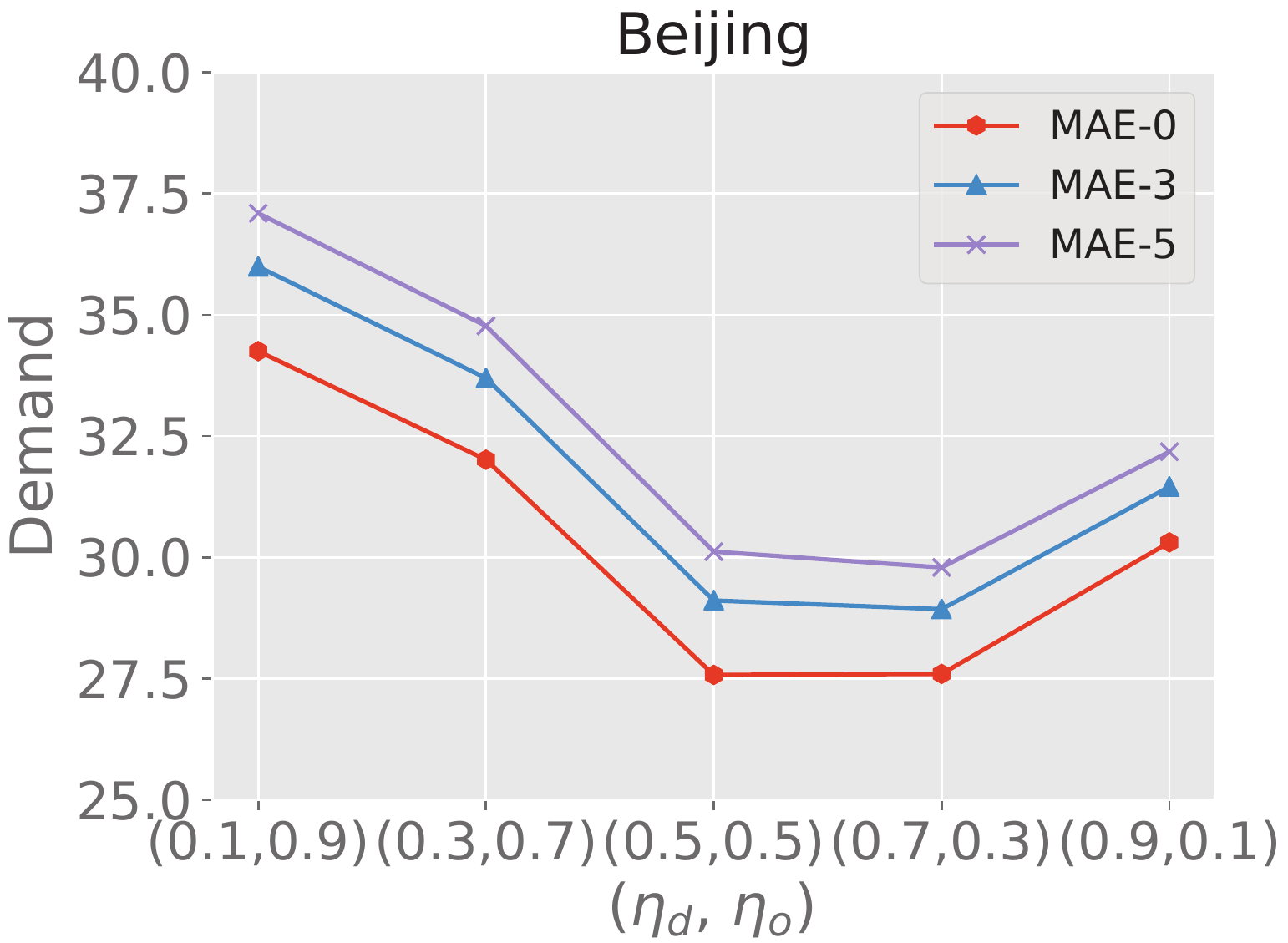}}
	\subfigure{
		\label{fig:eva1:sub7}
		\includegraphics[width=0.23\textwidth]{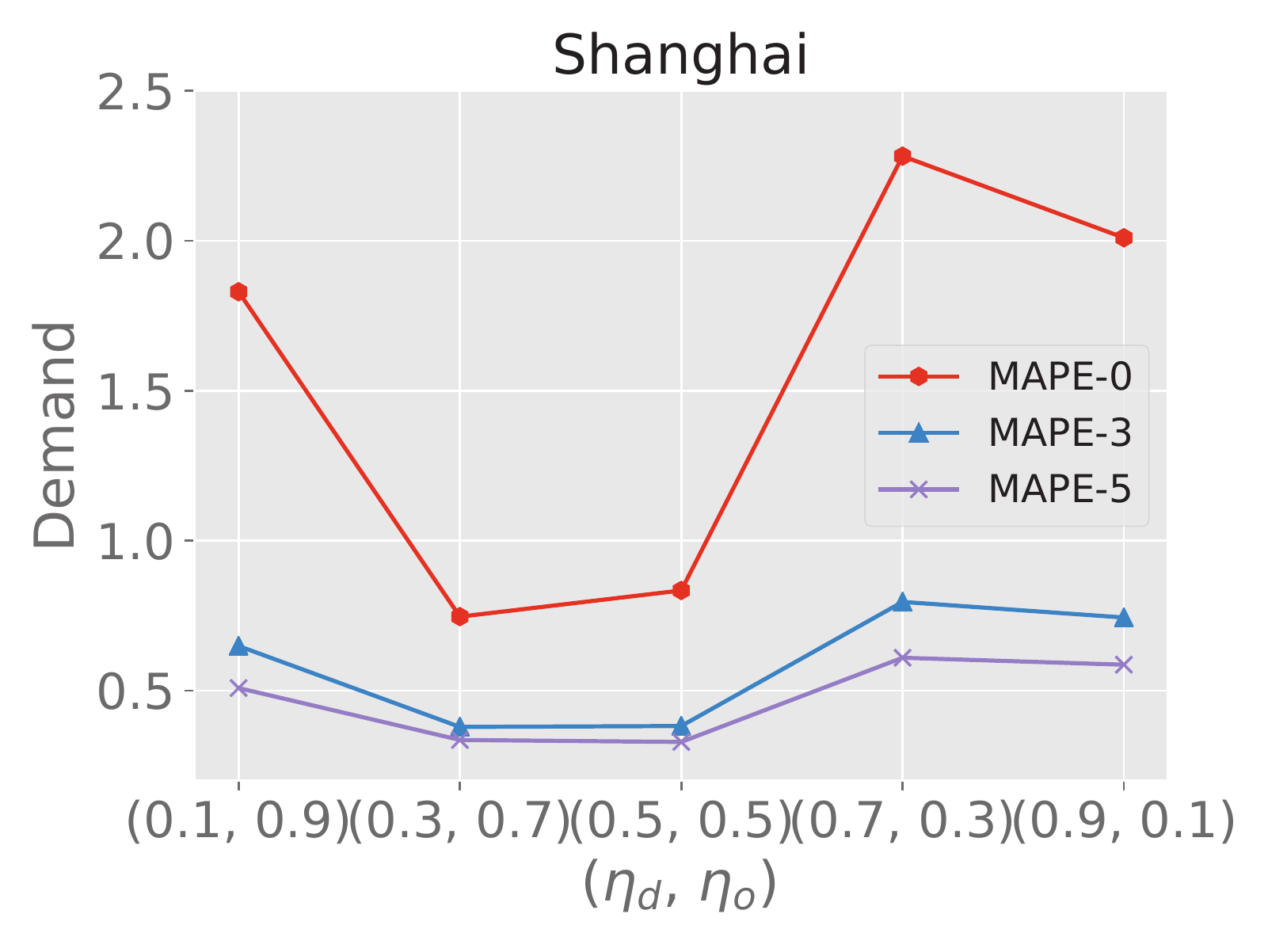}}	
	\subfigure{
		\label{fig:eva1:sub8}
		\includegraphics[width=0.23\textwidth]{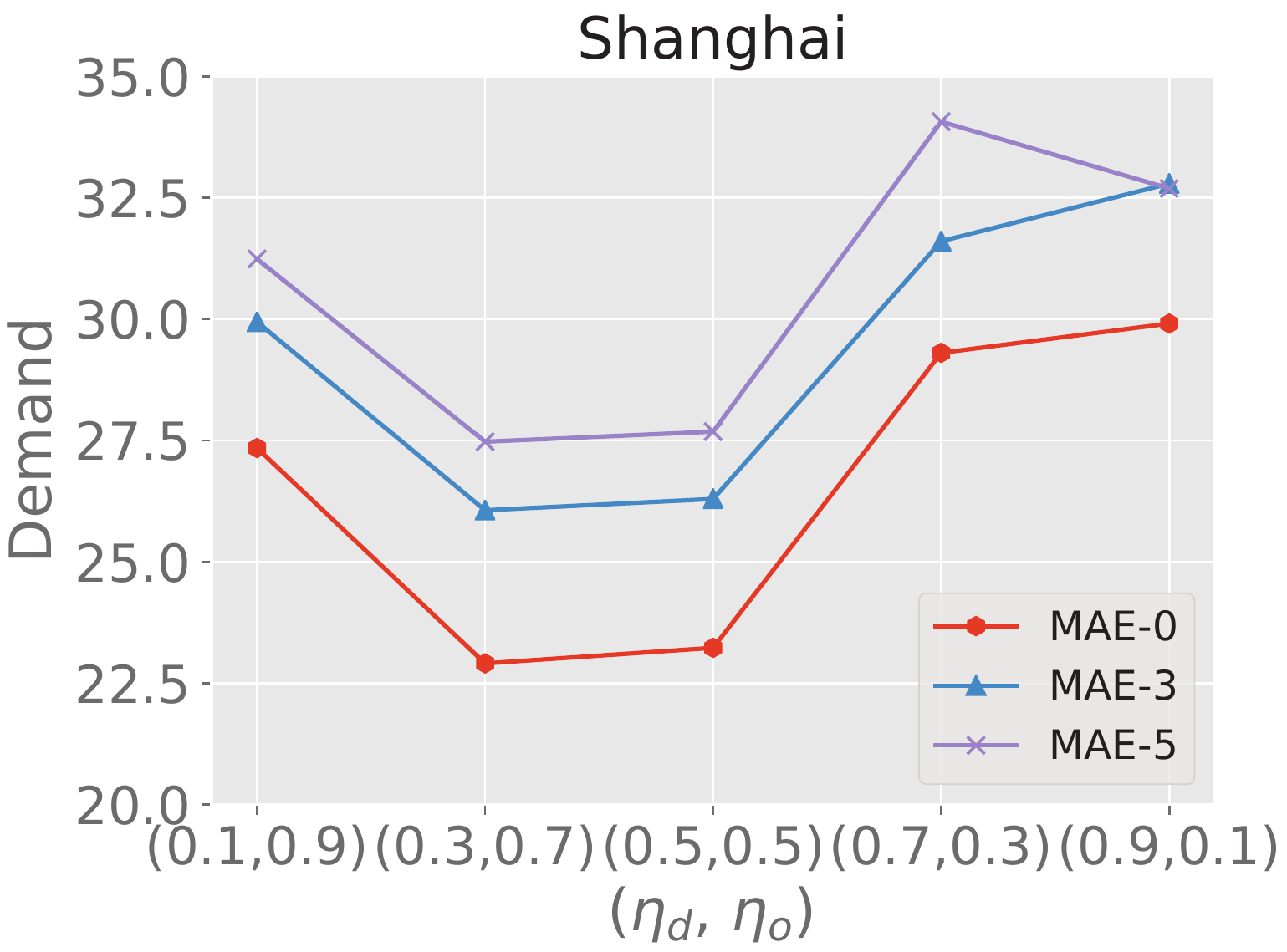}}
	\vspace{-2ex}
	\caption{\textbf{Performance on different weights of task losses}}
	\label{fig:eva1} 
	\vspace{-2ex}
\end{figure*}

\begin{figure*}[tb!]
	\centering
	\setlength{\abovecaptionskip}{0.15cm}
	\subfigure{
		\label{fig:eva2:sub1}
		\includegraphics[width=0.23\textwidth]{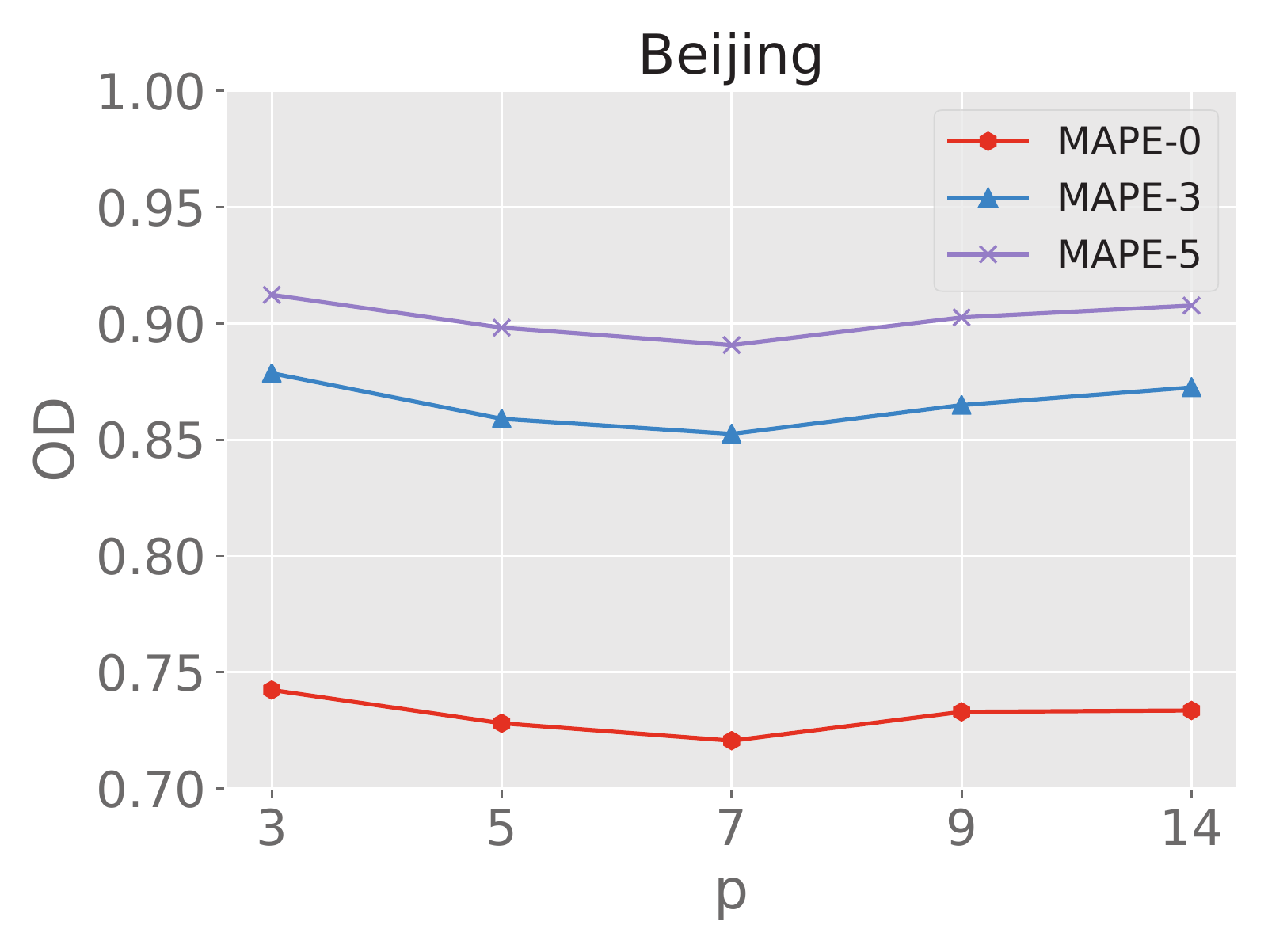}}
	\vspace{-2ex}
	\subfigure{
		\label{fig:eva2:sub2}
		\includegraphics[width=0.23\textwidth]{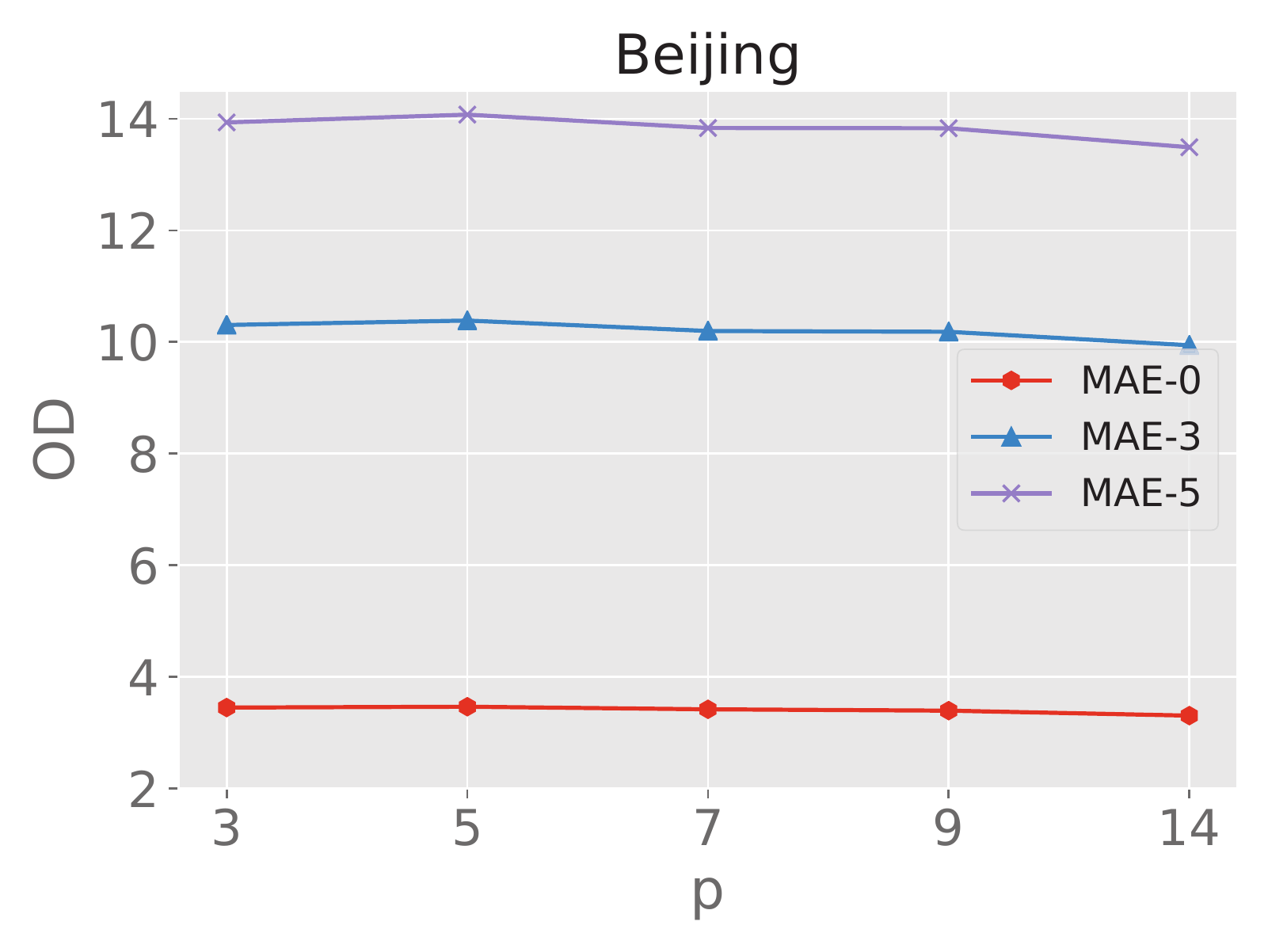}}
	\subfigure{
		\label{fig:eva2:sub3}
		\includegraphics[width=0.23\textwidth]{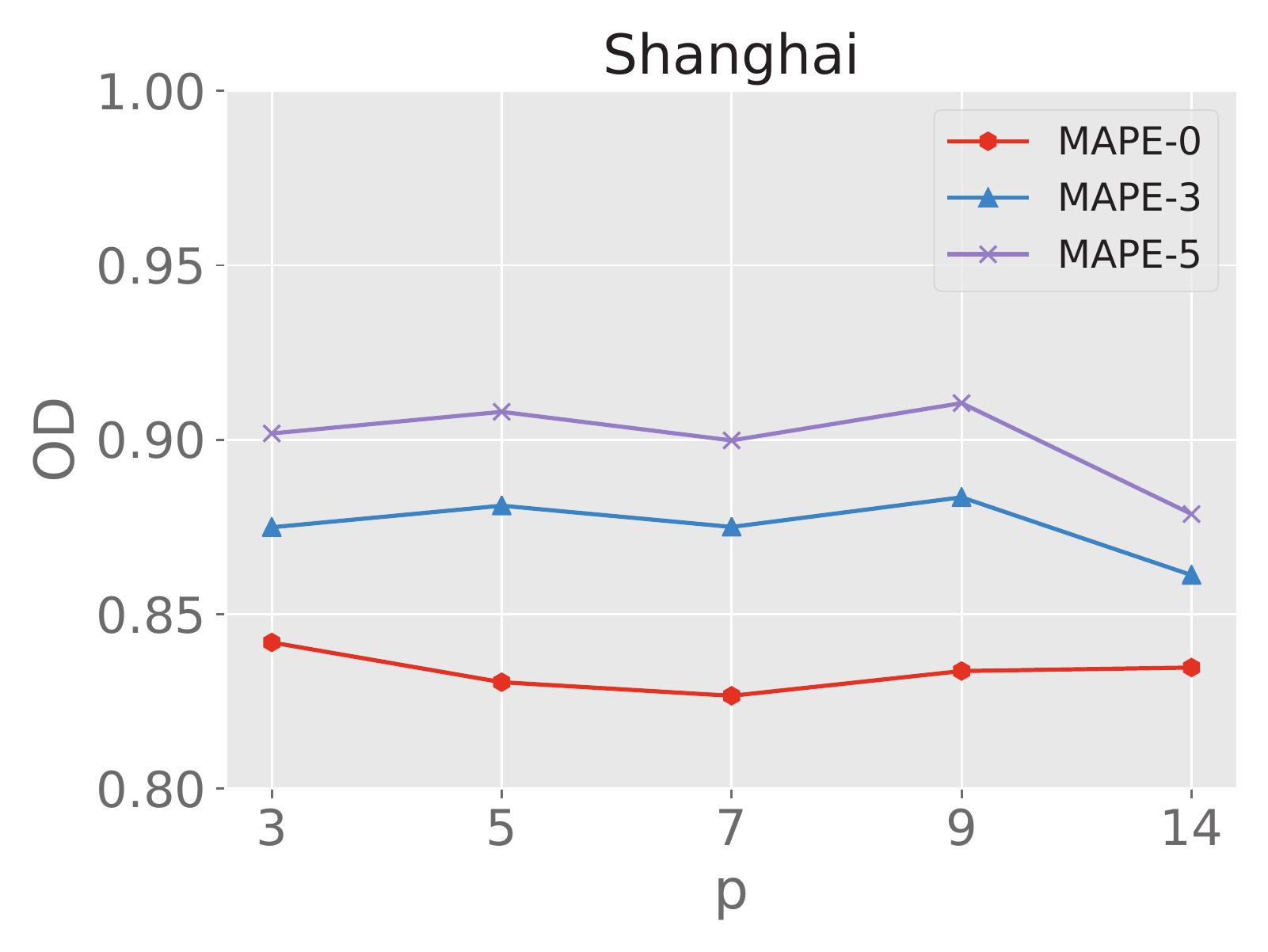}}	
	\subfigure{
		\label{fig:eva2:sub4}
		\includegraphics[width=0.23\textwidth]{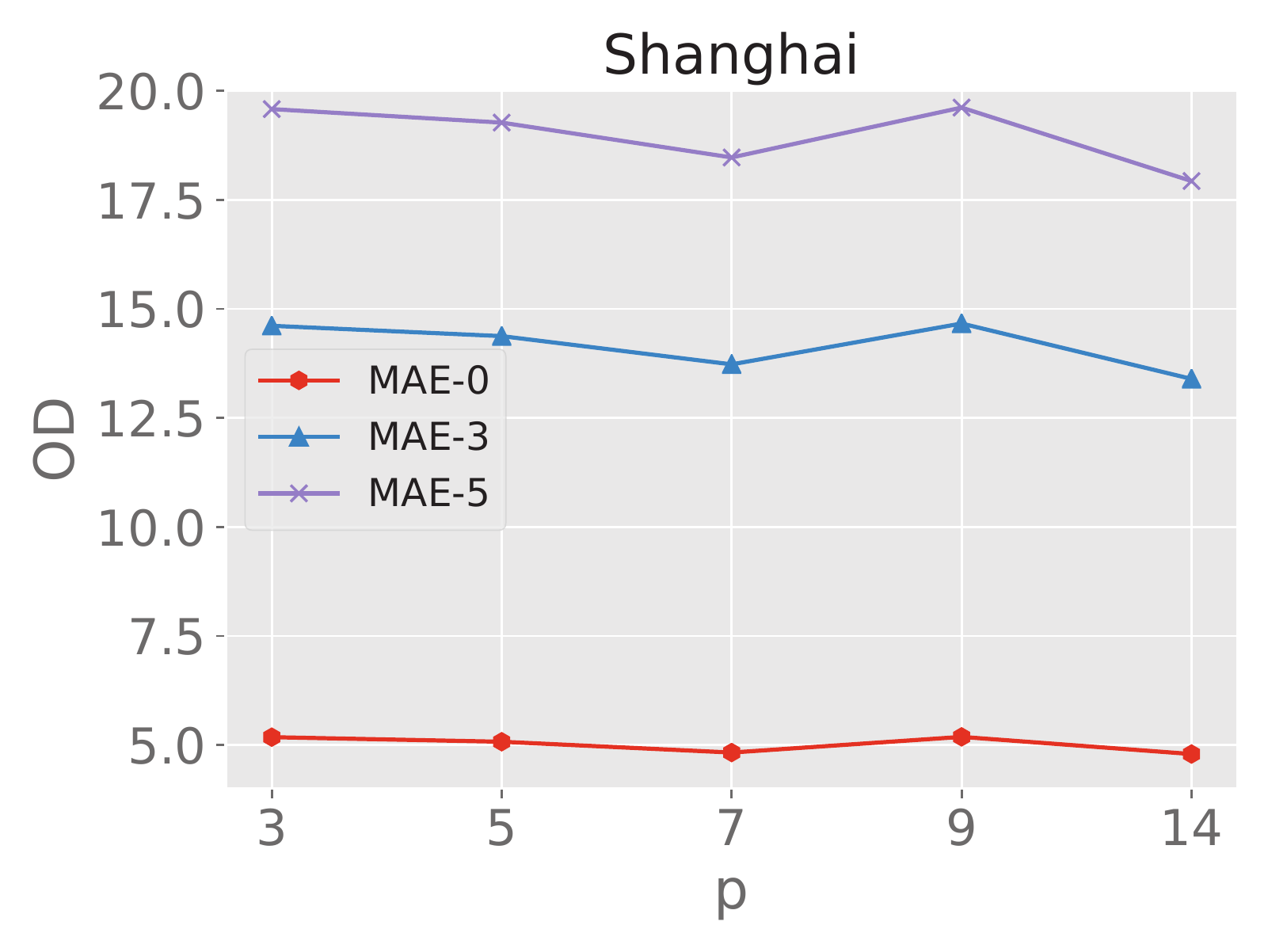}}
	\subfigure{
		\label{fig:eva2:sub5}
		\includegraphics[width=0.23\textwidth]{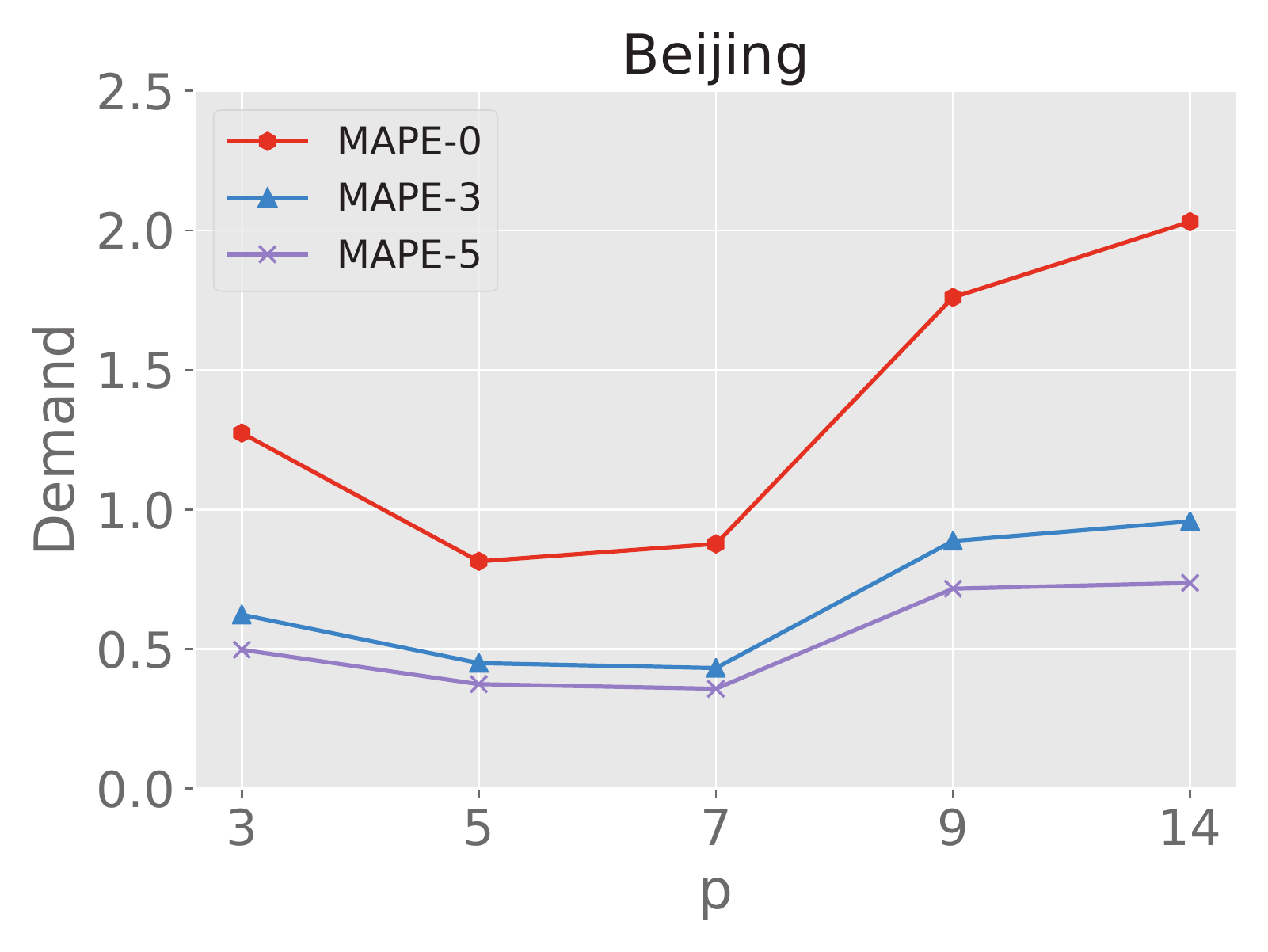}}
	\subfigure{
		\label{fig:eva2:sub6}
		\includegraphics[width=0.23\textwidth]{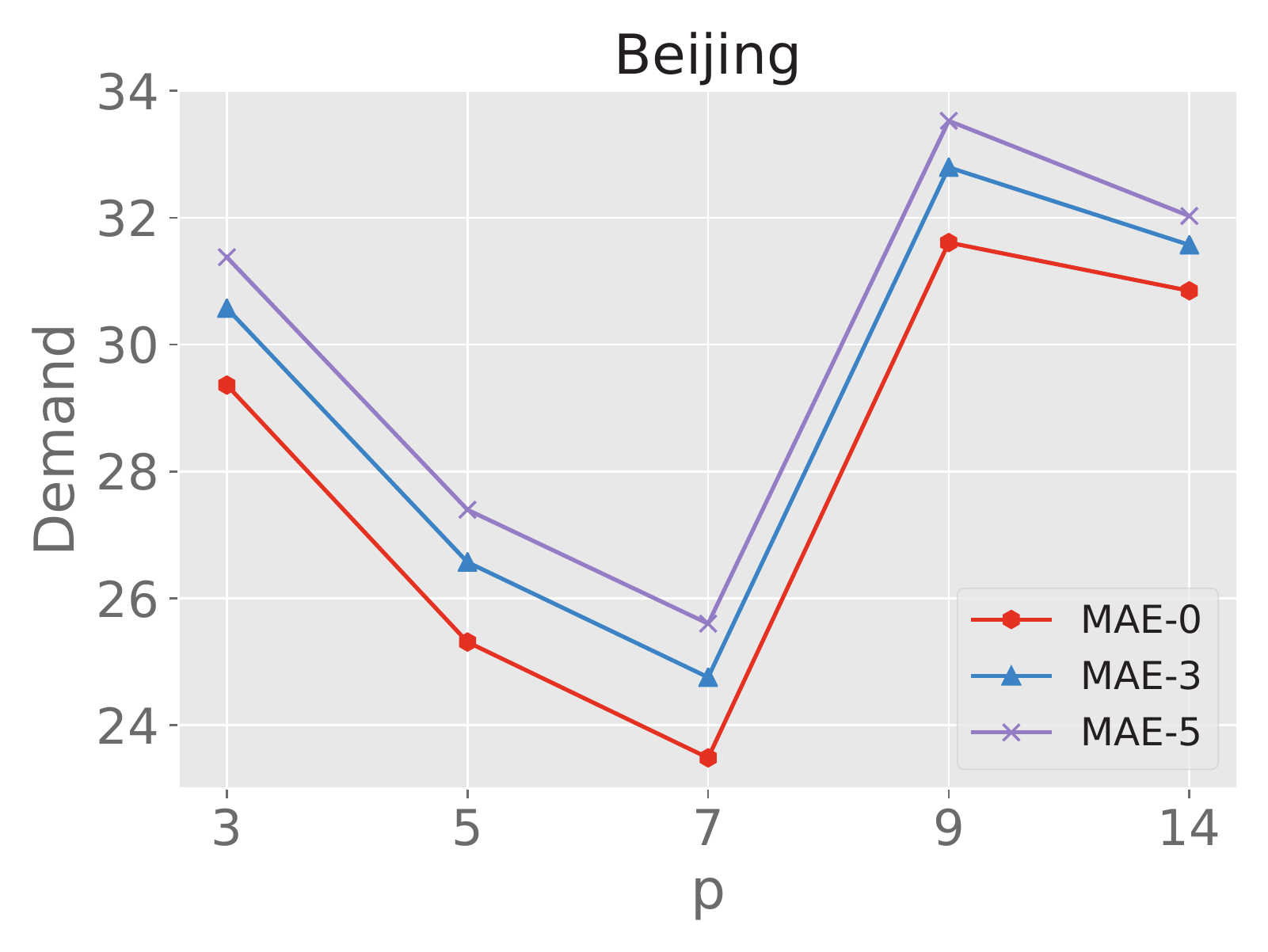}}
	\subfigure{
		\label{fig:eva2:sub7}
		\includegraphics[width=0.23\textwidth]{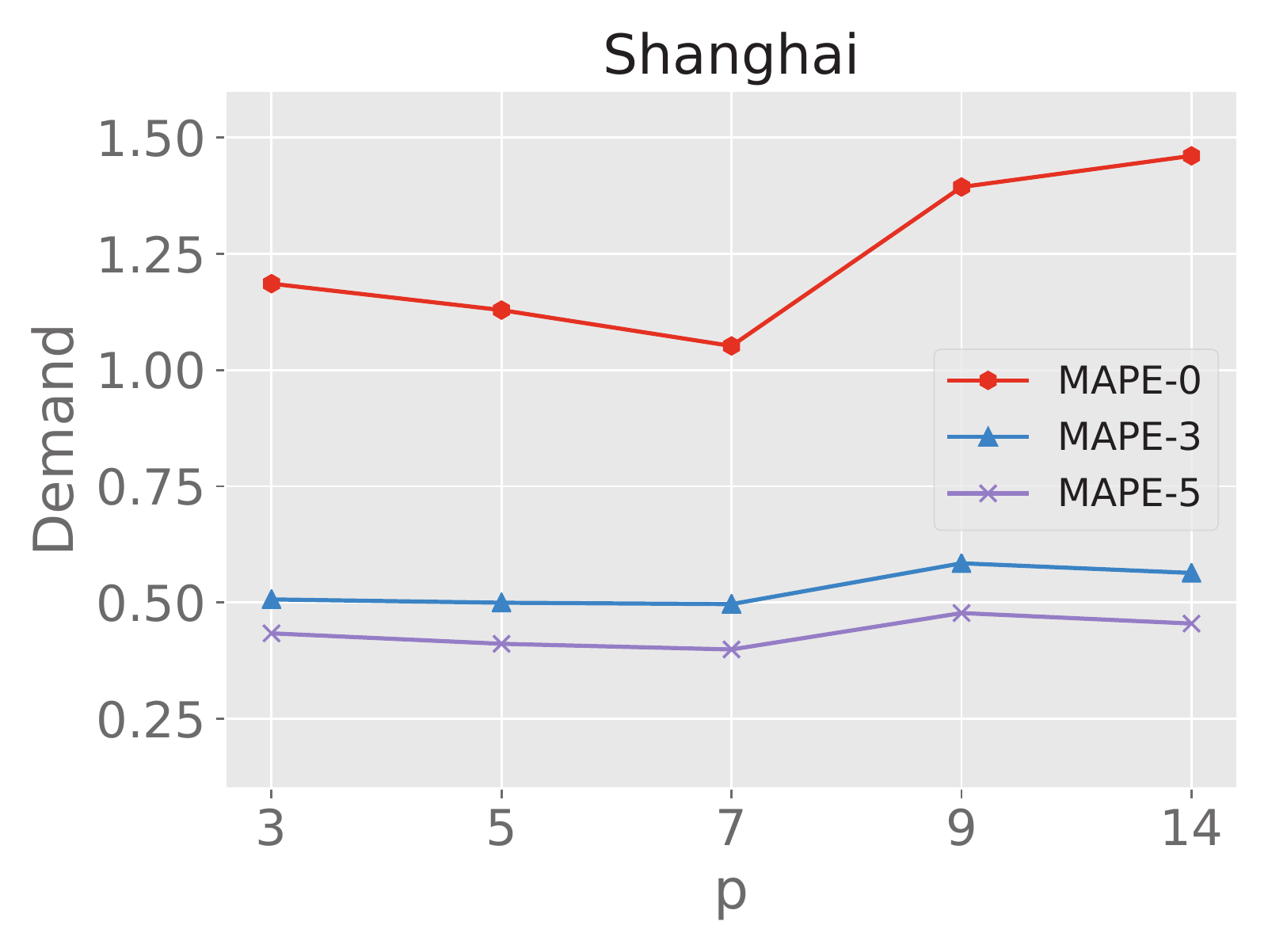}}	
	\subfigure{
		\label{fig:eva2:sub8}
		\includegraphics[width=0.23\textwidth]{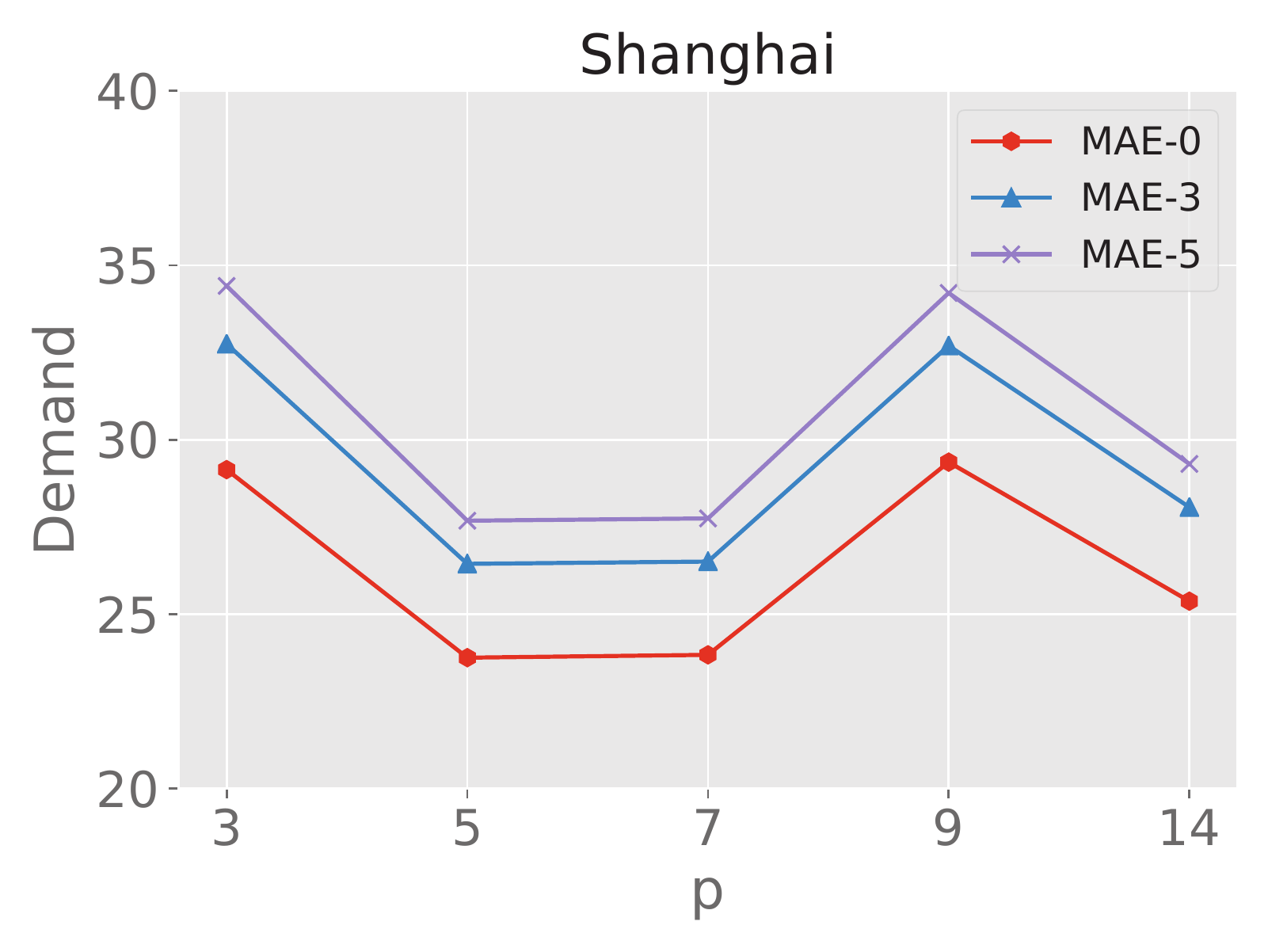}}
	\vspace{-2ex}
	\caption{\textbf{Performance on different numbers of historical time slots}}
	\label{fig:eva2} 
	\vspace{-2ex}
\end{figure*}

\paragraph{\textbf{Loss Weights}} As shown in Figure~\ref{fig:eva1}, we adjust the values of the loss weight pair $(\eta_{d}, \eta_{o})$ in both OD and Demand tasks. From the Figure, we can observe that:
\begin{itemize}
	\item Under the same loss weights, the results on different datasets show different patterns. On Beijing dataset, the model gains better overall performance with $(\eta_{d}, \eta_{o})=(0.7, 0.3)$. On Shanghai dataset, the weights around ($0.5, 0.5$) shows more advantages. We suppose this is related to the different intrinsic features of two datasets like the scale and sparsity of the data. For instance, the sparsity of Beijing dataset is more severe than Shanghai's, and then the accurate prediction on OD passenger demands relies more on the accuracy of Demand task, so the model needs a relatively larger $\eta_{o}$. 
	\item The Demand task is more sensitive to the loss weights than the OD Matrix task, especially on Shanghai dataset, which is possibly caused by the different demand distributions in different datasets. 
\end{itemize}

\paragraph{\textbf{Number of Historical Time Slots}}
Figure~\ref{fig:eva2} depicts the variation trend of model's performance under different numbers of historical time slots $P$ considered in each channel. Our findings are as follows:
\begin{itemize}
	\item The results show that the performance of the model does not always improve with an increasing value of $P$. Also, we can see an obvious trend from Figure~\ref{fig:eva2} that when we set $P$ based on a periodic value like the number of days in a week (i.e., $7$, $14$, etc.), the model shows a better performance overall.
	\item Combining Figure~\ref{fig:eva2} with Figure~\ref{fig:eva1}, Table~\ref{table:overall1} and Table~\ref{table:overall2}, we can see that the results on Shanghai dataset is worse and more fluctuant than on Beijing dataset. This indicates that this dataset has more complex temporal patterns, thus being more sensitive to the hyperparameter $P$.
\end{itemize}

\subsection{Scalability Analysis and Effectiveness of Pretraining}
\begin{figure*}[tb!]
	\centering
	\setlength{\abovecaptionskip}{0.15cm}
	\subfigure[]{
		\label{fig:ex6:sub1}
		\includegraphics[width=0.23\textwidth]{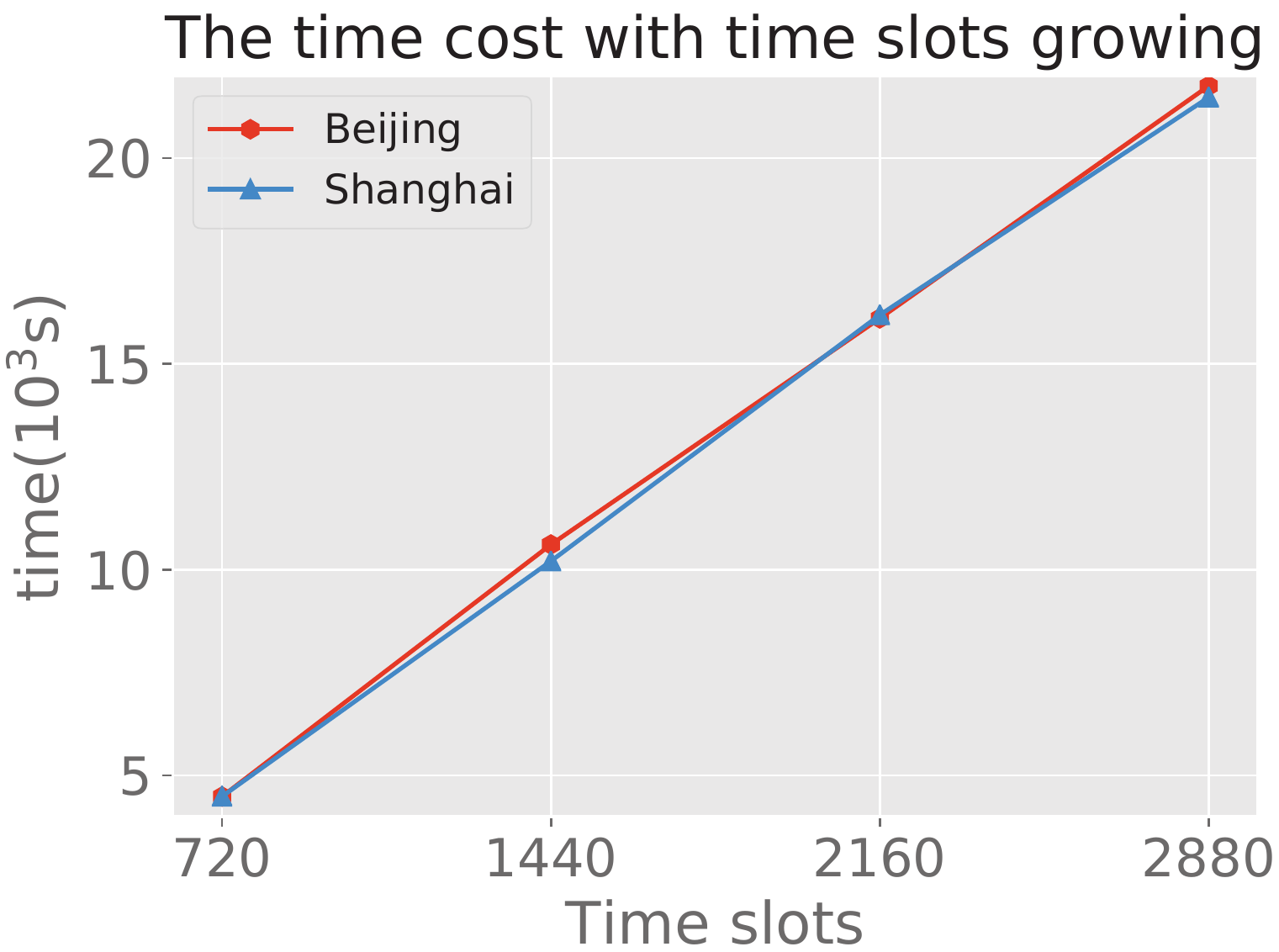}}
	\subfigure[]{
		\label{fig:ex6:sub2}
		\includegraphics[width=0.23\textwidth]{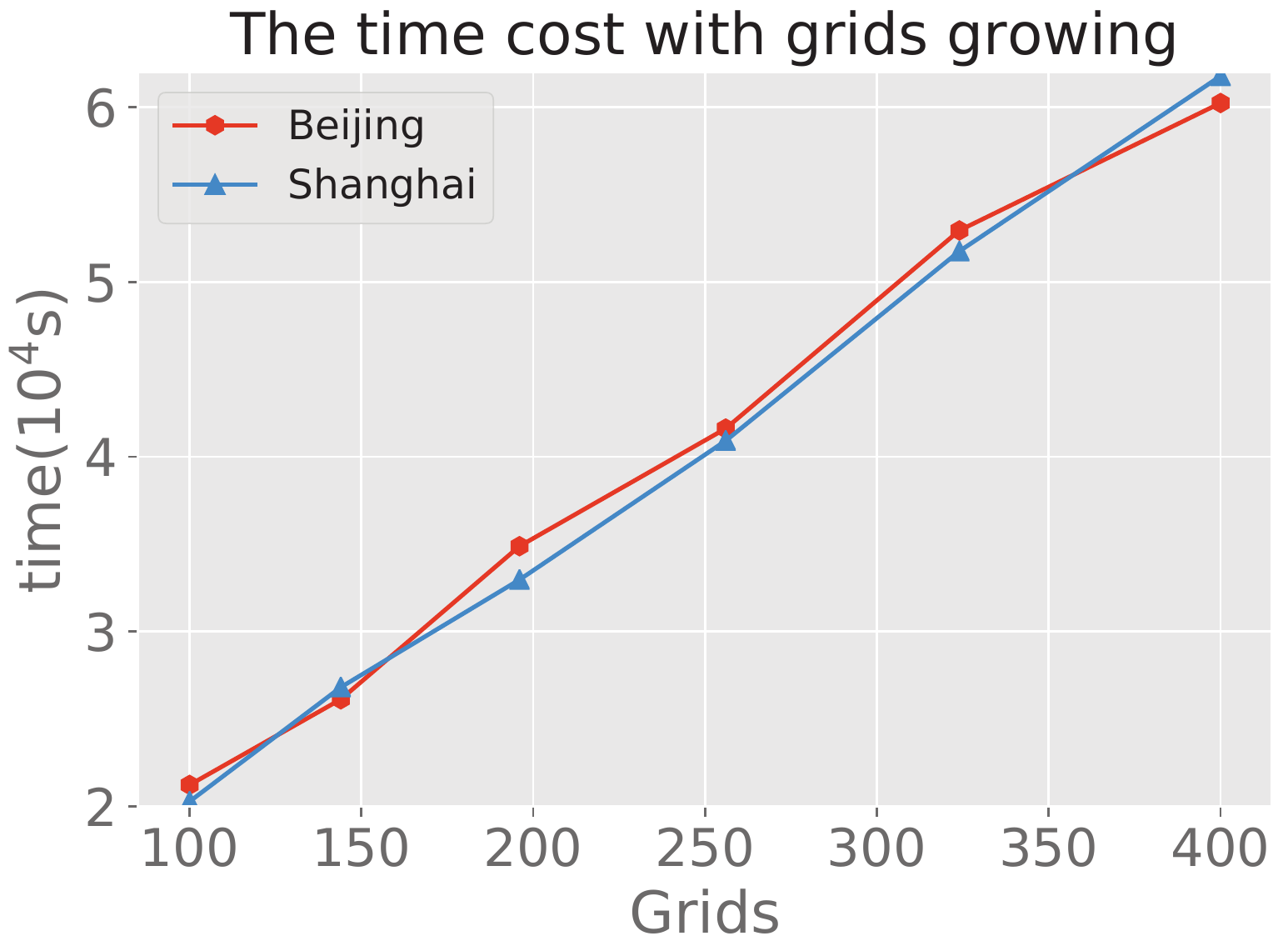}}
	\subfigure[]{
		\label{fig:ex6:sub3}
		\includegraphics[width=0.23\textwidth]{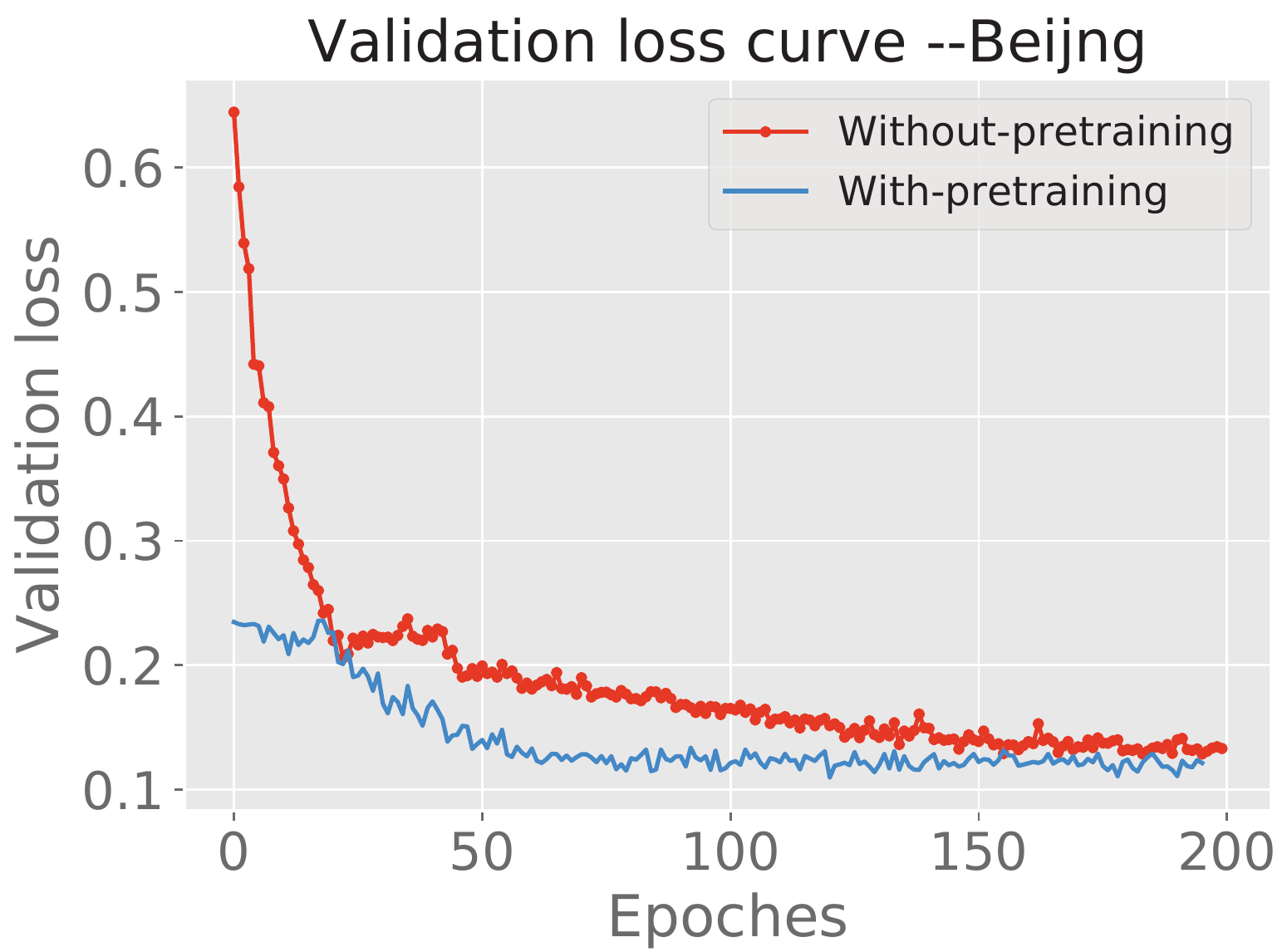}}
	\subfigure[]{
		\label{fig:ex6:sub4}
		\includegraphics[width=0.23\textwidth]{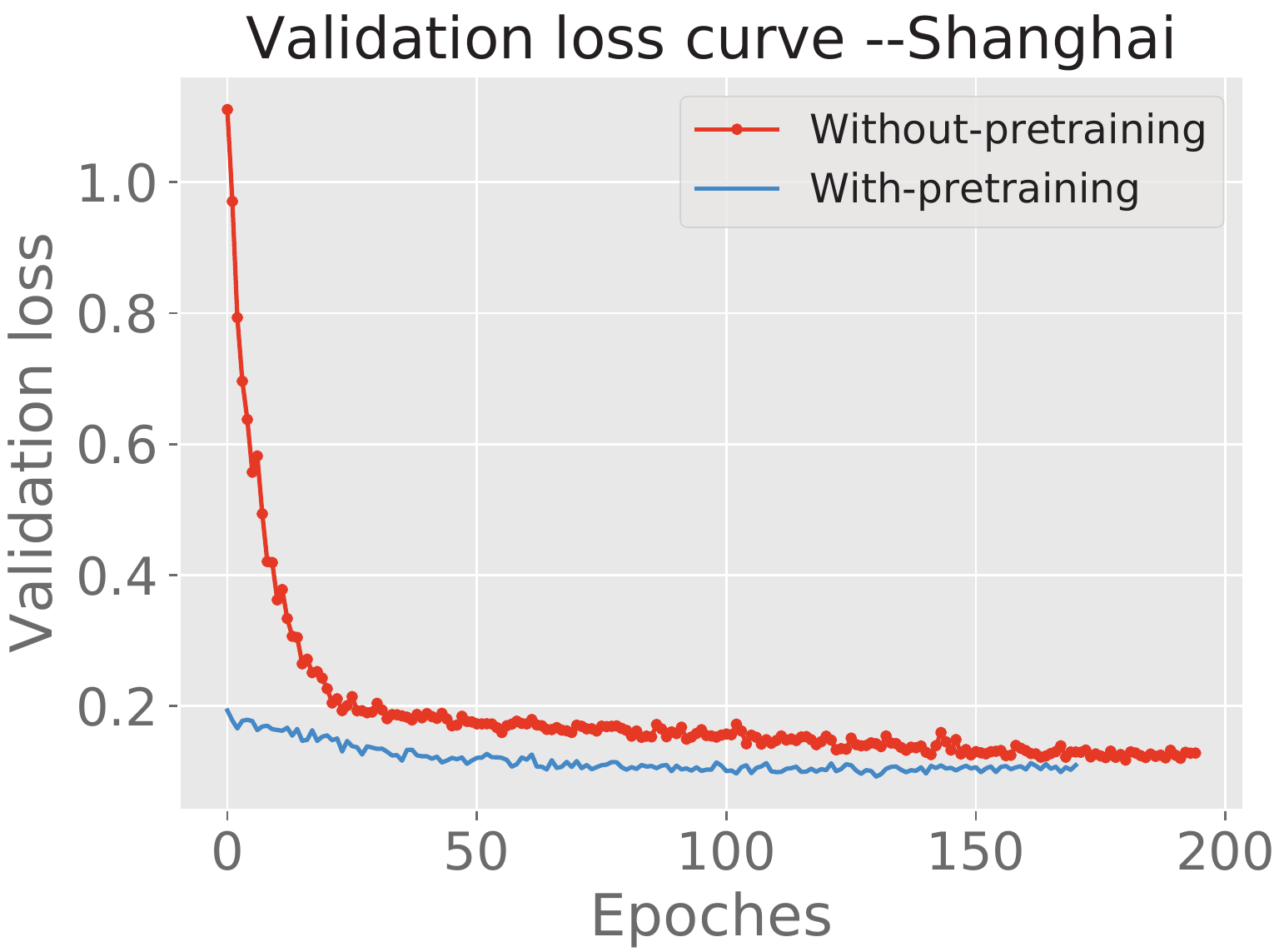}}
	\caption{\textbf{The time cost and the validation loss curve of our model}}
	\label{fig:ex6} 
\end{figure*}

In order to testify the model's scalability and the effectiveness of pretraining, we conduct several groups of experiments on both datasets and the results are shown in Figure \ref{fig:ex6}. In each group of experiments, except the variable parameter, the other parameters are set as default values following Section \ref{sec:expsetting}. 

\paragraph{\textbf{Scalability Analysis (RQ4)}} Figure \ref{fig:ex6:sub1} and \ref{fig:ex6:sub2} describe the training time cost. In the first group of experiment, we set the number of grids as $100$ while in the second one, we utilize $4$ months' data on both datasets. The summary of this part is as following:
\begin{itemize}
	\item It is obvious that as the number of the time slots and grids increase, the time cost of the model training is almost growing linearly, especially the time cost with the growing number of time slots.
	\item The time cost curve with increasing grids is less strictly linear. The possible reason is that the number of grids affects not only the data size but also the number of different neighbors in the spatial part which brings more complex impact to the training time than simple increasing the data size .
\end{itemize} 

\paragraph{\textbf{Effectiveness of Pretraining (RQ5)}} Figure \ref{fig:ex6:sub3} and \ref{fig:ex6:sub4} show the validation loss curve under the conditions of with-pretraining and without-pretraining. We can see from the figures:
\begin{itemize}
	\item The start loss of training process with pretraining is much lower than that without pretraining which means that the pretraining process would help fit the model on both tasks better.
	\item On both datasets,we can see that the pretraining offers faster convergence and a lower final loss, indicating that it is beneficial for our model. 
\end{itemize}

\begin{figure*}[tb!]
	\centering
	\setlength{\abovecaptionskip}{0.15cm}
	\subfigure[Mobility patterns at 8:00]{
		\label{fig:ex5:eight}
		\includegraphics[width=0.32\textwidth]{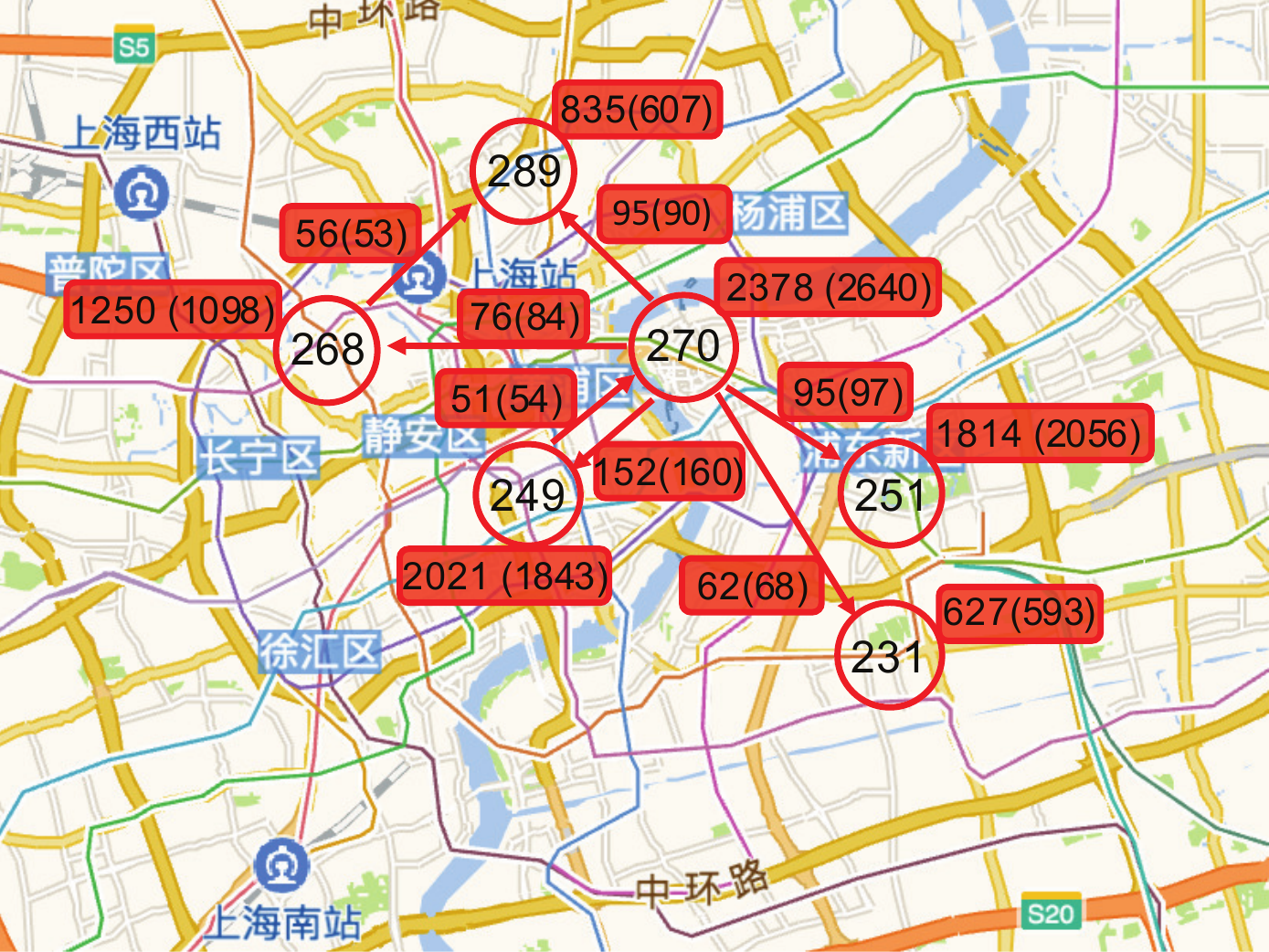}}
	\subfigure[Mobility patterns at 18:00]{
		\label{fig:ex5:eighteen}
		\includegraphics[width=0.32\textwidth]{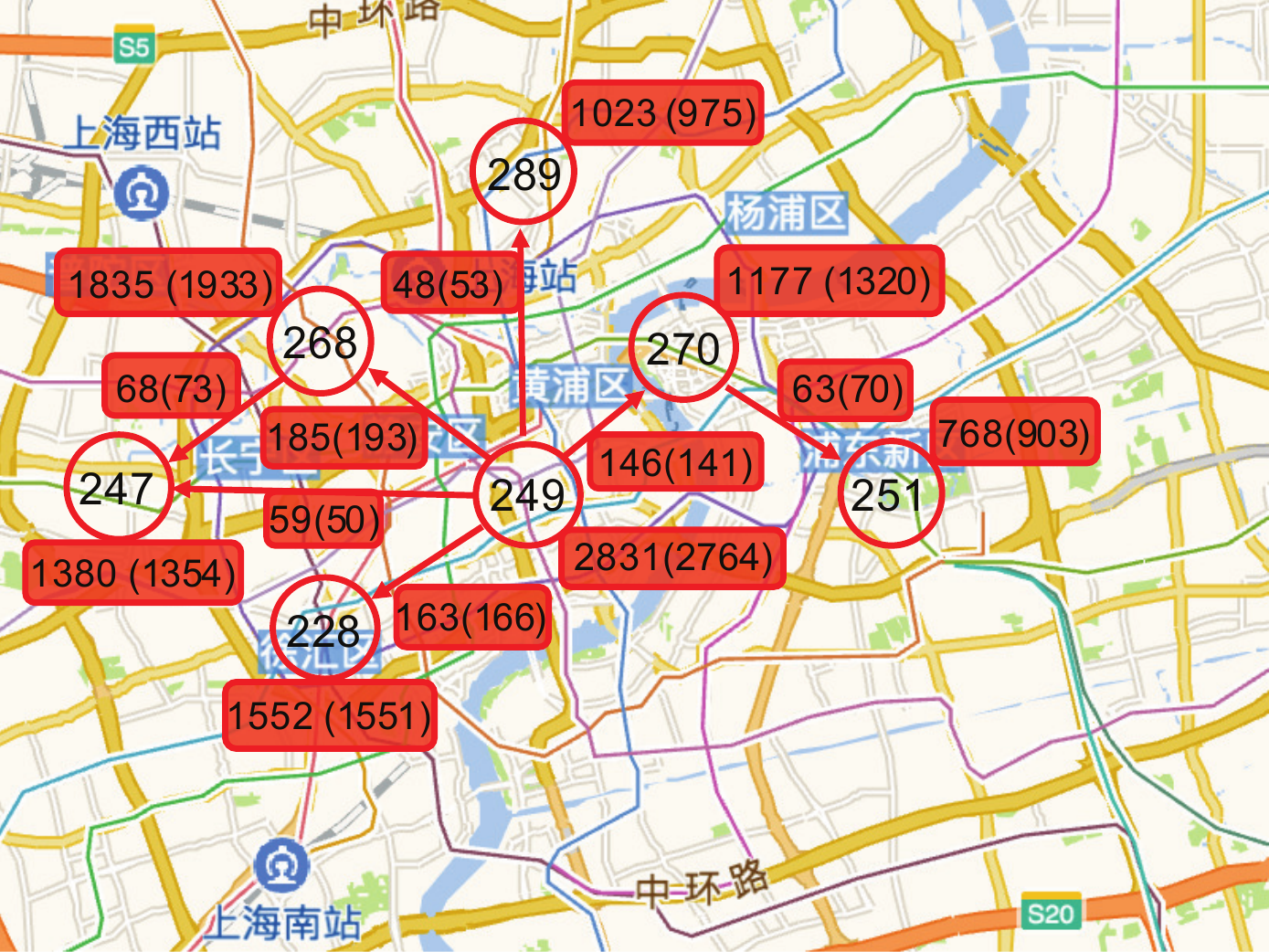}}
	\subfigure[Mobility patterns at 21:00]{
		\label{fig:ex5:twentyone}
		\includegraphics[width=0.32\textwidth]{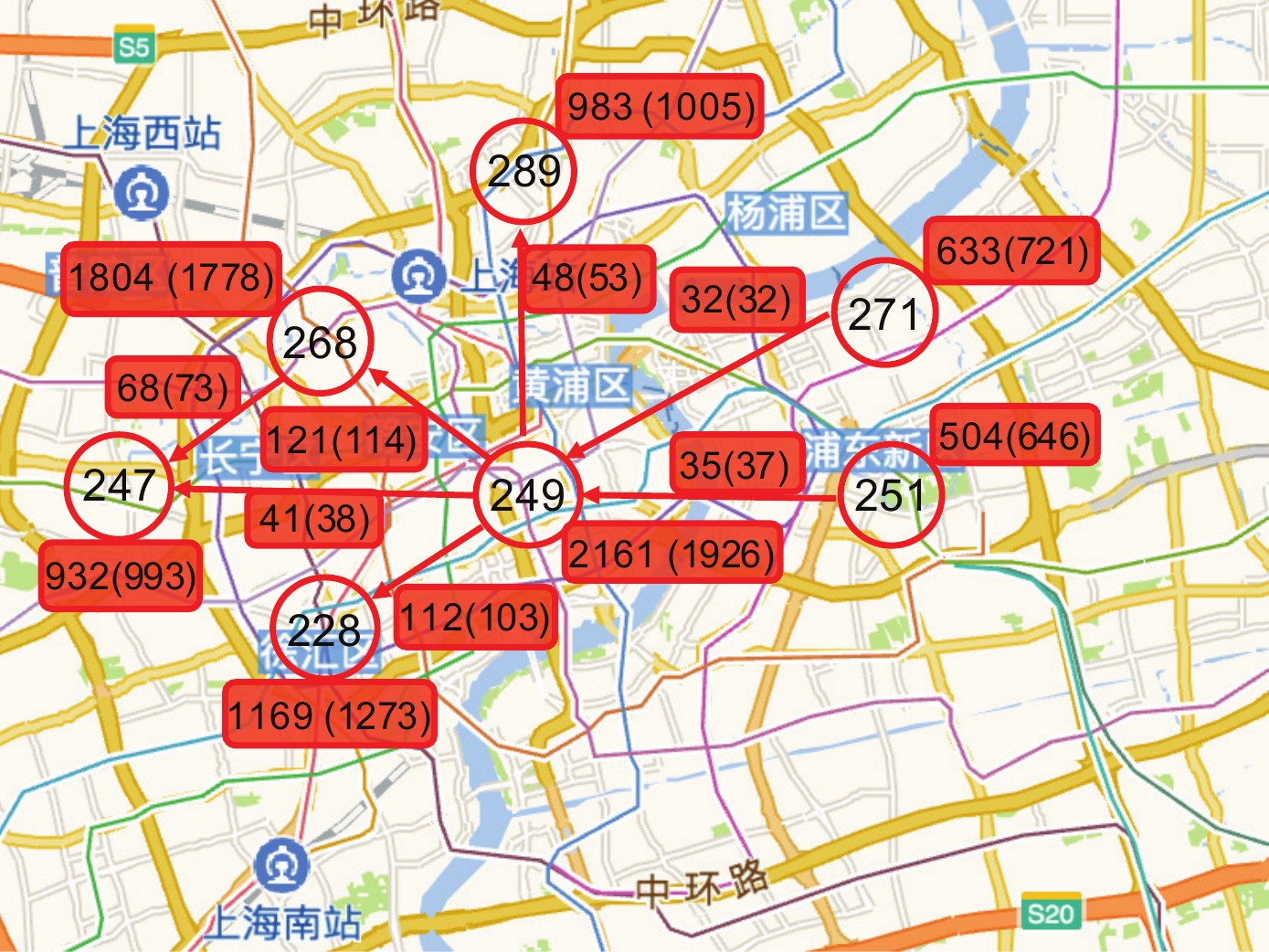}}	
	\caption{\textbf{The visualization of mobility patterns in Shanghai}}
	\label{fig:ex5} 
	\vspace{-2ex}
\end{figure*}

\subsection{Visualizing Learned Passenger Demand and Mobility Patterns (RQ6)}
In order to better understand the latent patterns learned by Gallat, we use Figure~\ref{fig:ex5} to visualize a part of passengers' demand patterns of the most popular regions in Shanghai predicted by Gallat during three different time slots. The red circle marks a region node on the map with its ID. The red rectangles and the arrows illustrate how many passengers transfer from one node to another. The first value in the rectangle is the predicted result of our model while the one in the bracket is the corresponding ground truth. From Figure~\ref{fig:ex5}, we can draw the following observations:
\begin{itemize}
	\item Figure~\ref{fig:ex5:eight} depicts the transferring relationship centered on node $270$ at 8:00 am. As node $270$ is a major residential area, it is the starting location for many people to leave home and march to several business districts and entertainment places, such as node $289$ and $251$, which cover a business district and the Shanghai Century Park, respectively. 
	\item Figure~\ref{fig:ex5:eighteen} demonstrates the passenger mobility centered on node $249$ which is a famous business district called Shanghai Xin Tian Di. Around 6:00 pm, a large amount of passengers leave this area to other residential or nightlife districts such as node $268$ and node $228$. There are also passengers going to node $247$, which has a large train station and an airport.
	\item Figure~\ref{fig:ex5:twentyone} illustrates the transferring relationship centered on node $249$ at 9:00 pm. Compared with Figure~\ref{fig:ex5:eight} and \ref{fig:ex5:eighteen}, it clearly shows that there are both passengers going to and leaving node $249$. It may be because $249$ also contains many nightlife venues like bars. There are also passengers travelling to node $247$ to take night trains or flights.
\end{itemize}

\section{Related Work}\label{related}
This section introduces the state-of-the-art studies related to our problem. We classify them into three categories.
\subsection{Sequence Based Prediction Problem}\label{traffic prediction}
Basically, our problem is a sequence based prediction problem. There are already many existing studies in this field \cite{tong2017simpler, wang2017deepsd, wang2019unified, wei2016zest, yao2018deep, hulot2018towards, liu2017functional, lai2018modeling, xu2020mtlm, xia2018exploring, guo2019streaming, chen2020sequence, su2020survey} which provide much inspiration for us. As the development of ride-hailing applications, we can collect more accuracy passenger demand data instead of traditional trajectory data from taxis to do research. Some studies \cite{tong2017simpler, wang2017deepsd, wang2019unified, wei2016zest, yao2018deep, hulot2018towards, liu2017functional, xu2020mtlm} have focused on passenger demand prediction via these kinds of data. 

\citet{tong2017simpler} put forward a unified linear regression model based on a distributed framework with more than 200 million features extracted from multi-source data to predict the passenger demand for each POI.
\citet{yao2018deep} propose a deep multi-view spatial-temporal network framework to model both spatial and temporal relations. Specifically, their model consists of three views: temporal view (modelling correlations between future demand values with near time points via LSTM), spatial view (modelling local spatial correlation via local CNN), and semantic view (modelling correlations among regions sharing similar temporal patterns). 
Wang and Wei et al. \cite{wang2019unified, wei2016zest} present a combined model to predict the passenger demand in a region at the future time slot and the model catches the temporal trend with a novel parameter and fuses the spatial and other related features by ANN based on multi-source data.
\citet{lai2018modeling} propose a novel framework, namely Long- and Short-term Time-series network (LSTNet) which uses the Convolution Neural Network (CNN) and the Recurrent Neural Network
(RNN) to extract short-term local dependency patterns among variables and to discover long-term patterns for time series trends.
There are some other works \cite{hulot2018towards, liu2017functional} based on bike sharing system.
\citet{hulot2018towards} focus on predicting the hourly demand for demand rentals and returns at each station of the bike sharing system and then they focus on determining decision intervals which are often used by bike sharing companies for their online rebalancing operations.
\citet{liu2017functional} develop a hierarchical station bike demand predictor which analyzes bike demands from functional zone level to station level. 
What's more, \citet{zhang2017deep, zhang2020flow} design different models to predict the in-flow and out-flow of people in a given area, but still they ignore the transferring relationship between different areas. 

All aforementioned methods have their own advantages but they model prediction problems as time series prediction problems and ignore the intrinsic connection and mobility between different areas or traffic interactions. Meanwhile, most of them need to draw support from sufficient multi-source data, which makes their model not that general and in low reproducibility.

\subsection{Graph Based Prediction Problem}\label{Graph-Based}
As traffic is based on networks consisting of lines and nodes, the traffic prediction can be naturally modeled into graph problems. Then graph-based methods \cite{deng2016latent, hu2020stochastic, gong2018network, li2018diffusion, liu2019contextualized, shi2020predicting, wang2019origin, geng2019spatiotemporal, seo2018structured, jiang2018deepurbanmomentum, su2019personalized, yu2017spatio, li2018go, jin2020recurrent, liang2020learning, wang2021gallat} can be used to solve them.

\citet{deng2016latent} define the traffic prediction based on the road network and given a series of road network snapshots, they propose a latent space model for road networks to learn the attributes of nodes in latent spaces which captures both topological and temporal properties.
\citet{li2018diffusion} propose to model the traffic flow as a diffusion process on a directed graph and introduce diffusion convolutional recurrent neural network for traffic forecasting that incorporates both spatial and temporal dependency in the traffic flow.
\citet{wang2019origin} design a grid-embedding based multi-task learning model where grid-embedding is designed to model the spatial relationships of different
areas, and LSTM based multi-task learning focuses on modelling temporal attributes and alleviating the data sparsity through subtasks.
\citet{geng2019spatiotemporal} present the spatial-temporal multi-graph convolution network where they first encode the non-Euclidean pair-wise correlations among regions into multiple graphs and then explicitly model these
correlations using multi-graph convolution. 
To utilize the global contextual information in modelling the temporal correlation, they further propose contextual gated recurrent neural network which augments recurrent neural network with a contextual-aware gating mechanism to re-weight different historical observations.
\citet{seo2018structured} propose the Graph Convolutional Recurrent Network (GCRN), a deep learning model able to predict structured sequences of data. Precisely, GCRN is a generalization of classical recurrent neural networks (RNN) to data structured by an arbitrary graph.
\citet{jiang2018deepurbanmomentum} build an online system via RNN to conduct the next short-term mobility predictions by using currently observed human mobility data.
\citet{yu2017spatio} propose a novel deep learning framework, spatiotemporal graph convolutional networks, to tackle the time series prediction problem in traffic domain. Instead of applying regular convolutional and recurrent units, they formulate the problem on graphs and build the model with complete convolutional structures, which enables much faster training speed with fewer parameters.

These works have either of the following two problems. Firstly, although they define the traffic prediction as a graph-based problem, they fail to provide a good representation of the graph, and in other word, they don't fully use the attributes of the graph, i.e., dynamics, direction and weight. Secondly, most of them utilize the traditional transductive graph convolution, but it is not friendly to cold-start nodes, that is, nodes have no interaction with others.
\subsection{Graph Representation Learning Method}\label{Representation}
Recently, the development of graph representation learning methods \cite{perozzi2014deepwalk, tang2015line, hamilton2017inductive, velivckovic2017graph, yin2017spatial, trivedi2018dyrep, zhou2018dynamic, zhang2018gaan, goyal2018dyngem, ma2018depthlgp, zhao2018rest, yin2019social, chen2019exploiting,  li2020community, liu2020dynamic, zheng2020reference} offer us some new thoughts to solve graph-based problems. 

The well-known methods Deepwalk \cite{perozzi2014deepwalk}, Line \cite{tang2015line} and node2vec \cite{grover2016node2vec} are both transductive and focused on static graphs. \citet{hamilton2017inductive} propose an inductive methods to learn representation on large graphs, which solves the cold start problem but it still only focuses on spatial perspective and ignores the dynamic attribute of graphs. \citet{velivckovic2017graph} employ attention mechanism in the node embedding process which gives us many inspirations but unfortunately it is also based on static graphs.
\citet{trivedi2018dyrep} build a novel modelling framework for dynamic graphs that posits representation learning as a latent mediation process bridging two observed processes namely dynamics of the network and dynamics on the networks. This model is further parameterized by a temporal-attentive representation network that encodes temporally evolving structural information into node representations, which in turn drives the nonlinear evolution of the observed graph dynamics. 
\citet{zhang2018gaan} propose a new network architecture, gated attention networks (GaAN). Unlike the traditional multi-head attention mechanism, which equally consumes all attention heads, GaAN uses a convolutional sub-network to control each attention head’s importance.
These two studies above both propose novel frameworks over dynamic graphs but they don't support the directed and weighted attributes.

\citet{goyal2018dyngem} present an efficient algorithm DynGEM based on recent advances in deep auto-encoders for graph embeddings. The major advantages of DynGEM include: the embedding is stable over time; it can handle growing dynamic graphs; and it has better running time than using static embedding methods on each snapshot of a dynamic graph.
\citet{ma2018depthlgp} propose a Deeply Transformed High-order Laplacian Gaussian Process (DepthLGP) method to infer embeddings for out-of-sample
nodes in dynamic graphs. DepthLGP combines the strength of nonparametric probabilistic modelling and deep learning.
However, these two works only focus on the representation learning over dynamic and weighted graphs but they neglect the directions of edges.

\section{Conclusion}\label{sec:conclusion}
In this paper, we define the passenger demand prediction problem in a new perspective, which is based on dynamic, directed and weighted graphs. To tackle this problem, we design a spatial-temporal attention network, Gallat, which includes spatial attention layer, temporal attention layer and transferring attention layer. In the spatial perspective, the spatial attention layer mimics the message passing process to formulate representation learning for each node by aggregating the information of all its neighbors. In this process, we add pre-weighted functions for three kinds of neighbors and utilize the attention mechanism to calculate the importance of different neighbors. In the temporal perspective, the temporal attention layer combines the learned representation of historical time slots and different channels via sefl-attention. Finally, we predict the passenger demand within an area first,  and learn a transferring probability via attention mechanism to obtain the final passenger mobility in the future time slots. We conduct extensive experiments to evaluate our model, which demonstrates that our model significantly outperforms all the baselines.


\begin{acks}
	This work is luckily supported by the National Key Research \& Development Program of China (2016YFB1000103) and Australian Research Council (Grant No.DP190101985, DP170103954).
	
\end{acks}

\bibliographystyle{ACM-Reference-Format}
\bibliography{ddw}


\begin{thebibliography}{56}


\ifx \showCODEN    \undefined \def \showCODEN     #1{\unskip}     \fi
\ifx \showDOI      \undefined \def \showDOI       #1{#1}\fi
\ifx \showISBNx    \undefined \def \showISBNx     #1{\unskip}     \fi
\ifx \showISBNxiii \undefined \def \showISBNxiii  #1{\unskip}     \fi
\ifx \showISSN     \undefined \def \showISSN      #1{\unskip}     \fi
\ifx \showLCCN     \undefined \def \showLCCN      #1{\unskip}     \fi
\ifx \shownote     \undefined \def \shownote      #1{#1}          \fi
\ifx \showarticletitle \undefined \def \showarticletitle #1{#1}   \fi
\ifx \showURL      \undefined \def \showURL       {\relax}        \fi
\providecommand\bibfield[2]{#2}
\providecommand\bibinfo[2]{#2}
\providecommand\natexlab[1]{#1}
\providecommand\showeprint[2][]{arXiv:#2}

\bibitem[\protect\citeauthoryear{Chen, Yin, Chen, Nguyen, Peng, and Li}{Chen
  et~al\mbox{.}}{2019}]%
        {chen2019exploiting}
\bibfield{author}{\bibinfo{person}{Hongxu Chen}, \bibinfo{person}{Hongzhi Yin},
  \bibinfo{person}{Tong Chen}, \bibinfo{person}{Quoc Viet~Hung Nguyen},
  \bibinfo{person}{Wen~Chih Peng}, {and} \bibinfo{person}{Xue Li}.}
  \bibinfo{year}{2019}\natexlab{}.
\newblock \showarticletitle{Exploiting Centrality Information with Graph
  Convolutions for Network Representation Learning}. In
  \bibinfo{booktitle}{\emph{2019 IEEE 35th International Conference on Data
  Engineering (ICDE)}}. \bibinfo{pages}{590--601}.
\newblock


\bibitem[\protect\citeauthoryear{Chen, Yin, Nguyen, Peng, Li, and Zhou}{Chen
  et~al\mbox{.}}{2020}]%
        {chen2020sequence}
\bibfield{author}{\bibinfo{person}{Tong Chen}, \bibinfo{person}{Hongzhi Yin},
  \bibinfo{person}{Quoc Viet~Hung Nguyen}, \bibinfo{person}{Wen~Chih Peng},
  \bibinfo{person}{Xue Li}, {and} \bibinfo{person}{Xiaofang Zhou}.}
  \bibinfo{year}{2020}\natexlab{}.
\newblock \showarticletitle{Sequence-aware factorization machines for temporal
  predictive analytics}. In \bibinfo{booktitle}{\emph{Proceedings - 2020 IEEE
  36th International Conference on Data Engineering, ICDE 2020}}.
  \bibinfo{pages}{1405--1416}.
\newblock


\bibitem[\protect\citeauthoryear{Cui, Henrickson, et~al\mbox{.}}{Cui
  et~al\mbox{.}}{2018}]%
        {cui2018traffic}
\bibfield{author}{\bibinfo{person}{Zhiyong Cui}, \bibinfo{person}{Kristian
  Henrickson}, {et~al\mbox{.}}} \bibinfo{year}{2018}\natexlab{}.
\newblock \showarticletitle{Traffic Graph Convolutional Recurrent Neural
  Network: A Deep Learning Framework for Network-Scale Traffic Learning and
  Forecasting}.
\newblock \bibinfo{journal}{\emph{ICLR}} (\bibinfo{year}{2018}).
\newblock


\bibitem[\protect\citeauthoryear{Deng et~al\mbox{.}}{Deng
  et~al\mbox{.}}{2016}]%
        {deng2016latent}
\bibfield{author}{\bibinfo{person}{Dingxiong Deng} {et~al\mbox{.}}}
  \bibinfo{year}{2016}\natexlab{}.
\newblock \showarticletitle{Latent space model for road networks to predict
  time-varying traffic}.
\newblock \bibinfo{journal}{\emph{SIGKDD}} (\bibinfo{year}{2016}).
\newblock


\bibitem[\protect\citeauthoryear{Geng, Li, et~al\mbox{.}}{Geng
  et~al\mbox{.}}{2019}]%
        {geng2019spatiotemporal}
\bibfield{author}{\bibinfo{person}{Xu Geng}, \bibinfo{person}{Yaguang Li},
  {et~al\mbox{.}}} \bibinfo{year}{2019}\natexlab{}.
\newblock \showarticletitle{Spatiotemporal multi-graph convolution network for
  ride-hailing demand forecasting}.
\newblock \bibinfo{journal}{\emph{AAAI}} (\bibinfo{year}{2019}).
\newblock


\bibitem[\protect\citeauthoryear{Gong, Li, Zhang, Liu, Zheng, and Kirsch}{Gong
  et~al\mbox{.}}{2018}]%
        {gong2018network}
\bibfield{author}{\bibinfo{person}{Yongshun Gong}, \bibinfo{person}{Zhibin Li},
  \bibinfo{person}{Jian Zhang}, \bibinfo{person}{Wei Liu}, \bibinfo{person}{Yu
  Zheng}, {and} \bibinfo{person}{Christina Kirsch}.}
  \bibinfo{year}{2018}\natexlab{}.
\newblock \showarticletitle{Network-wide Crowd Flow Prediction of Sydney Trains
  via customized Online Non-negative Matrix Factorization}.
\newblock  (\bibinfo{year}{2018}).
\newblock


\bibitem[\protect\citeauthoryear{Goyal, Kamra, He, and Liu}{Goyal
  et~al\mbox{.}}{2018}]%
        {goyal2018dyngem}
\bibfield{author}{\bibinfo{person}{Palash Goyal}, \bibinfo{person}{Nitin
  Kamra}, \bibinfo{person}{Xinran He}, {and} \bibinfo{person}{Yan Liu}.}
  \bibinfo{year}{2018}\natexlab{}.
\newblock \showarticletitle{Dyngem: Deep embedding method for dynamic graphs}.
\newblock \bibinfo{journal}{\emph{arXiv preprint}} (\bibinfo{year}{2018}).
\newblock


\bibitem[\protect\citeauthoryear{Grover and Leskovec}{Grover and
  Leskovec}{2016}]%
        {grover2016node2vec}
\bibfield{author}{\bibinfo{person}{Aditya Grover} {and} \bibinfo{person}{Jure
  Leskovec}.} \bibinfo{year}{2016}\natexlab{}.
\newblock \showarticletitle{node2vec: Scalable feature learning for networks}.
\newblock \bibinfo{journal}{\emph{SIGKDD}} (\bibinfo{year}{2016}).
\newblock


\bibitem[\protect\citeauthoryear{Guo, Yin, Wang, Chen, Zhou, and Quoc
  Viet~Hung}{Guo et~al\mbox{.}}{2019}]%
        {guo2019streaming}
\bibfield{author}{\bibinfo{person}{Lei Guo}, \bibinfo{person}{Hongzhi Yin},
  \bibinfo{person}{Qinyong Wang}, \bibinfo{person}{Tong Chen},
  \bibinfo{person}{Alexander Zhou}, {and} \bibinfo{person}{Nguyen Quoc
  Viet~Hung}.} \bibinfo{year}{2019}\natexlab{}.
\newblock \showarticletitle{Streaming Session-Based Recommendation}. In
  \bibinfo{booktitle}{\emph{Proceedings of the 25th ACM SIGKDD International
  Conference on Knowledge Discovery and Data Mining}}
  \emph{(\bibinfo{series}{KDD '19})}. \bibinfo{pages}{1569–1577}.
\newblock


\bibitem[\protect\citeauthoryear{Hamilton, Ying, et~al\mbox{.}}{Hamilton
  et~al\mbox{.}}{2017}]%
        {hamilton2017inductive}
\bibfield{author}{\bibinfo{person}{Will Hamilton}, \bibinfo{person}{Zhitao
  Ying}, {et~al\mbox{.}}} \bibinfo{year}{2017}\natexlab{}.
\newblock \showarticletitle{Inductive representation learning on large graphs}.
\newblock \bibinfo{journal}{\emph{NIPS}} (\bibinfo{year}{2017}).
\newblock


\bibitem[\protect\citeauthoryear{Hu, Yang, Guo, Jensen, and Xiong}{Hu
  et~al\mbox{.}}{2020}]%
        {hu2020stochastic}
\bibfield{author}{\bibinfo{person}{Jilin Hu}, \bibinfo{person}{Bin Yang},
  \bibinfo{person}{Chenjuan Guo}, \bibinfo{person}{Christian~S Jensen}, {and}
  \bibinfo{person}{Hui Xiong}.} \bibinfo{year}{2020}\natexlab{}.
\newblock \showarticletitle{Stochastic origin-destination matrix forecasting
  using dual-stage graph convolutional, recurrent neural networks}.
\newblock \bibinfo{journal}{\emph{ICDE}} (\bibinfo{year}{2020}).
\newblock


\bibitem[\protect\citeauthoryear{Hulot, Aloise, and Jena}{Hulot
  et~al\mbox{.}}{2018}]%
        {hulot2018towards}
\bibfield{author}{\bibinfo{person}{Pierre Hulot}, \bibinfo{person}{Daniel
  Aloise}, {and} \bibinfo{person}{Sanjay~Dominik Jena}.}
  \bibinfo{year}{2018}\natexlab{}.
\newblock \showarticletitle{Towards Station-Level Demand Prediction for
  Effective Rebalancing in Bike-Sharing Systems}.
\newblock \bibinfo{journal}{\emph{SIGKDD}} (\bibinfo{year}{2018}).
\newblock


\bibitem[\protect\citeauthoryear{Jiang, Song, et~al\mbox{.}}{Jiang
  et~al\mbox{.}}{2018}]%
        {jiang2018deepurbanmomentum}
\bibfield{author}{\bibinfo{person}{Renhe Jiang}, \bibinfo{person}{Xuan Song},
  {et~al\mbox{.}}} \bibinfo{year}{2018}\natexlab{}.
\newblock \showarticletitle{DeepUrbanMomentum: An Online Deep-Learning System
  for Short-Term Urban Mobility Prediction.}
\newblock \bibinfo{journal}{\emph{AAAI}} (\bibinfo{year}{2018}).
\newblock


\bibitem[\protect\citeauthoryear{Jin, Qu, Jin, and Ren}{Jin
  et~al\mbox{.}}{2020}]%
        {jin2020recurrent}
\bibfield{author}{\bibinfo{person}{Woojeong Jin}, \bibinfo{person}{Meng Qu},
  \bibinfo{person}{Xisen Jin}, {and} \bibinfo{person}{Xiang Ren}.}
  \bibinfo{year}{2020}\natexlab{}.
\newblock \showarticletitle{Recurrent Event Network: Autoregressive Structure
  Inference over Temporal Knowledge Graphs}. In
  \bibinfo{booktitle}{\emph{Proceedings of the 2020 Conference on Empirical
  Methods in Natural Language Processing (EMNLP)}}.
\newblock


\bibitem[\protect\citeauthoryear{Kingma and Ba}{Kingma and Ba}{2014}]%
        {kingma2014adam}
\bibfield{author}{\bibinfo{person}{Diederik~P Kingma} {and}
  \bibinfo{person}{Jimmy Ba}.} \bibinfo{year}{2014}\natexlab{}.
\newblock \showarticletitle{Adam: A method for stochastic optimization}.
\newblock \bibinfo{journal}{\emph{arXiv preprint}} (\bibinfo{year}{2014}).
\newblock


\bibitem[\protect\citeauthoryear{Kipf and Welling}{Kipf and Welling}{2017}]%
        {kipf2016semi}
\bibfield{author}{\bibinfo{person}{Thomas~N Kipf} {and} \bibinfo{person}{Max
  Welling}.} \bibinfo{year}{2017}\natexlab{}.
\newblock \showarticletitle{Semi-supervised classification with graph
  convolutional networks}.
\newblock \bibinfo{journal}{\emph{ICLR}} (\bibinfo{year}{2017}).
\newblock


\bibitem[\protect\citeauthoryear{Lai, Chang, Yang, et~al\mbox{.}}{Lai
  et~al\mbox{.}}{2018}]%
        {lai2018modeling}
\bibfield{author}{\bibinfo{person}{Guokun Lai}, \bibinfo{person}{Wei-Cheng
  Chang}, \bibinfo{person}{Yiming Yang}, {et~al\mbox{.}}}
  \bibinfo{year}{2018}\natexlab{}.
\newblock \showarticletitle{Modeling long-and short-term temporal patterns with
  deep neural networks}.
\newblock \bibinfo{journal}{\emph{SIGIR}} (\bibinfo{year}{2018}).
\newblock


\bibitem[\protect\citeauthoryear{Li, Cai, Deng, Wang, Sellis, and Xia}{Li
  et~al\mbox{.}}{2020}]%
        {li2020community}
\bibfield{author}{\bibinfo{person}{Jianxin Li}, \bibinfo{person}{Taotao Cai},
  \bibinfo{person}{Ke Deng}, \bibinfo{person}{Xinjue Wang},
  \bibinfo{person}{Timos Sellis}, {and} \bibinfo{person}{Feng Xia}.}
  \bibinfo{year}{2020}\natexlab{}.
\newblock \showarticletitle{Community-diversified influence maximization in
  social networks}.
\newblock \bibinfo{journal}{\emph{Information Systems}}  \bibinfo{volume}{92}
  (\bibinfo{year}{2020}), \bibinfo{pages}{101522}.
\newblock


\bibitem[\protect\citeauthoryear{Li, Zheng, Wang, and Zhou}{Li
  et~al\mbox{.}}{2018b}]%
        {li2018go}
\bibfield{author}{\bibinfo{person}{Lei Li}, \bibinfo{person}{Kai Zheng},
  \bibinfo{person}{Sibo Wang}, {and} \bibinfo{person}{Xiaofang Zhou}.}
  \bibinfo{year}{2018}\natexlab{b}.
\newblock \showarticletitle{Go Slow to Go Fast: Minimal On-Road Time Route
  Scheduling with Parking Facilities Using Historical Trajectory}.
\newblock \bibinfo{journal}{\emph{The VLDB Journal}}  \bibinfo{volume}{27}
  (\bibinfo{year}{2018}), \bibinfo{pages}{321--345}.
\newblock


\bibitem[\protect\citeauthoryear{Li, Yu, Shahabi, and Liu}{Li
  et~al\mbox{.}}{2018a}]%
        {li2018diffusion}
\bibfield{author}{\bibinfo{person}{Yaguang Li}, \bibinfo{person}{Rose Yu},
  \bibinfo{person}{Cyrus Shahabi}, {and} \bibinfo{person}{Yan Liu}.}
  \bibinfo{year}{2018}\natexlab{a}.
\newblock \showarticletitle{Diffusion convolutional recurrent neural network:
  Data-driven traffic forecasting}.
\newblock \bibinfo{journal}{\emph{ICLR}} (\bibinfo{year}{2018}).
\newblock


\bibitem[\protect\citeauthoryear{Liang and Zhang}{Liang and Zhang}{2020}]%
        {liang2020learning}
\bibfield{author}{\bibinfo{person}{Wenwei Liang} {and} \bibinfo{person}{Wei
  Zhang}.} \bibinfo{year}{2020}\natexlab{}.
\newblock \showarticletitle{Learning Social Relations and Spatiotemporal
  Trajectories for Next Check-in Inference}.
\newblock \bibinfo{journal}{\emph{IEEE Transactions on Neural Networks and
  Learning Systems}} (\bibinfo{year}{2020}), \bibinfo{pages}{1--11}.
\newblock


\bibitem[\protect\citeauthoryear{Liao, Zhang, et~al\mbox{.}}{Liao
  et~al\mbox{.}}{2018}]%
        {liao2018deep}
\bibfield{author}{\bibinfo{person}{Binbing Liao}, \bibinfo{person}{Jingqing
  Zhang}, {et~al\mbox{.}}} \bibinfo{year}{2018}\natexlab{}.
\newblock \showarticletitle{Deep sequence learning with auxiliary information
  for traffic prediction}.
\newblock \bibinfo{journal}{\emph{SIGKDD}} (\bibinfo{year}{2018}).
\newblock


\bibitem[\protect\citeauthoryear{Liu, Sun, et~al\mbox{.}}{Liu
  et~al\mbox{.}}{2017}]%
        {liu2017functional}
\bibfield{author}{\bibinfo{person}{Junming Liu}, \bibinfo{person}{Leilei Sun},
  {et~al\mbox{.}}} \bibinfo{year}{2017}\natexlab{}.
\newblock \showarticletitle{Functional zone based hierarchical demand
  prediction for bike system expansion}.
\newblock \bibinfo{journal}{\emph{SIGKDD}} (\bibinfo{year}{2017}).
\newblock


\bibitem[\protect\citeauthoryear{Liu, Qiu, Li, Wang, Ouyang, and Lin}{Liu
  et~al\mbox{.}}{2019}]%
        {liu2019contextualized}
\bibfield{author}{\bibinfo{person}{Lingbo Liu}, \bibinfo{person}{Zhilin Qiu},
  \bibinfo{person}{Guanbin Li}, \bibinfo{person}{Qing Wang},
  \bibinfo{person}{Wanli Ouyang}, {and} \bibinfo{person}{Liang Lin}.}
  \bibinfo{year}{2019}\natexlab{}.
\newblock \showarticletitle{Contextualized Spatial--Temporal Network for Taxi
  Origin-Destination Demand Prediction}.
\newblock \bibinfo{journal}{\emph{IEEE TITS}} (\bibinfo{year}{2019}).
\newblock


\bibitem[\protect\citeauthoryear{Liu, Huang, Yu, Song, Fan, and Dong}{Liu
  et~al\mbox{.}}{2020}]%
        {liu2020dynamic}
\bibfield{author}{\bibinfo{person}{Zhijun Liu}, \bibinfo{person}{Chao Huang},
  \bibinfo{person}{Yanwei Yu}, \bibinfo{person}{Peng Song},
  \bibinfo{person}{Baode Fan}, {and} \bibinfo{person}{Junyu Dong}.}
  \bibinfo{year}{2020}\natexlab{}.
\newblock \showarticletitle{Dynamic Representation Learning for Large-Scale
  Attributed Networks}. In \bibinfo{booktitle}{\emph{Proceedings of the 29th
  ACM International Conference on Information and Knowledge Management
  (CIKM)}}. \bibinfo{pages}{1005–1014}.
\newblock


\bibitem[\protect\citeauthoryear{Ma, Cui, and Zhu}{Ma et~al\mbox{.}}{2018}]%
        {ma2018depthlgp}
\bibfield{author}{\bibinfo{person}{Jianxin Ma}, \bibinfo{person}{Peng Cui},
  {and} \bibinfo{person}{Wenwu Zhu}.} \bibinfo{year}{2018}\natexlab{}.
\newblock \showarticletitle{DepthLGP: learning embeddings of out-of-sample
  nodes in dynamic networks}.
\newblock \bibinfo{journal}{\emph{AAAI}} (\bibinfo{year}{2018}).
\newblock


\bibitem[\protect\citeauthoryear{Mnih and Hinton}{Mnih and Hinton}{2009}]%
        {mnih2009a}
\bibfield{author}{\bibinfo{person}{Andriy Mnih} {and}
  \bibinfo{person}{Geoffrey~E Hinton}.} \bibinfo{year}{2009}\natexlab{}.
\newblock \showarticletitle{A scalable hierarchical distributed language
  model}. In \bibinfo{booktitle}{\emph{NIPS}}.
\newblock


\bibitem[\protect\citeauthoryear{Perozzi, Al-Rfou, et~al\mbox{.}}{Perozzi
  et~al\mbox{.}}{2014}]%
        {perozzi2014deepwalk}
\bibfield{author}{\bibinfo{person}{Bryan Perozzi}, \bibinfo{person}{Rami
  Al-Rfou}, {et~al\mbox{.}}} \bibinfo{year}{2014}\natexlab{}.
\newblock \showarticletitle{Deepwalk: Online learning of social
  representations}.
\newblock \bibinfo{journal}{\emph{SIGKDD}} (\bibinfo{year}{2014}).
\newblock


\bibitem[\protect\citeauthoryear{Ross}{Ross}{2015}]%
        {girshick2015fast}
\bibfield{author}{\bibinfo{person}{Girshick Ross}.}
  \bibinfo{year}{2015}\natexlab{}.
\newblock \showarticletitle{Fast-rnn}. In \bibinfo{booktitle}{\emph{ICCV}}.
\newblock


\bibitem[\protect\citeauthoryear{Seo, Defferrard, et~al\mbox{.}}{Seo
  et~al\mbox{.}}{2018}]%
        {seo2018structured}
\bibfield{author}{\bibinfo{person}{Youngjoo Seo}, \bibinfo{person}{Micha{\"e}l
  Defferrard}, {et~al\mbox{.}}} \bibinfo{year}{2018}\natexlab{}.
\newblock \showarticletitle{Structured sequence modeling with graph
  convolutional recurrent networks}.
\newblock \bibinfo{journal}{\emph{ICONIP}} (\bibinfo{year}{2018}).
\newblock


\bibitem[\protect\citeauthoryear{Shi, Yao, Guo, Li, Zhang, Ye, Li, and Liu}{Shi
  et~al\mbox{.}}{2020}]%
        {shi2020predicting}
\bibfield{author}{\bibinfo{person}{Hongzhi Shi}, \bibinfo{person}{Quanming
  Yao}, \bibinfo{person}{Qi Guo}, \bibinfo{person}{Yaguang Li},
  \bibinfo{person}{Lingyu Zhang}, \bibinfo{person}{Jieping Ye},
  \bibinfo{person}{Yong Li}, {and} \bibinfo{person}{Yan Liu}.}
  \bibinfo{year}{2020}\natexlab{}.
\newblock \showarticletitle{Predicting Origin-Destination Flow via
  Multi-Perspective Graph Convolutional Network}.
\newblock \bibinfo{journal}{\emph{ICDE}} (\bibinfo{year}{2020}).
\newblock


\bibitem[\protect\citeauthoryear{Su, Cong, Chen, Zheng, and Zheng}{Su
  et~al\mbox{.}}{2019}]%
        {su2019personalized}
\bibfield{author}{\bibinfo{person}{Han Su}, \bibinfo{person}{Guanglin Cong},
  \bibinfo{person}{Wei Chen}, \bibinfo{person}{Bolong Zheng}, {and}
  \bibinfo{person}{Kai Zheng}.} \bibinfo{year}{2019}\natexlab{}.
\newblock \showarticletitle{Personalized Route Description Based on Historical
  Trajectories}.
\newblock \bibinfo{journal}{\emph{CIKM}} (\bibinfo{year}{2019}),
  \bibinfo{pages}{79--88}.
\newblock


\bibitem[\protect\citeauthoryear{Su, Liu, Zheng, Zhou, and Zheng}{Su
  et~al\mbox{.}}{2020}]%
        {su2020survey}
\bibfield{author}{\bibinfo{person}{Han Su}, \bibinfo{person}{Shuncheng Liu},
  \bibinfo{person}{Bolong Zheng}, \bibinfo{person}{Xiaofang Zhou}, {and}
  \bibinfo{person}{Kai Zheng}.} \bibinfo{year}{2020}\natexlab{}.
\newblock \showarticletitle{A Survey of Trajectory Distance Measures and
  Performance Evaluation}.
\newblock \bibinfo{journal}{\emph{The VLDB Journal}}  \bibinfo{volume}{29}
  (\bibinfo{year}{2020}), \bibinfo{pages}{3--32}.
\newblock


\bibitem[\protect\citeauthoryear{Tang, Qu, et~al\mbox{.}}{Tang
  et~al\mbox{.}}{2015}]%
        {tang2015line}
\bibfield{author}{\bibinfo{person}{Jian Tang}, \bibinfo{person}{Meng Qu},
  {et~al\mbox{.}}} \bibinfo{year}{2015}\natexlab{}.
\newblock \showarticletitle{Line: Large-scale information network embedding}.
\newblock \bibinfo{journal}{\emph{WWW}} (\bibinfo{year}{2015}).
\newblock


\bibitem[\protect\citeauthoryear{Tong et~al\mbox{.}}{Tong
  et~al\mbox{.}}{2017}]%
        {tong2017simpler}
\bibfield{author}{\bibinfo{person}{Yongxin Tong} {et~al\mbox{.}}}
  \bibinfo{year}{2017}\natexlab{}.
\newblock \showarticletitle{The simpler the better: a unified approach to
  predicting original taxi demands based on large-scale online platforms}.
\newblock \bibinfo{journal}{\emph{SIGKDD}} (\bibinfo{year}{2017}).
\newblock


\bibitem[\protect\citeauthoryear{Trivedi, Farajtabar, Biswal, and Zha}{Trivedi
  et~al\mbox{.}}{2019}]%
        {trivedi2018dyrep}
\bibfield{author}{\bibinfo{person}{Rakshit Trivedi}, \bibinfo{person}{Mehrdad
  Farajtabar}, \bibinfo{person}{Prasenjeet Biswal}, {and}
  \bibinfo{person}{Hongyuan Zha}.} \bibinfo{year}{2019}\natexlab{}.
\newblock \showarticletitle{DyRep: Learning Representations over Dynamic
  Graphs}.
\newblock \bibinfo{journal}{\emph{ICLR}} (\bibinfo{year}{2019}).
\newblock


\bibitem[\protect\citeauthoryear{Vaswani, Shazeer, Parmar, Uszkoreit, Jones,
  Gomez, et~al\mbox{.}}{Vaswani et~al\mbox{.}}{2017}]%
        {vaswani2017attention}
\bibfield{author}{\bibinfo{person}{Ashish Vaswani}, \bibinfo{person}{Noam
  Shazeer}, \bibinfo{person}{Niki Parmar}, \bibinfo{person}{Jakob Uszkoreit},
  \bibinfo{person}{Llion Jones}, \bibinfo{person}{Gomez}, {et~al\mbox{.}}}
  \bibinfo{year}{2017}\natexlab{}.
\newblock \showarticletitle{Attention is all you need}.
\newblock \bibinfo{journal}{\emph{NIPS}} (\bibinfo{year}{2017}).
\newblock


\bibitem[\protect\citeauthoryear{Veli{\v{c}}kovi{\'c}, Cucurull,
  et~al\mbox{.}}{Veli{\v{c}}kovi{\'c} et~al\mbox{.}}{2017}]%
        {velivckovic2017graph}
\bibfield{author}{\bibinfo{person}{Petar Veli{\v{c}}kovi{\'c}},
  \bibinfo{person}{Guillem Cucurull}, {et~al\mbox{.}}}
  \bibinfo{year}{2017}\natexlab{}.
\newblock \showarticletitle{Graph attention networks}.
\newblock \bibinfo{journal}{\emph{arXiv preprint}} (\bibinfo{year}{2017}).
\newblock


\bibitem[\protect\citeauthoryear{Wang, Cao, Li, et~al\mbox{.}}{Wang
  et~al\mbox{.}}{2017}]%
        {wang2017deepsd}
\bibfield{author}{\bibinfo{person}{Dong Wang}, \bibinfo{person}{Wei Cao},
  \bibinfo{person}{Jian Li}, {et~al\mbox{.}}} \bibinfo{year}{2017}\natexlab{}.
\newblock \showarticletitle{DeepSD: Supply-demand prediction for online
  car-hailing services using deep neural networks}.
\newblock \bibinfo{journal}{\emph{ICDE}} (\bibinfo{year}{2017}).
\newblock


\bibitem[\protect\citeauthoryear{Wang, Lin, et~al\mbox{.}}{Wang
  et~al\mbox{.}}{2019a}]%
        {wang2019unified}
\bibfield{author}{\bibinfo{person}{Yuandong Wang}, \bibinfo{person}{Xuelian
  Lin}, {et~al\mbox{.}}} \bibinfo{year}{2019}\natexlab{a}.
\newblock \showarticletitle{A Unified Framework with Multi-source Data for
  Predicting Passenger Demands of Ride Services}.
\newblock \bibinfo{journal}{\emph{TKDD}} (\bibinfo{year}{2019}).
\newblock


\bibitem[\protect\citeauthoryear{Wang, Yin, Chen, Wo, et~al\mbox{.}}{Wang
  et~al\mbox{.}}{2019b}]%
        {wang2019origin}
\bibfield{author}{\bibinfo{person}{Yuandong Wang}, \bibinfo{person}{Hongzhi
  Yin}, \bibinfo{person}{Hongxu Chen}, \bibinfo{person}{Tianyu Wo},
  {et~al\mbox{.}}} \bibinfo{year}{2019}\natexlab{b}.
\newblock \showarticletitle{Origin-destination matrix prediction via graph
  convolution: a new perspective of passenger demand modeling}.
\newblock \bibinfo{journal}{\emph{SIGKDD}} (\bibinfo{year}{2019}).
\newblock


\bibitem[\protect\citeauthoryear{Wang, Yin, Chen, Liu, Wang, Wo, and Xu}{Wang
  et~al\mbox{.}}{2021}]%
        {wang2021gallat}
\bibfield{author}{\bibinfo{person}{Yuandong Wang}, \bibinfo{person}{Hongzhi
  Yin}, \bibinfo{person}{Tong Chen}, \bibinfo{person}{Chunyang Liu},
  \bibinfo{person}{Ben Wang}, \bibinfo{person}{Tianyu Wo}, {and}
  \bibinfo{person}{Jie Xu}.} \bibinfo{year}{2021}\natexlab{}.
\newblock \showarticletitle{Gallat: A Spatiotemporal Graph Attention Network
  for Passenger Demand Prediction}. In \bibinfo{booktitle}{\emph{2021 IEEE 37th
  International Conference on Data Engineering (ICDE)}}.
\newblock


\bibitem[\protect\citeauthoryear{Wei, Wang, et~al\mbox{.}}{Wei
  et~al\mbox{.}}{2016}]%
        {wei2016zest}
\bibfield{author}{\bibinfo{person}{Hua Wei}, \bibinfo{person}{Yuandong Wang},
  {et~al\mbox{.}}} \bibinfo{year}{2016}\natexlab{}.
\newblock \showarticletitle{Zest: a hybrid model on predicting passenger demand
  for chauffeured car service}.
\newblock \bibinfo{journal}{\emph{CIKM}} (\bibinfo{year}{2016}).
\newblock


\bibitem[\protect\citeauthoryear{Xia, Wang, Kong, Wang, Li, and Liu}{Xia
  et~al\mbox{.}}{2018}]%
        {xia2018exploring}
\bibfield{author}{\bibinfo{person}{Feng Xia}, \bibinfo{person}{Jinzhong Wang},
  \bibinfo{person}{Xiangjie Kong}, \bibinfo{person}{Zzhibo Wang},
  \bibinfo{person}{Jianxin Li}, {and} \bibinfo{person}{Chengfei Liu}.}
  \bibinfo{year}{2018}\natexlab{}.
\newblock \showarticletitle{Exploring Human Mobility Patterns in Urban
  Scenarios: A Trajectory Data Perspective}.
\newblock \bibinfo{journal}{\emph{IEEE Communications Magazine}}
  \bibinfo{volume}{56}, \bibinfo{number}{3} (\bibinfo{year}{2018}),
  \bibinfo{pages}{142--149}.
\newblock


\bibitem[\protect\citeauthoryear{Xu, Zhang, Cheng, and Xu}{Xu
  et~al\mbox{.}}{2020}]%
        {xu2020mtlm}
\bibfield{author}{\bibinfo{person}{Saijun Xu}, \bibinfo{person}{Ruoqian Zhang},
  \bibinfo{person}{Wanjun Cheng}, {and} \bibinfo{person}{Jiajie Xu}.}
  \bibinfo{year}{2020}\natexlab{}.
\newblock \showarticletitle{MTLM: a multi-task learning model for travel time
  estimation}.
\newblock \bibinfo{journal}{\emph{GeoInformatica}} (\bibinfo{year}{2020}).
\newblock


\bibitem[\protect\citeauthoryear{Xue, Qi, Xie, Zhang, Huang, and Li}{Xue
  et~al\mbox{.}}{2015}]%
        {xue2015solving}
\bibfield{author}{\bibinfo{person}{Andy~Yuan Xue}, \bibinfo{person}{Jianzhong
  Qi}, \bibinfo{person}{Xing Xie}, \bibinfo{person}{Rui Zhang},
  \bibinfo{person}{Jin Huang}, {and} \bibinfo{person}{Yuan Li}.}
  \bibinfo{year}{2015}\natexlab{}.
\newblock \showarticletitle{Solving the data sparsity problem in destination
  prediction}.
\newblock \bibinfo{journal}{\emph{VLDBJ}} (\bibinfo{year}{2015}).
\newblock


\bibitem[\protect\citeauthoryear{Yao, Wu, Ke, Tang, et~al\mbox{.}}{Yao
  et~al\mbox{.}}{2018}]%
        {yao2018deep}
\bibfield{author}{\bibinfo{person}{Huaxiu Yao}, \bibinfo{person}{Fei Wu},
  \bibinfo{person}{Jintao Ke}, \bibinfo{person}{Xianfeng Tang},
  {et~al\mbox{.}}} \bibinfo{year}{2018}\natexlab{}.
\newblock \showarticletitle{Deep multi-view spatial-temporal network for taxi
  demand prediction}.
\newblock \bibinfo{journal}{\emph{AAAI}} (\bibinfo{year}{2018}).
\newblock


\bibitem[\protect\citeauthoryear{Yin, Wang, Zheng, Li, Yang, and Zhou}{Yin
  et~al\mbox{.}}{2019}]%
        {yin2019social}
\bibfield{author}{\bibinfo{person}{Hongzhi Yin}, \bibinfo{person}{Qinyong
  Wang}, \bibinfo{person}{Kai Zheng}, \bibinfo{person}{Zhixu Li},
  \bibinfo{person}{Jiali Yang}, {and} \bibinfo{person}{Xiaofang Zhou}.}
  \bibinfo{year}{2019}\natexlab{}.
\newblock \showarticletitle{Social influence-based group representation
  learning for group recommendation}.
\newblock \bibinfo{journal}{\emph{ICDE}} (\bibinfo{year}{2019}).
\newblock


\bibitem[\protect\citeauthoryear{Yin, Wang, Wang, Chen, and Zhou}{Yin
  et~al\mbox{.}}{2017}]%
        {yin2017spatial}
\bibfield{author}{\bibinfo{person}{Hongzhi Yin}, \bibinfo{person}{Weiqing
  Wang}, \bibinfo{person}{Hao Wang}, \bibinfo{person}{Ling Chen}, {and}
  \bibinfo{person}{Xiaofang Zhou}.} \bibinfo{year}{2017}\natexlab{}.
\newblock \showarticletitle{Spatial-Aware Hierarchical Collaborative Deep
  Learning for POI Recommendation}.
\newblock \bibinfo{journal}{\emph{IEEE Transactions on Knowledge and Data
  Engineering}} \bibinfo{volume}{29}, \bibinfo{number}{11}
  (\bibinfo{year}{2017}), \bibinfo{pages}{2537--2551}.
\newblock


\bibitem[\protect\citeauthoryear{Yu, Yin, et~al\mbox{.}}{Yu
  et~al\mbox{.}}{2018}]%
        {yu2017spatio}
\bibfield{author}{\bibinfo{person}{Bing Yu}, \bibinfo{person}{Haoteng Yin},
  {et~al\mbox{.}}} \bibinfo{year}{2018}\natexlab{}.
\newblock \showarticletitle{Spatio-temporal graph convolutional networks: A
  deep learning framework for traffic forecasting}.
\newblock \bibinfo{journal}{\emph{IJCAI}} (\bibinfo{year}{2018}).
\newblock


\bibitem[\protect\citeauthoryear{Zhang, Shi, Xie, Ma, King, and Yeung}{Zhang
  et~al\mbox{.}}{2018}]%
        {zhang2018gaan}
\bibfield{author}{\bibinfo{person}{Jiani Zhang}, \bibinfo{person}{Xingjian
  Shi}, \bibinfo{person}{Junyuan Xie}, \bibinfo{person}{Hao Ma},
  \bibinfo{person}{Irwin King}, {and} \bibinfo{person}{Dit-Yan Yeung}.}
  \bibinfo{year}{2018}\natexlab{}.
\newblock \showarticletitle{Gaan: Gated attention networks for learning on
  large and spatiotemporal graphs}.
\newblock \bibinfo{journal}{\emph{arXiv preprint}} (\bibinfo{year}{2018}).
\newblock


\bibitem[\protect\citeauthoryear{Zhang, Zheng, and Qi}{Zhang
  et~al\mbox{.}}{2017}]%
        {zhang2017deep}
\bibfield{author}{\bibinfo{person}{Junbo Zhang}, \bibinfo{person}{Yu Zheng},
  {and} \bibinfo{person}{Dekang Qi}.} \bibinfo{year}{2017}\natexlab{}.
\newblock \showarticletitle{Deep spatio-temporal residual networks for citywide
  crowd flows prediction}.
\newblock \bibinfo{journal}{\emph{AAAI}} (\bibinfo{year}{2017}).
\newblock


\bibitem[\protect\citeauthoryear{Zhang, Zheng, Sun, and Qi}{Zhang
  et~al\mbox{.}}{2020}]%
        {zhang2020flow}
\bibfield{author}{\bibinfo{person}{Junbo Zhang}, \bibinfo{person}{Yu Zheng},
  \bibinfo{person}{Junkai Sun}, {and} \bibinfo{person}{Dekang Qi}.}
  \bibinfo{year}{2020}\natexlab{}.
\newblock \showarticletitle{Flow Prediction in Spatio-Temporal Networks Based
  on Multitask Deep Learning}.
\newblock \bibinfo{journal}{\emph{TKDE}} (\bibinfo{year}{2020}).
\newblock


\bibitem[\protect\citeauthoryear{Zhao, Shang, Wang, Zheng, Nguyen, and
  Zheng}{Zhao et~al\mbox{.}}{2018}]%
        {zhao2018rest}
\bibfield{author}{\bibinfo{person}{Yan Zhao}, \bibinfo{person}{Shuo Shang},
  \bibinfo{person}{Yu Wang}, \bibinfo{person}{Bolong Zheng},
  \bibinfo{person}{Quoc Viet~Hung Nguyen}, {and} \bibinfo{person}{Kai Zheng}.}
  \bibinfo{year}{2018}\natexlab{}.
\newblock \showarticletitle{REST: A Reference-based Framework for
  Spatio-temporal Trajectory Compression}.
\newblock \bibinfo{journal}{\emph{KDD}} (\bibinfo{year}{2018}),
  \bibinfo{pages}{2797--2806}.
\newblock


\bibitem[\protect\citeauthoryear{Zheng, Zhao, Lian, Zheng, Li, and Zhou}{Zheng
  et~al\mbox{.}}{2020}]%
        {zheng2020reference}
\bibfield{author}{\bibinfo{person}{Kai Zheng}, \bibinfo{person}{Yan Zhao},
  \bibinfo{person}{Defu Lian}, \bibinfo{person}{Bolong Zheng},
  \bibinfo{person}{Guanfeng Li}, {and} \bibinfo{person}{Xiaofang Zhou}.}
  \bibinfo{year}{2020}\natexlab{}.
\newblock \showarticletitle{Reference-Based Framework for Spatio-Temporal
  Trajectory Compression and Query Processing}.
\newblock \bibinfo{journal}{\emph{IEEE Transactions on Knowledge and Data
  Engineering}} \bibinfo{volume}{32}, \bibinfo{number}{11}
  (\bibinfo{year}{2020}), \bibinfo{pages}{2227--2240}.
\newblock


\bibitem[\protect\citeauthoryear{Zhou, Yang, et~al\mbox{.}}{Zhou
  et~al\mbox{.}}{2018}]%
        {zhou2018dynamic}
\bibfield{author}{\bibinfo{person}{Lekui Zhou}, \bibinfo{person}{Yang Yang},
  {et~al\mbox{.}}} \bibinfo{year}{2018}\natexlab{}.
\newblock \showarticletitle{Dynamic network embedding by modeling triadic
  closure process}.
\newblock \bibinfo{journal}{\emph{AAAI}} (\bibinfo{year}{2018}).
\newblock


\end{thebibliography}

\end{document}